%% file: sample.tex
\documentclass{ustthesis}

\usepackage{amsmath, amssymb, epsfig, enumerate, calc}
\usepackage{algorithm}
\usepackage{color,graphicx}
\usepackage{siunitx}
\usepackage{soul}
\usepackage{cite}
\usepackage{url}
\usepackage{graphics} 
\usepackage{subcaption}

\usepackage{bookmark}
\usepackage{hyperref} 
\usepackage{textcomp}
\usepackage{nth}
\usepackage{indentfirst}
\usepackage{siunitx}
\usepackage{changepage}
\usepackage[table]{xcolor}
\usepackage{tabularx}
\usepackage{booktabs, multirow}
\usepackage{multicol}
\usepackage{ragged2e}
\usepackage{threeparttable}
\usepackage{subcaption}
\usepackage{caption}
\usepackage[justification=raggedright]{caption}
\usepackage[noend]{algpseudocode}

\definecolor{ref}{RGB}{10,138,218}  
\definecolor{table_color}{RGB}{250,148,147}  

\usepackage[a4paper, margin=25mm,textheight=247mm,textwidth=145mm]{geometry} 

\newcommand{\tab}[1]{\hspace{3mm}}

\title{Scalable Vision-Based 3D Object Detection and Monocular Depth Estimation for Autonomous Driving}  
\author{Yuxuan Liu}     

\degree{\PhD}             

\department{Electronic and Computer Engineering}       

\advisor{Prof. Andrew Wing On POON}     

\coadvisor{Prof. Kam Tim WOO}     

\depthead{Prof. Andrew Wing On POON}    

\defencedate{2024}{01}{23}      

\begin{document}

\maketitle

%



\input{chapter/signature}

\input{chapter/ack}

\tableofcontents


\listoffigures


\listoftables


\input{chapter/abstract}




\chapter{Introduction}
\label{chp_intro}
\input{chapter/01_introduction}
\newpage

\chapter{Monocular 3D Detection}
\label{chp_background}
\input{chapter/03_mono3d}
\newpage

\chapter{Stereo 3D Detection}
\label{chp_background}
\input{chapter/04_stereo3d}
\newpage

\chapter{Joint Dataset Training for Monocular 3D Detection}
\label{chp_background}

\input{chapter/06_scaleup}
\newpage

\chapter{Unsupervised Monocular Depth Prediction}
\label{chp_background}
\input{chapter/05_monodepth}
\newpage

\chapter{Conclusion}
\label{chp_background}
\input{chapter/07_conclusion}
\newpage

\bibliographystyle{plain}

\bibliography{references}

\input{chapter/s6_appendices}

\end{document}

%% file: chapter/signature.tex
\signature
		\begin{figure}[!hb]
		\begin{tabular}{ll}
		Thesis Examination Committee &  \\[8pt]
	1. Prof. Andrew W. POON (Supervisor)   &  Department of Electronic and Computer Engineering \\[8pt]
            2. Prof. Kam Tim WOO (Co-Supervisor)   &  Department of Electronic and Computer Engineering \\[8pt]
		3. Prof. Ling Shi              &  Department of Electronic and Computer Engineering \\[8pt]
		4. Prof. Wei Zhang               &  Department of Electronic and Computer Engineering \\[8pt]
		5. Prof. Qiong Luo               &  Department of Computing Science and Engineering \\[8pt]
		6. Prof. Rong Xiong (External Examiner)  & College of Control Science and Engineering \\[8pt]
		                                   &  Zhejiang University \\[8pt]
		\end{tabular}
		\end{figure}
\endsignature

%% file: chapter/ack.tex
\acknowledgments

First and foremost, I am profoundly grateful to Prof. Liu for his invaluable guidance and mentorship throughout my doctoral studies in Robotics and Autonomous Driving. His pragmatic approach to problem-solving and his deep insights into the broader implications of robotic systems have been instrumental in shaping my academic journey. It has been a privilege to be part of his team and to contribute to the esteemed robotics institute over these transformative four years.

Moreover, my sincere appreciation extends to the members of my thesis examination committee: Prof Andrew Wing On POON, Prof. Kam Tim Woo, Prof. Ling Shi, Prof. Wei Zhang, Prof. Qiong Luo, and Prof. Rong Xiong. Their meticulous review and insightful suggestions have significantly enhanced the quality of my thesis.

The successful completion of my research is a testament not only to my efforts but also to the collaborative spirit and support of my supervisor and colleagues. I am immensely thankful for the inspiration and motivation derived from my interactions with fellow students, whose passion and intelligence have been a constant source of encouragement. Special thanks go to my co-workers, particularly Dr. Hengli Wang, Dr. Zhenhua Xu, Dr. Peng Yun, Dr. Huaiyang Huang, Dr. Peide Cai, Dr. Yuxuang Sun, Dr. Rui Fan, Dr. XinXing Chen, Mr. Xiaoyang Yan, Mr. Fulong Ma, Mr. Xiangcheng Hu, and Mrs. Lu Gan. Additionally, my gratitude extends to the members of the RAM Lab, especially Dr. Jianhao Jiao, Dr. Sukai Wang, Ms. Xiaodong Mei, Mr. Bowen Yang, Mr. Jingwen Yu, Mr. Tianyu Liu, Mr. Yingbing Chen, and Mr. Mingkai Tang. The bonds formed during these four and a half years are invaluable and I eagerly anticipate their continuation in the future.

Furthermore, I am also deeply thankful to teachers Shiomi and Aisu for their support during challenging times and for encouraging me to expand my horizons. Their guidance has opened new avenues in my life, and I am excited about the prospects of forging new connections in the future.

Lastly, but certainly not least, I must acknowledge the unwavering support of my family. The past four years have been a period of significant challenges and rapid changes, both in my personal life and in the broader societal context. The constant love and support from those closest to me, especially my family, have been my bedrock. Their encouragement has been pivotal in reaching this milestone in my academic career.

\endacknowledgments

%% file: chapter/abstract.tex
\begin{abstract}

3D perception serves as a cornerstone in the realm of autonomous driving. Vision-based 3D perception methods, which rely solely on camera inputs to reconstruct a 3D environment, have seen significant advancements due to the proliferation of deep learning techniques. Despite these strides, existing frameworks still encounter performance bottlenecks and often necessitate substantial amounts of LiDAR-annotated data, limiting their practical deployment across diverse autonomous driving platforms at a larger scale.

This dissertation is a multifaceted contribution to the advancement of vision-based 3D perception technologies. In the first segment, the thesis introduces structural enhancements to both monocular and stereo 3D object detection algorithms. By integrating ground-referenced geometric priors into monocular detection models, this research achieves unparalleled accuracy in benchmark evaluations for monocular 3D detection. Concurrently, the work refines stereo 3D detection paradigms by incorporating insights and inferential structures gleaned from monocular networks, thereby augmenting the operational efficiency of stereo detection systems.

The second segment is devoted to data-driven strategies and their real-world applications in 3D vision detection. A novel training regimen is introduced that amalgamates datasets annotated with either 2D or 3D labels. This approach not only augments the detection models through the utilization of a substantially expanded dataset but also facilitates economical model deployment in real-world scenarios where only 2D annotations are readily available.

Lastly, the dissertation presents an innovative pipeline tailored for unsupervised depth estimation in autonomous driving contexts. Extensive empirical analyses affirm the robustness and efficacy of this newly proposed pipeline. Collectively, these contributions lay a robust foundation for the widespread adoption of vision-based 3D perception technologies in autonomous driving applications.

\end{abstract}

%% file: chapter/01_introduction.tex
\section{Autonomous Driving Application and System}

In the contemporary era, marked by significant strides in deep learning and data-driven methodologies, autonomous driving has evolved from being an academic subject to a real-world solution with widespread recognition. Leading automotive manufacturers like Tesla and Nio have incorporated co-pilot functionalities into their vehicles \cite{TeslaAI}, thereby reducing the cognitive load on human drivers. Furthermore, autonomous technologies have penetrated other sectors; for example, fully automated systems in advanced ports such as Guangzhou Nansha Port have bolstered both safety and operational efficiency \cite{GuangzhouPort}. The COVID-19 pandemic has acted as a catalyst, expediting the integration of autonomous vehicles into urban landscapes for contactless delivery services \cite{Hercules}. The continual advancements in artificial intelligence and engineering are projected to expand the range of applications for autonomous driving, further enhancing both safety and efficiency in everyday life.

\begin{figure}[htb]
\includegraphics[width=1.0\columnwidth]{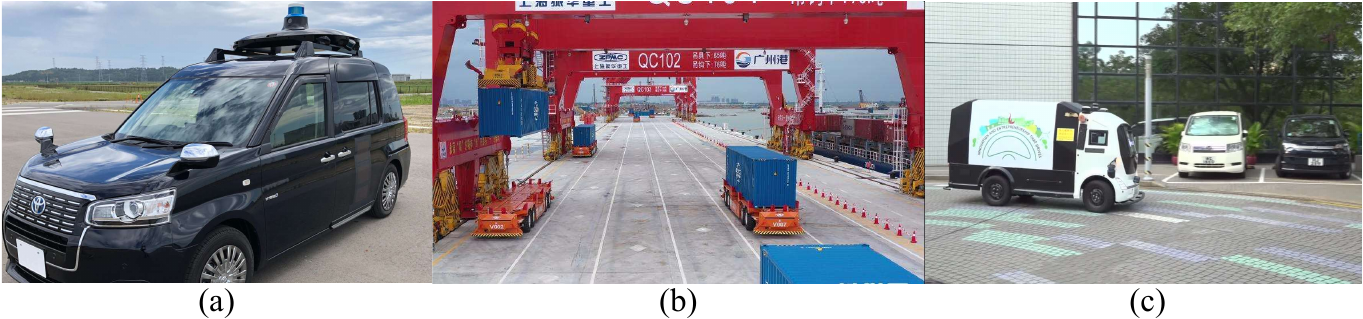}
\caption{Examples of autonomous driving applications. (a): autonomous Japan taxi tested in Tokyo \cite{Autoware}; (b): autonomous logistic platform fueling the Guangzhou Nansha Port \cite{GuangzhouPort}; (c): autonomous vehicle for contactless delivery services in HKUST campus \cite{Hercules}}
\label{fig_autoware}
\end{figure}

The software architecture of standard autonomous driving runtime systems is intrinsically modular, aligning well with the distributed computational paradigm inherent in robotics \cite{Autoware, Hercules, ROS2}. A reference structure, inspired by the Autoware platform\cite{Autoware}, serves to illustrate this modularity in figure~\ref{fig_autoware}. In a typical setup, an array of sensors including Global Navigation Satellite Systems (GNSS), Radar, LiDAR, Inertial Measurement Units (IMU), and cameras, serve as the eyes and ears of the vehicle, continuously capturing environmental data. This information is processed through a sequence of specialized modules for localization, perception, and planning/control, culminating in real-time vehicular control outputs via the interface.

\begin{figure}[htb]
\includegraphics[width=1.0\columnwidth]{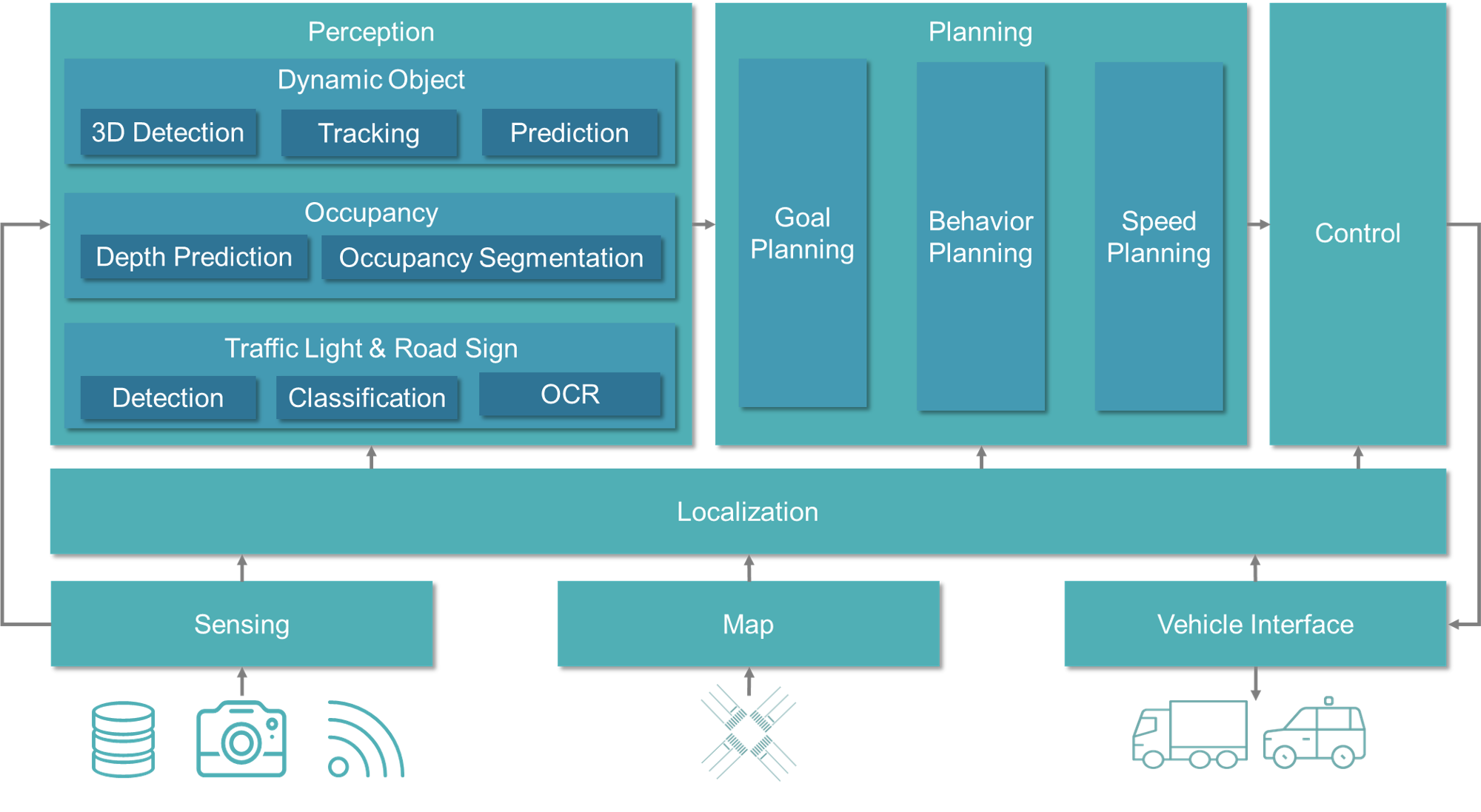}
\caption{An illustrative modular framework for autonomous driving systems \cite{Autoware}, drawn from Autoware's design principles. The focus of this thesis primarily lies in the 3D-related tasks inside the vision-based perception module, a critical component within the larger perception challenge.}
\label{fig_autoware}
\end{figure}

The role of the perception module is to interpret this sensor data and construct a real-time model of the surrounding environment. Inputs for this module are primarily sourced from recent frames obtained from LiDARs and cameras. The outputs are sophisticated 3D environmental information that feeds into downstream tasks. Advances in deep learning have empowered perception modules to generate increasingly complex environmental representations \cite{RNGDetPP, CenterLineDet}. However, this increasing complexity comes with its own set of challenges, including heightened data and computational requirements, especially in real-world applications.

\begin{figure}[htb]
\includegraphics[width=1.0\columnwidth]{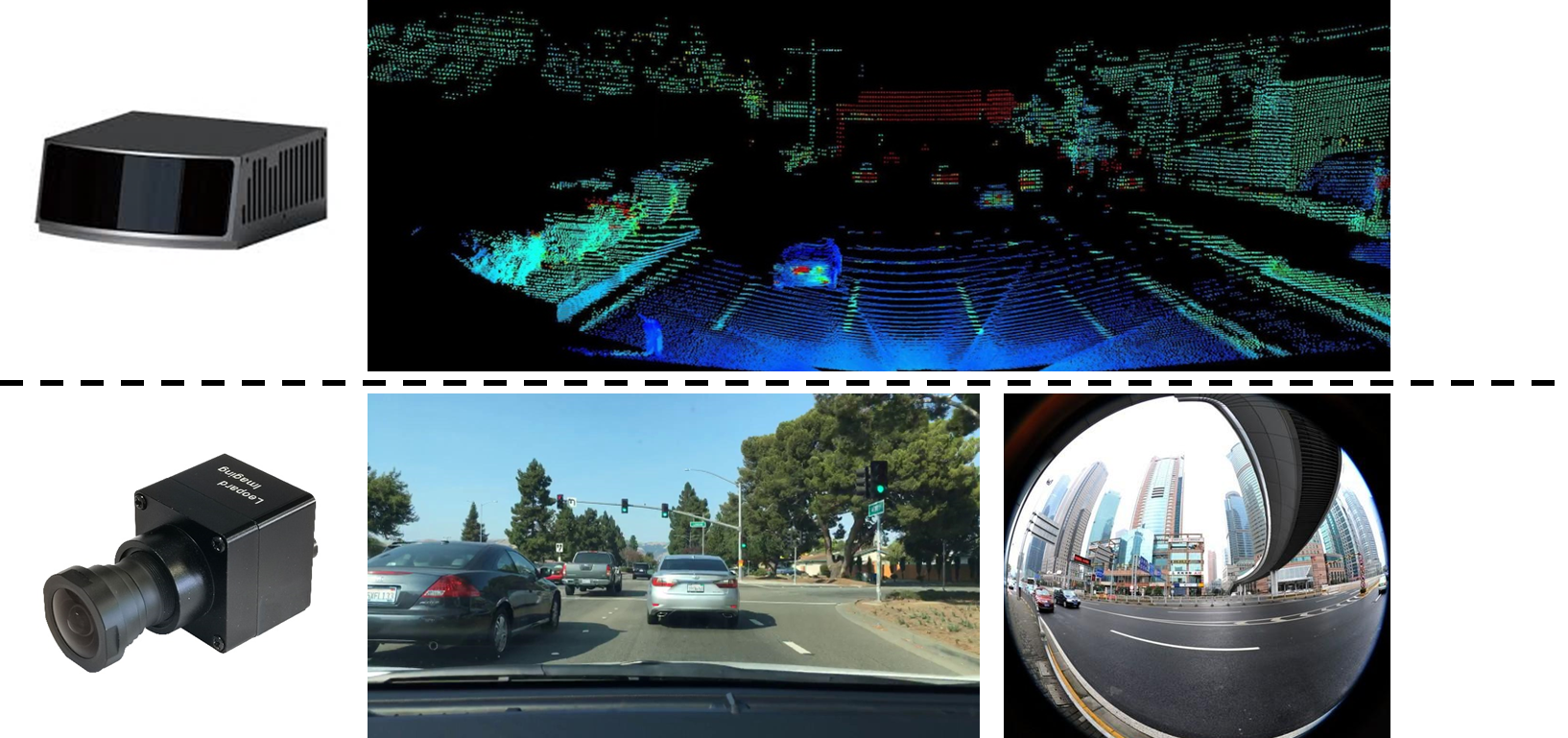}
\caption{Illustrative example for a LiDAR and a camera.}
\label{fig_sensors}
\end{figure}

The integration of advanced sensing technologies in autonomous vehicles is a critical factor in their ability to perceive and interact with the surrounding environment effectively. Among these technologies, LiDAR systems and cameras are predominantly utilized shown in figure~\ref{fig_sensors}. LiDAR systems are especially valued for their capacity to provide precise three-dimensional measurements within their field of view (FOV). Ongoing advancements in manufacturing techniques are anticipated to enhance the resolution and reduce the cost of LiDAR systems, making them more accessible and efficient. Concurrently, the use of cameras in vehicles has become widespread. Cameras, in many respects, are considered to deliver comprehensive semantic information, which is crucial for human-like perception and decision-making in driving scenarios. There is a growing consensus that a camera-only configuration might represent the most cost-effective approach for minimal hardware deployment in self-driving cars. This minimal configuration approach aligns with the intention to reduce hardware costs while maintaining functional efficacy. This thesis aims to assess the feasibility, challenges, and potential of relying solely on camera-based systems in the autonomous driving domain.

\section{Vision-Based 3D Perception System}

\begin{figure}
\includegraphics[width=1.0\columnwidth]{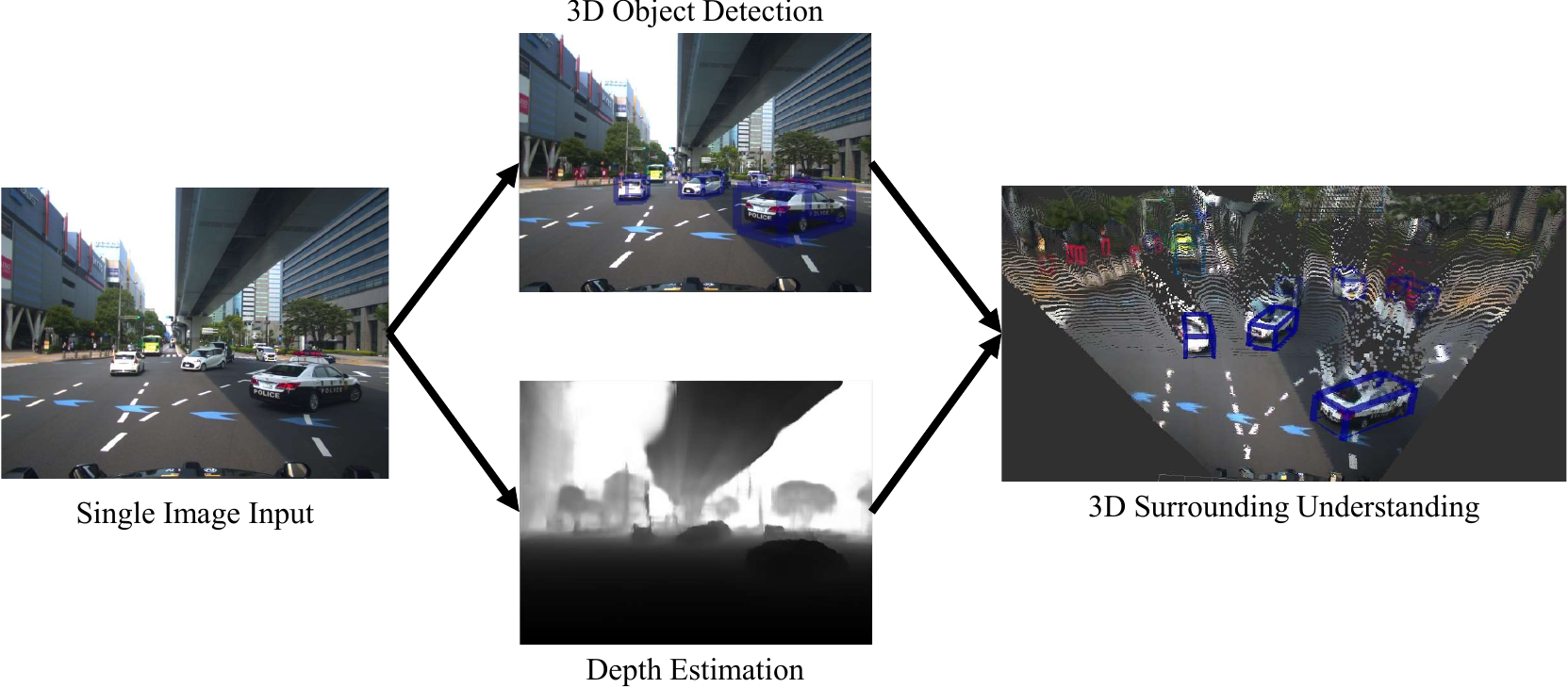}
\caption{An illustration of vision-based 3D object detection and depth prediction for 3D scene perception from a single image input. The figure is created with models proposed in Chapter 4 and Chapter 5 on completely unseen images.}
\label{fig_vision_tasks}
\end{figure}

As perception modules in autonomous driving systems have evolved, there's been a noteworthy pivot from reliance on sophisticated sensor hardware and prebuilt maps to the employment of deep learning algorithms for intricate environmental representation \cite{Hercules, Autoware, CenterLineDet}. Within this paradigm shift, vision-based 3D perception systems, which exclusively utilize camera imagery, have come to the forefront as both a fervent research topic and a practical avenue for reducing hardware deployment costs. Opting for camera-only setups minimizes the need for substantial alterations to existing vehicular architectures compared to multi-modal sensor configurations \cite{TeslaAI}.

In lieu of depth data from additional sensors like LiDAR, vision-based 3D perception systems typically employ multi-view triangulation techniques with a stereo setting or deep learning methods to extrapolate depth information from captured images \cite{Liu2023FSNetTASE, tom2019howdepth, qin2018vins}. These systems then bear the responsibility for accurately identifying and predicting the dynamics of objects and environmental elements in the vicinity of the vehicle.

This thesis concentrates on the crucial aspects of vision-based perception: 3D object detection and depth prediction for the three-dimensional understanding of the surrounding environment. The ability to detect dynamic objects such as vehicles, pedestrians, and other road users forms the basis for subsequent interactive planning. Depth prediction is key to recognizing map elements and detecting general objects in three dimensions. The deep technical challenge of these two tasks is how we obtain depth geometrically or learn from data. These raise several challenges for both research and practical deployment, as outlined below:

\begin{itemize}
    \item{\textbf{Accuracy.} The absence of direct depth measurements from additional sensors like LiDAR makes even well-established multi-view triangulation techniques less reliable in the highly dynamic contexts of autonomous driving. Furthermore, deep learning models tasked with this kind of perception need to concurrently achieve semantic reasoning and geometric interpretation, which inherently places constraints on the attainable accuracy of the system.}
    \item{\textbf{Real-time Deployment.}  While many existing solutions incorporate intricate architectures to integrate multi-view triangulation or other geometric insights within deep neural networks, the complexity of these models must be curtailed to ensure their viability on resource-constrained mobile platforms.}
    \item{\textbf{Data Utilization.} The majority of current deep learning frameworks rely heavily on datasets that include both LiDAR and camera inputs or are annotated by LiDAR. This dependency poses a challenge for training or fine-tuning models on data sourced solely from camera-equipped platforms. If such pre-trained models do not generalize well to novel environments, their real-world applicability is compromised.}
\end{itemize}

\section{Thesis Contribution and Outline}

\begin{figure}
\centering
\includegraphics[width=0.9\columnwidth]{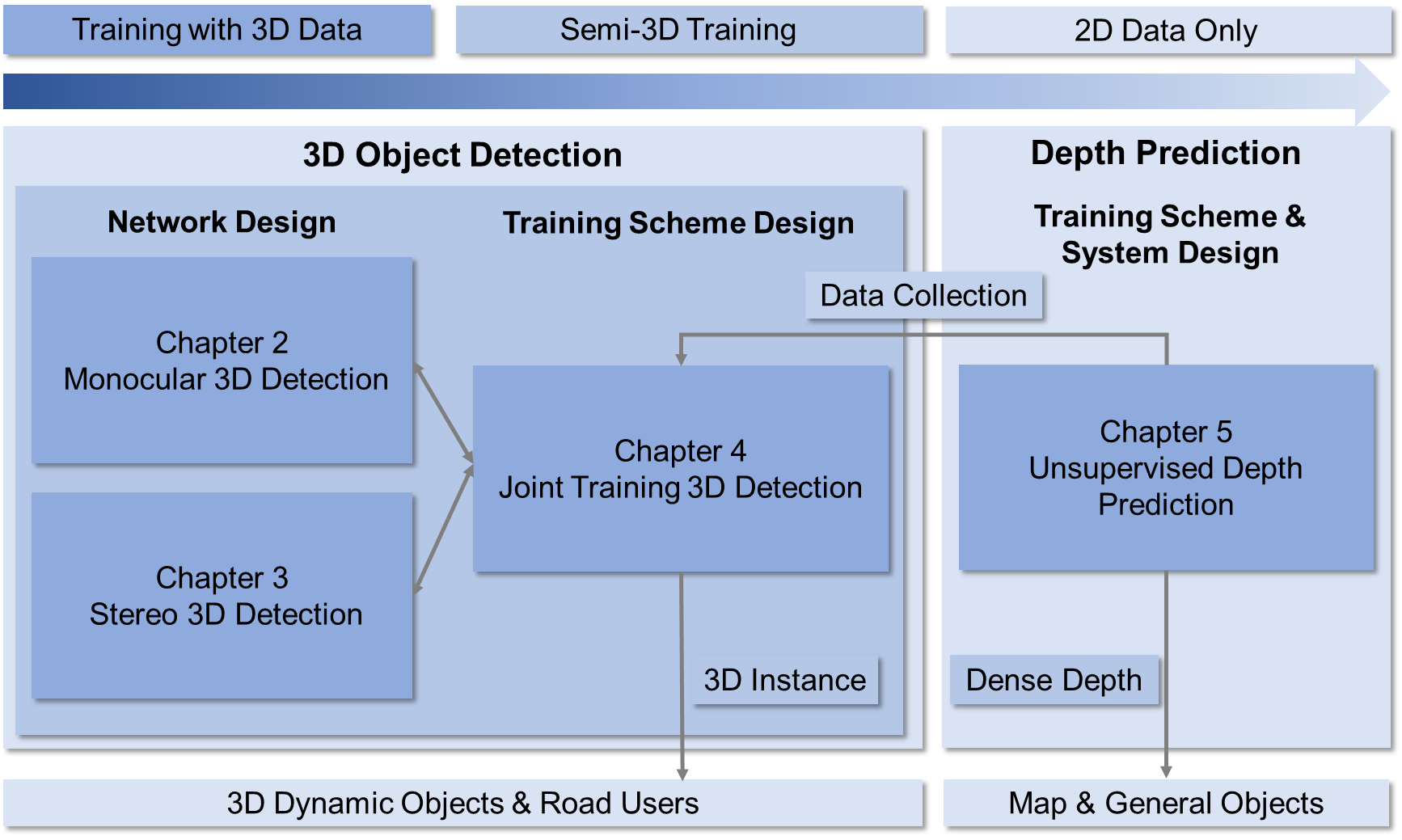}
\caption{This thesis is systematically divided into two principal segments. The first segment concentrates on refining the architecture and training methodologies of 3D object detection networks. The latter segment is devoted to the development of a comprehensive framework for full-scale unsupervised depth prediction. Collectively, these two segments contribute to a nuanced 3D understanding of both dynamic and static elements within the surrounding environment.}
\label{fig_thesie_struct}
\end{figure}

This thesis represents a significant stride in enhancing vision-based 3D perception modules within autonomous driving systems, as illustrated in figure~\ref{fig_thesie_struct}. The research is bifurcated into two primary segments: the first focuses on the development and optimization of vision-based 3D object detection algorithms, while the second is devoted to the application of monocular depth prediction for comprehensive occupancy estimation.

This research commences with the introduction of novel network module designs specifically tailored for monocular 3D detection networks. A key innovation in these designs is the incorporation of additional depth cues, which are intended to significantly enhance the accuracy of object detection.  Building upon these insights, the study further extends its scope to the design of network structures for stereo 3D detection. These structures are meticulously crafted to retain the essential multi-view geometry inherent to stereo imagery also the efficiency of monocular networks. Following this, a novel training methodology for monocular 3D detectors is unveiled. This methodology facilitates training on combined datasets, fostering large-scale training processes and enabling cost-effective knowledge distillation in practical scenarios. As a result of these advancements, the refined networks demonstrate enhanced capability in detecting and tracking critical road users and objects with improved confidence and efficiency in a three-dimensional space. 

In its final phase, the thesis proposes a pioneering approach to unsupervised monocular depth prediction. This approach establishes a comprehensive pipeline for understanding the 3D environment, dramatically reducing the need for label-intensive methods. This unsupervised depth prediction framework addresses the limitations of prior 3D detection networks, particularly their inability to recognize map elements and rare, unlabelled obstacles in three dimensions. Together, these two segments of the thesis coalesce to form the foundational pillars of a fully realized vision-based 3D perception system.

The thesis is structured as follows:

\begin{itemize}
    \item In Chapter 1, an introductory overview of the modular architectures in autonomous driving systems is presented, followed by a survey of challenges and evolutions in vision-based 3D perception. This chapter also outlines the thesis contributions.
    \item In Chapter 2, the intricacies of monocular 3D object detection are investigated. A ground-aware detection algorithm that significantly improves upon existing state-of-the-art methodologies on detection accuracy was proposed. The work was published as \cite{liu2021GAC}.
    \item In Chapter 3, building on insights garnered in Chapter 2, the computational efficiency of stereo 3D object detection frameworks was refined with a brand-new inference structure. achieving an unparalleled balance between performance and computational demand. The work was published as \cite{Liu2021YOLOStereo3D}.
    \item Chapter 4 addresses the challenges of data annotation and model deployment of vision-based 3D detectors. A versatile training regime is introduced, capable of accommodating datasets with varied labeling strategies and enabling joint training on both 2D and 3D annotated data, leading to the training of models on a broad scale of data and the distillation of 3D knowledge to scenes solely annotated in 2D. The work was published as \cite{ma2023dataset}.
    \item In Chapter 5, the focus shifts to unsupervised monocular depth prediction, where a robust training framework that eliminates the need for labeled data is proposed. This chapter also includes an extension to fisheye cameras and validates the approach through real-world testing without LiDAR. The work was published as \cite{Liu2023FSNetTASE}.
    \item In Chapter 6, the thesis concludes with a summary and an exploration of potential avenues for future research.
\end{itemize}

%% file: chapter/03_mono3d.tex
\input{mono3d_secs/sec-0-introduction}

\input{mono3d_secs/sec-1-related-works}
\input{mono3d_secs/sec-2-methods}

\input{mono3d_secs/sec-3-experiments}

\input{mono3d_secs/sec-4-conclusion}

%% file: mono3d_secs/sec-0-introduction.tex
\section{Introduction}

The endeavor to simultaneously deduce the position, orientation, and dimensions of an object in three dimensions from a single, well-calibrated RGB image within an autonomous driving context is fundamentally a challenging problem. 
Lidar-based methods and stereo-vision-based methods, which respectively obtain depths and distance information from lidar measurements and triangulation, can achieve superior performance \cite{Yun2018Focal}\cite{wang2020PointTrackNet}\cite{chen2019ImageDe}\cite{Li2019Stereo}. Monocular vision systems, in contrast, offer a cost-effective and flexible alternative to the LiDAR-based modalities. Moreover, they display a heightened resilience to variations in extrinsic camera parameters when compared to their stereo-vision counterparts. Given these advantages, the pursuit of 3D object detection leveraging a solitary camera remains an active and vibrant area of research, despite the inherent challenge presented by the absence of direct depth cues.

Recent advancements in monocular 3D object detection primarily leverage geometric constraints between the 3D object and its 2D image projection. Techniques like ShiftRCNN \cite{Li2019ShiftRCNN}, SS3D \cite{Jorgensen2019SS3D}, and RTM3D \cite{Li2020RTM3DRM} have enhanced depth and orientation estimation by solving the Perspective-n-Point problem under noisy conditions.

These methods are extensions of the broader challenge in monocular 6D pose estimation. Monocular 6D pose estimation benchmarks like LINEMOD \cite{hinterstoisser2012linemod} are based on the assumption that the CAD models of the objects of interest are known. However, we do not have access to accurate car models for each vehicle in autonomous driving scenes. This limitation curtails the effectiveness of monocular 3D object detectors in these contexts.

In the realm of autonomous driving and mobile robotics, it is safe to assume that most of the important dynamic objects are on a ground plane, and the camera is mounted at a certain height above the ground. Some traditional depth prediction methods also note the importance of ground planes and introduce a similar "floor-wall" assumption for indoor environments \cite{Delage2006FloorWall}, \cite{Chun2013FloorDetectionDepthEstimation}.
Such perspective priors on the ground plane, not presented in \textit{general} monocular 6D pose estimation problems, and provide a significant amount of information for geometric reasoning for monocular 3D object detection in driving scenes.  
 Few recent works \textit{explicitly} inject the perspective priors on the ground plane into a neural network.

This chapter introduces two novel procedures to explicitly enable a monocular object detector to utilize ground plane reasoning.

The first procedure is anchor filtering, where the invariance was explicitly broken in the neural network predictions. Given a prior distance between an anchor and its distance to the camera, we back-project the anchor to 3D. Since all objects of interest are located around the ground plane, 3D anchors far from the ground plane were filtered out during training and testing. This operation focuses the network on positions where objects of interest are likely to appear. We will further introduce this procedure in Section~\ref{sec:anchor_method}.

\input{mono3d_secs/pics/motivation.tex}

The second procedure is a ground-aware convolution module. The motivation of this module is illustrated by Figure~\ref{fig:motivation}. For a human, ground pixels around a car are useful to estimate the car's 3D position, orientation, and dimensions. For an anchor-based detector, features at the center are responsible for estimating all the car's 3D parameters.  However, to infer depth with ground pixels like a human, the network model needs to perform the following steps from the center of an object (e.g. the red dot in the figure) \begin{enumerate}
    \item identifying the contact points of the object and the ground plane (e.g. the blue curve beneath the car).
    \item computing the 3D position of the contact points with perspective geometry.
    \item gathering information from these contact points with a receptive field focusing downwards.
\end{enumerate}

A standard object detection or depth prediction network is built to have a uniform receptive field, and neither perspective geometry priors nor camera parameters are provided to the network. Thus, it is non-trivial to train a standard neural network to make inferences like a human.

The ground-aware convolution module is designed to direct the network towards incorporating ground-based reasoning into its inferences. We encode the prior depth value of each pixel point as an additional feature map, encouraging each pixel to incorporate features from pixels below it. The specifics of this module will be detailed in Section~\ref{sec:lookground}.

Incorporating the two proposed procedures into the network, a one-stage framework with explicit ground plane hypothesis usage is proposed. The network is fast thanks to its clean structure and can run at about 20 frames-per-second (FPS) on a modern GPU.

Additionally, the ground-aware convolution module is adapted into a U-Net-based structure for monocular depth prediction, achieving state-of-the-art (SOTA) performance on the KITTI dataset.

The contribution of the chapter is three-fold.
\begin{itemize}
    \item Identifying the benefit of learning from the ground plane priors in urban scenes for 3D reasoning from images.
    \item Introducing a processing method and a ground-aware convolution module in monocular 3D object detection to use the ground plane hypothesis.
    \item Evaluating the proposed module and design methods on the KITTI 3D object detection benchmark and the depth prediction benchmark with competitive results.
\end{itemize}

%% file: mono3d_secs/pics/motivation.tex
\begin{figure}
  \centering
    \includegraphics[width=0.9\linewidth]{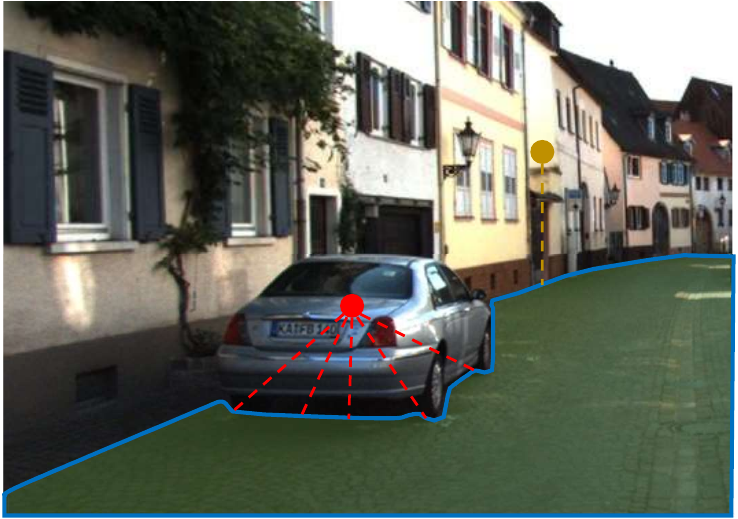}
    \caption{Contact points with the ground plane are important in inferreing 3D information of an object. Predicting depths of background pixels (e.g., the brown point) also rely on the geometry of the ground plane. Best viewed in color.}
    \label{fig:motivation}
  \end{figure}

%% file: mono3d_secs/sec-1-related-works.tex
\section{Related Works}
\label{section:Relate}

\subsection{Pseudo-LiDAR for Monocular 3D Object Detection}
The idea of pseudo-LiDAR, reconstructing point clouds from mono or stereo images, has led to the recent advances in 3D detection \cite{wang2018pseudo}\cite{Ma2019AM3D}\cite{Vianney2019RefinedMPL}\cite{Weng2019Plidar}\cite{Ku2019MonoPSR}.
Pseudo-LiDAR methods usually reconstruct the point cloud from a single RGB image with off-the-shelf depth prediction networks, which limit their performance.
Moreover, the current SOTA monocular depth prediction networks generally take about 0.05s per frame, which significantly limits the inference speed of pseudo-lidar detection pipelines.

\subsection{One-Stage Detection for Monocular 3D Object Detection}
Several recent advances in monocular 3D object detection directly regress 3D bounding boxes in a one-stage object detection framework. 

\textit{Optimization-based Methods:}
SS3D \cite{Jorgensen2019SS3D} concurrently estimated 2D bounding boxes, depth, orientation, dimensions, and 3D corners. Nonlinear optimization was applied to merge all these predictions.
Shift-RCNN\cite{Li2019ShiftRCNN} also estimated 3D information in a 2D anchor and applied a small sub-network instead of a nonlinear solver.
More recent methods, SMOKE\cite{liu2020SMOKE} and RTM3D \cite{Li2020RTM3DRM} incoperate the aforementioned optimization scheme into the anchor-free object detector CenterNet \cite{zhou2019objects}.

\textit{3D Anchor-based Methods:}
M3D-RPN \cite{Brazil2019M3DRPN} introduced 3D priors in 2D anchors, and also emphasized the importance of the ground plane hypothesis.
It also introduced height-wise convolution while D4LCN \cite{Ding2019D4LCN} introduced depth-guided convolution.
Both techniques came at a high cost to efficiency and only utilized the ground plane hypothesis implicitly.

We point out that anchor-based methods are still better than anchor-free methods in 3D detection. Anchor-free detectors implicitly require the network to learn the correlation between the object's apparent size and its distance value. In contrast, anchor-based detectors can embed this in an anchor's preprocessing. As a result, we develop our framework upon anchor-based detectors.

To our knowledge, our proposed framework is the first 3D anchor-based method to explicitly utilize the ground plane hypothesis of driving scenes in monocular 3D detection and achieves the SOTA performance at the time of writing.

\subsection{Supervised Monocular Depth Prediction With Deep Learning}

Supervised monocular depth prediction is another hot research topic closely related to monocular 3D object detection. 

DORN \cite{Fu2018DORN} and SoftDorn \cite{diaz2019SoftDorn} proposed to treat the depth estimation problem as an ordinal regression problem to improve the convergence rate. BTS \cite{Lee2019BTS} proposed the local planar guidance module and incorporated normal information to constraint the depth prediction results in the scenes. BANet \cite{Aich2020BANet}, meanwhile, proposed a bidirectional attention network to improve the receptive fields and global information understanding of depth prediction networks. 

Many of the methods above focus on depth prediction for multiple datasets and scenarios. Images in datasets like NYUv2\cite{SilbermanECCV12NYU} and DIODE\cite{diode_dataset} are taken from various viewpoints, and it is hard to extract floor priors, unlike the cases in driving scenes. As a result, the neural networks mentioned above do not utilize the camera's extrinsic parameters to extract environment priors, and the absolute scale is lacking during the network inference process.

%% file: mono3d_secs/sec-2-methods.tex
\section{Methods}
\label{section:Methods}

This section elaborates on the methods applied in this chapter. First, it presents the formulation of the detection network's inference results and the data preprocessing procedure. 
Second, the ground-aware convolution module that extracts depth priors from the ground plane hypothesis is introduced.
Finally, we present the network's architecture with other major modifications in the training and inferencing process.

\subsection{Anchors Preprocessing}
\label{sec:anchor_method}
\subsubsection{Anchors Definition}

We follow the idea from YOLO \cite{yolov3} to densely predict bounding boxes with dense anchors.
Each anchor on the image also acts as a proposal of an object in 3D. A 3D anchor consists of a 2D bounding box
parameterized by $[x, y, w_{2d}, h_{2d}]$, where $(x, y)$ is the center of the 2D box and $(w_{2d}, h_{2d})$ is the width and height;
3D centers of an object are presented as $[cx, cy, z]$, where $(cx, cy)$ is the center of the object projected on the
image plane and $z$ is the depth; $[w_{3d}, h_{3d}, l_{3d}]$ corresponds to the width, height and length of the 3D bounding box, and $[sin(\alpha), cos(\alpha)]$ is the sine and cosine value of the observation angle $\alpha$.

\input{mono3d_secs/network.tex}

\subsubsection{Priors Extraction from Anchors}
The shape and size of an anchor or an object are highly correlated with the depth. In some prior methods \cite{Brazil2019M3DRPN}, the mean of the depth is computed for each pre-defined anchor box, while variance is computed globally instead. The global variance is only computed to normalize the targets for the neural net.

We further observe that the variance of the depth $z$ of an anchor is inversely proportional to the object's size in the image. Thus, we consider each anchor as a distribution with individual mean and variance of the object proposal in 3D. To collect prior statistical knowledge in the anchors, we iterate through the training set and collect all objects sharing a large intersection-over-union (IoU) with the box for each anchor box with a different shape.
Then we calculate the mean and variance of the depth $z$, $sin(\alpha)$ and $cos(\alpha)$ for each pre-defined anchor box. 
We can significantly lower the prior variance of the depth $z$ for large anchor boxes / close objects.

Since we have considered anchors as distributions of 3D proposals, the associated 3D targets should not deviate much from the expectation. We utilize the fact that most objects of interest should be on the ground plane. Each anchor, centering at $(u, v)$ with pre-computed mean depth $\hat z$, can be back-projected to 3D:
\begin{equation}
    x_{3d} = \frac{u - c_x}{f_x} \hat z \;\;\;\;\;\;\;\;\; y_{3d} = \frac{v - c_y}{f_y} \hat z,
\end{equation}
where $(c_x, c_y)$ is the camera's principal point and $(f_x, f_y)$ is the camera's focal length. Anchors with $y_{3d}$ too far from the ground will be filtered out from training and testing. Such a strategy allows the network to train with 3D anchors around the region of interest and simplify the classification problem.

\subsection{Ground-Aware Convolution Module}
\label{sec:lookground}
Ground-aware convolution is designed to guide the object center to extract features and reason the depth from its contact point; the structure is presented in Figure~\ref{fig:LOOKGROUND}.

To first inject perspective geometry into the network, we encode the prior depth value $z$ of each pixel point, assuming that it is on the ground. The perspective geometry foundation is presented in Figure~\ref{fig:geometric}.

According to the ideal pin-hole camera model, the relation between the depth $z$ and height $y_{3d}$ can be obtained as:
\begin{equation}
    \label{eq:forward}
    z \cdot v = f_y \cdot y_{3d} + c_y \cdot z + T_y,
\end{equation}
where $f_y, c_y,$ and $T_y$ are focal lengths, the principal point coordinate and relative translation respectively, and $v$ is the pixel's y-coordinate in the image. 
\input{mono3d_secs/pics/geometric.tex}

 Assume we know the expected elevation $EL$ of the camera from the ground (1.65 meters in the KITTI dataset \cite{Geiger2012KITTI}). The distance from the ground plane pixel to the camera in $z$ can be solved from Equation \ref{eq:forward}
 as:
\begin{equation}
    \label{eq:solve_z}
    z = \frac{f_y \cdot EL + T_y}{v - c_y}.
\end{equation}

We note that the function is not continuous around the vanishing line of the ground plane ($v=c_y$), and, as indicated in Figure~\ref{fig:geometric}, physically unachievable for $v < c_y$. To detour from such a problem, we first propose to encode the depth value as the disparity of a virtual stereo setup (baseline $B=0.54 m$, similar to the KITTI stereo setup), and we derive the virtual disparity
\begin{equation}
    d = f_y \cdot B \frac{v-c_y}{f_y \cdot EL + T_y}    
\end{equation}
based on the depth $z$ in Equation \ref{eq:solve_z}. Rectified Linear Unit (ReLU) activation ($max(x, 0)$) is then applied to suppress pixels with disparity smaller than zero, which is physically unachievable for forward-facing cameras. After these two steps, the depth priors of the image becomes spatially continuous and consistent.

Inspired by CoordinateConv \cite{Liu2018CoordConv}, we treat this depth prior as an additional feature map with the same spatial size as the base feature map. Each element in the feature map is now encoded with depth priors assuming it is on the ground.

\input{mono3d_secs/pics/lookground.tex}
As motivated in Figure~\ref{fig:motivation}, pixels at the center of the object need to query the depth and image features from contact points, which are usually below the object centers.

Each point $p_i$ in the feature map will then dynamically predict an offset $\delta_{yi}$ as if it is the center of a foreground object
$$
\delta_{yi} = \delta^0_{yi} + \Delta_i = \frac{\hat h}{2EL - \hat h} \cdot (v - c_y) + \Delta_i,
$$, where $\hat h$ is the height of the object (we fix this to be the average height of foreground objects of the dataset), $\Delta_i$ is the residual predicted by the convolution networks. 

Then, as shown in Figure~\ref{fig:LOOKGROUND}, we extract features $f_i'$ at position $p_i + \delta_{yi}$ using linear interpolation. The extracted features $f_i'$ are merged back to the original point $p_i$ with a residual connection.

The ground-aware convolution module mimics how humans utilize the ground plane in depth perception. It extracts geometric priors and features from pixels beneath. The other part of the network is then responsible for predicting the depth residual between the priors and the targets. The module is differentiable and trained end-to-end with the entire network.

\begin{table*}[h]
    \centering
    \footnotesize
    \caption{3D Object Detection Results of Car on KITTI Test Set}
    \def\arraystretch{1.4}
    \setlength\tabcolsep{2pt}
    \begin{tabular*}{1.0\textwidth}{ l|c|c|c|c|c|c|r}
        \toprule
        {\bf Methods} & {\bf 3D Easy} & {\bf 3D Moderate} & {\bf 3D Hard} & {\bf BEV Easy} & {\bf BEV Moderate} & {\bf BEV Hard} & {\bf Time}  \\ 
        \midrule
        MonoPSR\cite{Ku2019MonoPSR}      & 10.76 \% & 7.25 \% & 5.85 \% & 18.33 \% & 12.58 \% & 9.91 \% & 0.2s\\
        PLiDAR\cite{Weng2019Plidar}       & 10.76 \% & 7.50 \% & 6.10 \% & 21.27 \% & 13.92 \% & 11.25 \% & 0.1s\\
        SS3D\cite{Jorgensen2019SS3D}          & 10.78 \% & 7.68 \% & 6.51 \% & 16.33 \% & 11.52 \% & 9.93 \% & 0.05s\\
        MonoDIS\cite{Simonelli2019MonoDIS}      & 10.37 \% & 7.94 \% & 6.40 \% & 17.23 \% & 13.19 \% & 11.12 \% & 0.1s\\
        M3D-RPN\cite{Brazil2019M3DRPN}      & 14.76 \% & 9.71 \% & 7.42 \% & 21.02 \% & 13.67 \% & 10.42 \% & 0.16s\\
        RTM3D\cite{Li2020RTM3DRM}        & 14.41 \% & 10.34 \% & 8.77 \% & 19.17 \% & 14.20 \% & 11.99 \% & 0.05s\\
        AM3D\cite{Ma2019AM3D}         & 16.50 \% & 10.74 \% & 9.52 \% & 25.03 \% & 17.32 \% & \textbf{14.91} \% & 0.4s\\
        D4LCN\cite{Ding2019D4LCN}        & 16.65 \% & 11.72 \% & 9.51 \% & 22.51 \% & 16.02 \% & 12.55 \% & 0.2s\\
        \textbf{Ours}  & \textbf{21.65 \%} & \textbf{13.25 \%} & \textbf{9.91 \%} & \textbf{29.81 \%} & \textbf{17.98 \%} & 13.08 \% & \textbf{0.05s}\\
        \bottomrule
    \end{tabular*} \label{tab:test_results}
\end{table*}

\subsection{Network Architecture for Monocular 3D Detection}

The inference structure of the network is presented in Figure~\ref{fig:network}.
We adopt ResNet-101 \cite{He2015Resnet} as the backbone network, and we only take features at scale $1/16$.
The feature map is then fed into the classification branch and regression branch.

The classification branch consists of two convolutional layers, while the regression branch is composed of a ground-aware convolution module followed by a convolutional output layer.

The shape of the output tensor from the classification branch $C$ is $(B, \frac{W}{16}, \frac{H}{16}, K * \#anchors)$, where $K$ represents the number of classes and $\#anchors$ means the number of anchors per pixel.
    The output tensor from the regression branch is $(B, \frac{W}{16}, \frac{H}{16}, 12 * \#anchors)$.
      There are nine parameters for each anchor: four for 2D bounding box estimation, three for object center predictions, three for dimension predictions, and two more for observation angle predictions.

\subsubsection{Loss Functions}

The total loss $\mathcal{L}$ is the aggregation of classification loss for objectness $L_{cls}$, and regression loss for other parameters $L_{reg}$:
$$
    \mathcal{L} = L_{cls} + L_{reg}.
$$
We adopt focal loss \cite{Yun2018Focal}, \cite{Lin2018Focal} for the classification of objectness and cross-entropy loss for the multi-bin classification
of width, height, and length.
Other parameters, $[x_{2d}, y_{2d}, w_{2d},$ $ h_{2d}, cx, cy, z, w_{3d}, h_{3d}, l_{3d}, sin(\alpha), cos(\alpha)]$,
are normalized based on the anchors' prior parameters and optimized through smoothed-L1 loss \cite{Girshick2015Fastrcnn}.

\subsubsection{Post Optimization}
We follow \cite{Brazil2019M3DRPN} to apply hill-climbing algorithms as a post-optimization procedure.
By perturbating the observation angle and depth estimation,
the algorithm incrementally maximizes the IoU between the directly estimated 2D bounding box and the 2D bounding box projected from the 3D bounding box to the image plane.

The original implementation optimizes the depth and observation angle concurrently.
With repeated experiments, we find that optimizing only the observation angle produces even better results in the validation set. Concurrently optimizing two variables could overfit to the sparse 3D-2D constraints and affect the accuracy of the 3D prediction.

\subsection{Network Architecture for Monocular Depth Prediction}
We adopt a U-Net \cite{UNetFB15Ronneberger} structure for supervised dense depth prediction.
We select a pretrained ResNet-34 \cite{He2015Resnet} as the backbone encoder. 

In the decoding phase, the features are bilinearly upsampled, followed by two convolution layers and concat with the skip connections. We add a ground-aware convolution module before the two convolution layers in the decoder.

The depth prediction network densely predicts the logarithm of depth from each image with a $(B,1,H, W)$ tensor $y=\log{z}$. We provide supervision on each output scale $l$. The total loss is the sum of a scale-invariant (SI) loss $\mathcal{L}_{SI}$ \cite{diaz2019SoftDorn} and a smoothness loss $\mathcal{L}_{smooth}$ \cite{monodepth17} with hyperparameter $\alpha$:
\begin{equation}
    \mathcal{L} = \sum_l (\mathcal{L}_{SI} + \alpha \mathcal{L}_{smooth}).   
\end{equation}
SI loss is commonly used to simultaneously minimize the mean-square-error (MSE) and improve global consistency.
Smoothness loss is needed because the supervision from the KITTI dataset \cite{Geiger2012KITTI} is sparse and lacks local consistency.
The SI loss and smoothness loss are computed with the following equations:
\begin{align}
    \mathcal{L}_{SI} &= \frac{1}{n} \sum_i d_i^2 - \frac{\lambda}{n^2} (\sum_i d_i)^2 \\
    \mathcal{L}_{smooth} &= \frac{1}{N} \sum_i |\partial_xz^l_i|e^{-||\partial_x I^l_i||} + |\partial_yz^l_i|e^{-||\partial_y I^l_i||}, 
\end{align}
where $d_i = \log{z_i} - \log{z_i^*}$, $n$ is the number of valid pixels,  $\lambda \in [0,1]$ is a hyperparameter balancing the absolute MSE loss and relative scale loss, $N$ is the number of total pixels, and $\partial_xI$ and $\partial_yI$ are the gradients of the input images.

%% file: mono3d_secs/network.tex
\begin{figure}
    \centering
        \includegraphics[width=1.0\linewidth]{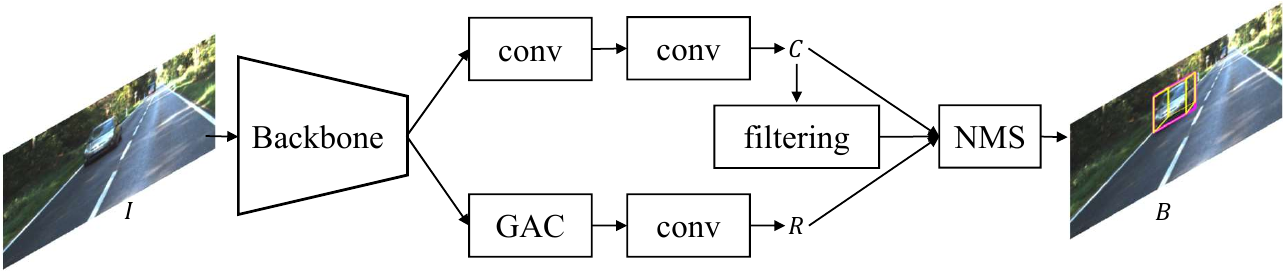}

    \caption{Network structure for 3D object detection. We extract features from image $I$ and predict classfication tensor $C$ and regression tensor $R$. We filter anchors far from the ground before post-processing and produce the final bounding boxes $B$. 
    }
    \label{fig:network}
\end{figure}

%% file: mono3d_secs/pics/geometric.tex
\begin{figure}
    \centering
    \includegraphics[width=0.85\linewidth]{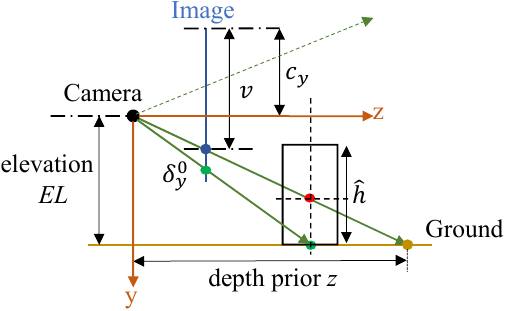}
    \caption{Perspective geometry for the GAC module. When we calculate the vertical offsets $\delta_y^0$, we assume pixels are foreground object centers. When we compute the depth priors $z$, we assume pixels are on the ground because they are features to be queried. }
    \label{fig:geometric}
  \end{figure}

%% file: mono3d_secs/pics/lookground.tex
\begin{figure}
    \centering
        \includegraphics[width=1.0\linewidth]{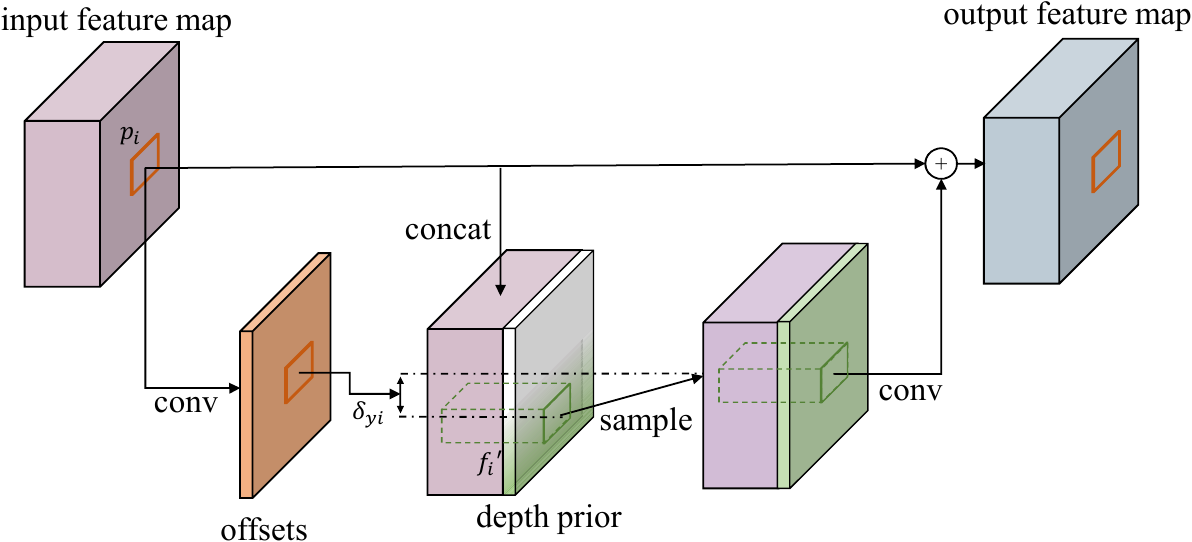}

    \caption{Ground-aware convolution. The network predicts the offsets in the vertical direction, and we sample the corresponding features and depth priors from pixels below. Depth priors are computed with perspective geometry with ground plane assumption.
    }
    \label{fig:LOOKGROUND}

\end{figure}

%% file: mono3d_secs/sec-3-experiments.tex
\section{Experiments}
\label{section:Experiments}

\subsection{Dataset and Training Setups}

We first evaluate the proposed monocular 3D detection network on the KITTI benchmark \cite{Geiger2012KITTI}. The dataset consists of 7,481 training frames and 7,518 test frames.
Chen \textit{et al.} \cite{Chen2015kittisplit} further splits the training set into 3,712 training frames and 3,769 validation frames.

We first determine the hyperparameters of the network with a family of smaller networks fine-tuned on Chen's split \cite{Chen2015kittisplit}.
Then, we retrain the final network on the entire training set with the same hyperparameters before uploading the result for testing on the KITTI server. 
The ablation study that follows is also conducted on the validation set of Chen's split.

Similar to RTM3D \cite{Li2020RTM3DRM}, we double the training set by utilizing images both from the left and right RGB cameras (only RGB images from the left camera are used in validation and final testing) and use random horizontal mirroring as data augmentation (not applied in validation and testing), which significantly enlarges the training set and improve performance. 
The top 100 pixels of each image are cropped to speed up inference, and the cropped input images are scaled to $288 \times 1280$ for the model submitted to the KITTI server, which is similar to the original scale of the images.
The feature map produced by the backbone, therefore, has a shape of $18 \times 80$. Regression loss and classification loss that are too small in magnitude (1e-3) are clipped to prevent overfitting.
The network is trained with a batch size of 8 on a single Nvidia 1080Ti GPU.
During inference, the network is fed one image at a time, and the total average processing time, including file IO and post-optimization, is 0.05s per frame.

\input{mono3d_secs/depth_result_validation.tex}
\input{mono3d_secs/pics/cherry_picking}
\subsection{Evaluation Metric and Results for 3D Detection}

As pointed out by  Simonelli \textit{et al.} \cite{Simonelli2019MonoDIS} and the KITTI team, evaluating performance with 40 recall positions ($AP_{40}$) instead of the 11 recall positions ($AP_{11}$) proposed in the original Pascal VOC benchmark\cite{Everingham10pascal} could eliminate the problematic results presented in the lowest recall bin.
Therefore, we present our results on the test set and also ablation studies based on $AP_{40}$.

The results are presented in Table~\ref{tab:test_results} alongside those of other SOTA monocular 3D detection methods based on the KITTI benchmark.

The proposed network significantly outperforms existing methods on easy and moderate vehicles. 
We do expect ground-aware convolutions to produce more accurate predictions for close-up vehicles with clear borders with the ground plane.

Qualitative results are presented in Figure~\ref{fig:examples}. The model shown here shares the same hyperparameters as the model submitted to the KITTI server but is only trained on the training sub-split. In the images on the left-hand side of the figure, cars are mostly detected and estimated accurately. The effect of the GAC module is also visualized.

We present several typical failure cases on the right-hand side of Figure~\ref{fig:examples}, and in the top-right image, the network does not detect a heavily obscured car.
In the middle-right image, truncated cars and a car that is quite far away are not detected.
We acknowledge that the network could still have trouble detecting small objects.
We show the bottom-right image to demonstrate cases in which the network give an inaccurate estimation of the 3D dimensions of a car because, as stated in Section~\ref{section:Methods}, it is still difficult to estimate the width, length, and height of an object merely by semantic information in the image. 

We provide an ablation study of the model in Section~\ref{section:Discussion}.

\subsection{Experiments on Monocular Depth Prediction}

We further evaluate the proposed depth prediction network in the KITTI depth prediction benchmark \cite{Geiger2012KITTI}.  The dataset for monocular depth prediction consists of 42949 training frames, 1000 validation samples, and 500 test samples, annotated with sparse point clouds.

The input images are cropped to $352 \times 1216$ during training and testing.  In the loss function, we applied $\alpha$= 0.3, $\lambda$=0.3 through grid-search on the validation set. The network is also trained with a batch size of 8 on a single Nvidia 1080Ti GPU. 

Scale-invariant log error (SILog) is the primary metric used in the KITTI benchmark to evaluate depth prediction algorithms. 

The results are presented in Table~\ref{tab:depth_validation_results}. The proposed network produces one of the best performances on the KITTI dataset, providing competitive results compared with SOTA methods. We also show that the network improves significantly against the baseline U-Net Model.

Qualitative results are presented in Figure~\ref{fig:depth_examples}. Depth predictions inside the range of LiDAR are generally consistent. Depth predictions along long, vertical objects like trees are consistent thanks to the ground aware convolution module. We point out that there are still artifacts around the edge of objects and areas without supervision because the network receives no post-processing and little pre-training.
The depth prediction results show that the proposed module and the proposed network can improve depth inferencing from monocular images in autonomous driving scenes.

\input{mono3d_secs/cherry_picking_depth.tex}
\section{Model Analysis and Discussion}
\label{section:Discussion}
In this section, we further analyze the performance of the proposed method and discuss the effectiveness of each design choice. The experiments will focus more on monocular 3D detection.
We conduct ablation studies to validate the contribution of anchor preprocessing and the ground-aware convolution module.

\begin{table*}
   \caption{3D Detection Ablation Study Results of Car on KITTI Validation Set}
   \centering
   \footnotesize
   \def\arraystretch{1.4}
   \setlength\tabcolsep{2pt}
   \begin{tabular*}{1.0\textwidth}{  @{\extracolsep{\fill}} l| c|c }
      \toprule
       {\bf Methods} & {\bf $IoU\ge 0.7$ 3D Easy/Moderate/Hard } &  {\bf $IoU\ge 0.5$ 3D Easy/Moderate/Hard}  \\ \midrule
       \textbf{Baseline Model}  & \textbf{23.63 \%}/  \textbf{16.16 \%}/   \textbf{12.06} \% & \textbf{60.92 \%}/   \textbf{42.18} \%/  \textbf{32.02 \%}  \\
       \hline
    w/o Anchor Filtering & 21.39 \%/   14.35 \%/ 11.11  \% & 59.76 \%/    41.00 \%/   31.10 \% \\
    w OHEM & 22.45 \%/   15.10 \%/ 11.29  \% & 60.71 \%/    42.01 \%/   31.88 \% \\
    \hline
    w Conv                 & 21.57 \%/   15.26 \%/ 11.35  \% & 58.17 \%/    41.17 \%/   32.58 \% \\
    w DisparityConv       & 22.13 \%/   15.42 \%/ 11.34  \% & 60.13 \%/    41.62 \%/   33.07 \% \\
    w Deformable Conv      & 22.16 \%/   15.71 \%/ 11.75  \% & 62.24 \%/    43.93 \%/   33.76 \% \\

       \bottomrule
   \end{tabular*} \label{tab:mono3d_ablation_study}
\end{table*}
\subsection{Anchor Preprocessing}
\label{subsection:ablation_data}
We first conduct experiments on anchor filtering. In the experiment, we do not filter out unnecessary anchors during training and testing. We notice that the proposed filtering will filter out half of the negative anchors, so we also conduct an experiment against Online Hard Example Mining (OHEM), where we filter out half of the easy negative anchors during training\cite{OHEM2016Shrivastava}.

As shown in Table~\ref{tab:mono3d_ablation_study}, the baseline model outperforms the ablated one and OHEM.  The baseline model performs better at 3D inference. We also point out that there is almost no difference in 2D detection between the two models.

Generally, a one-stage single-scale object detector not only needs to classify background from the foreground but also needs to select anchors with the correct scales at foreground pixels, which also means selecting the proper depth prior. Filtering off-the-ground anchors during training and testing significantly lowers the learning burden for the classification branch of the object detector. Thus the classification branch can focus more on selecting the right anchors for foreground pixels. Such a method, as a result, also outperforms position-invariant filtering methods like OHEM.

\subsection{Ground-Aware Convolution Module}
Intuitively, basic convolutions provide a uniform receptive field for each pixel, and the network could implicitly learn to adjust its receptive field by fine-tuning the weights of multiple convolution layers. Deformable convolutions \cite{Zhu2018Deformv2} further explicitly encourage the network to adapt its receptive field according to each pixel's surrounding context. Compared with deformable convolutions, the ground-aware convolution module fixes the search direction and allows a larger search range.

We substitute the proposed module with basic convolutions, disparity-conditioned convolutions (i.e., convolution with the depth prior as an additional feature map), and deformable convolutions to examine the performance.
The results are shown in Table~\ref{tab:mono3d_ablation_study}. The experiments with deformable convolutions demonstrate better 2D detection results. 

Deformable convolution can enhance the performance with a generally larger receptive field. While disparity-conditioned convolution provides the network with prior depths, the receptive field of the network is lacking. These two modules improve the performance, but the proposed module has better results by a considerable margin.

%% file: mono3d_secs/depth_result_validation.tex
\begin{table*}
    \centering
    \caption{Depth Prediction Results on KITTI Test Set}
    
    \begin{tabular*}{0.8\textwidth}{ l|c|c|c|c }
        \toprule
        {\bf Methods} & {\bf SILog} & {\bf sqErrorRel} & {\bf absErrorRel} & {\bf iRMSE}  \\ 
        \midrule
         
        PAP\cite{PAP2019zhang}          & 13.08 & 2.72 \% & 10.27 \% & 13.95 \\
        VNL\cite{Yin2019VNLNet}         & 12.65 & \textbf{2.46} \% & 10.15 \% & 13.02  \\
        SoftDorn\cite{diaz2019SoftDorn} & 12.39 & 2.49 \% & 10.10 \% & 13.48  \\
        \midrule
        Base U-Net             & 12.78 & 3.11 \% & 10.12 \% & 13.46 \\
        \textbf{Ours}           & \textbf{12.13} & 2.61 \% & \textbf{9.41} \% & \textbf{12.65} \\
        \bottomrule
    \end{tabular*} \label{tab:depth_validation_results}
\end{table*}

%% file: mono3d_secs/pics/cherry_picking.tex
\begin{figure*}
    \centering
    \includegraphics[{width=1.0\textwidth}]{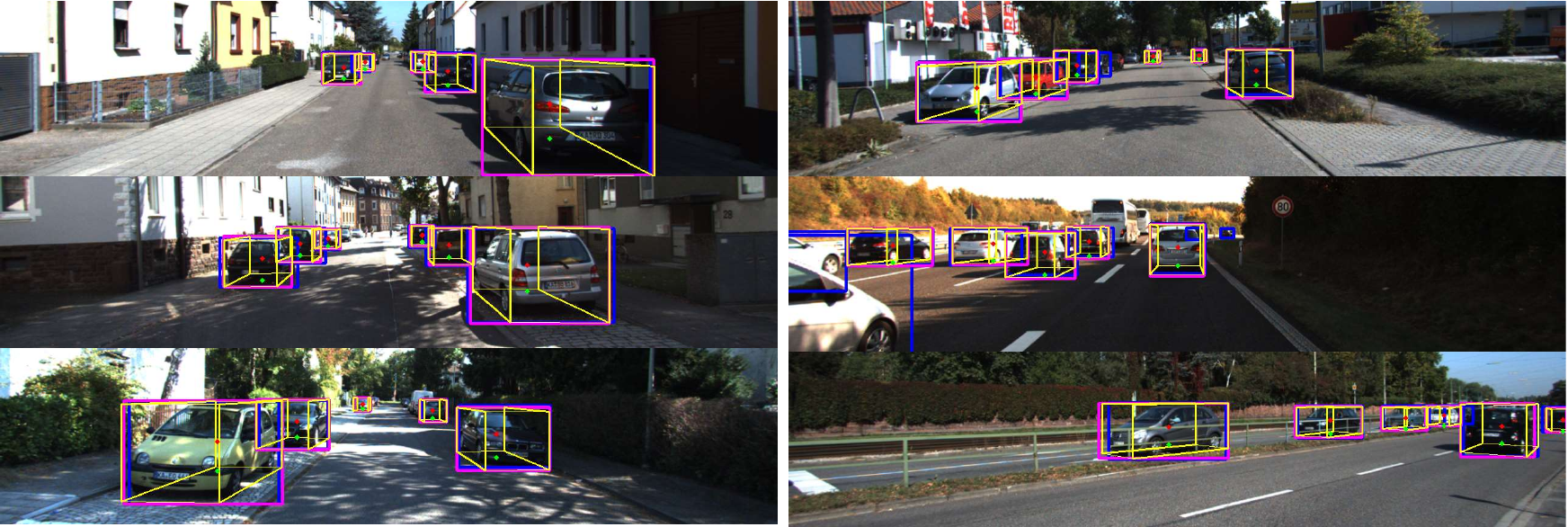}
    \caption{Qualitative examples from validation sets. Blue boxes, pink boxes and yellow boxes indicate the ground truth
    2D bounding box, estimated 2D bounding box, and estimated 3D bounding box respectively. Red points are object centers and green points visualize the offsets $\delta_{yi}$ in the GAC module.
    }

    \label{fig:examples}
\end{figure*}

%% file: mono3d_secs/cherry_picking_depth.tex
\begin{figure*}
    \centering
    \includegraphics[{width=0.9\textwidth}]{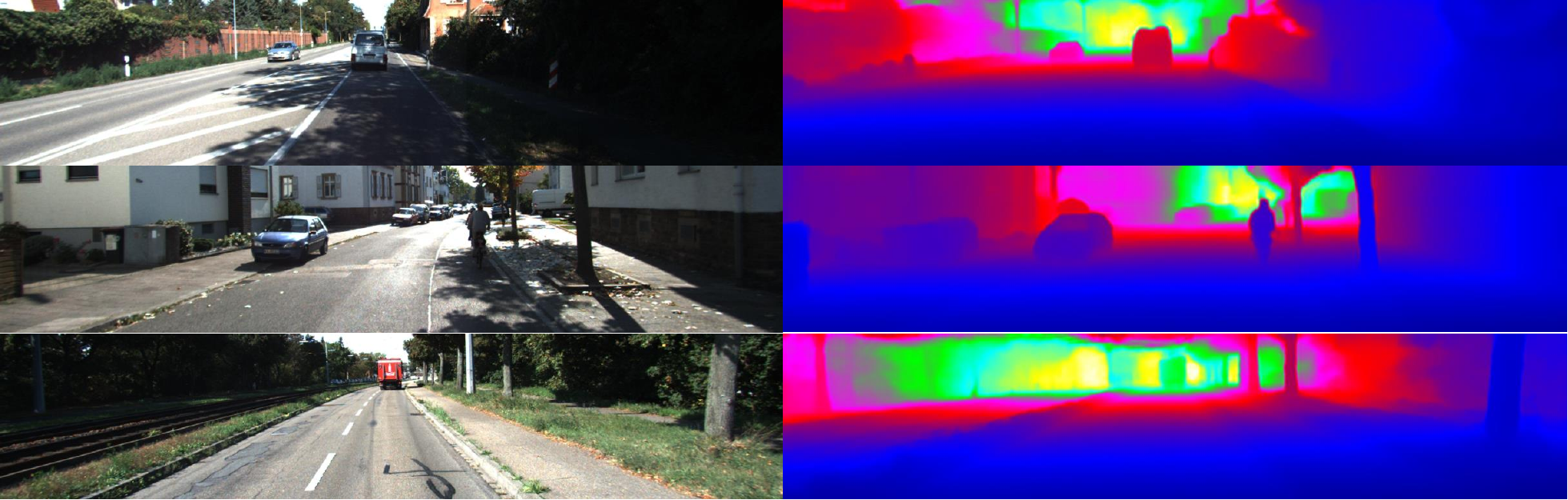}

    \caption{Qualitative examples of depth prediction from validation sets. The depth maps on the right are rendered with the official color map.
    }

    \label{fig:depth_examples}
\end{figure*}

%% file: mono3d_secs/sec-4-conclusion.tex
\section{Conclusion}
\label{section:Conclusion}

In this chapter, we have detailed the development of a ground-aware monocular 3D object detection framework tailored for autonomous driving scenarios. This framework innovatively combines statistical and geometric priors with data-driven approaches to refine the problem of monocular 3D detection.

We introduced two key advancements: an anchor filtering procedure and a ground-aware convolution module. The anchor filtering procedure infuses ground plane priors and statistical priors into anchor placements, refining the focus of the detection process. The ground-aware convolution module, on the other hand, equips the network with critical geometric priors and contextual hints, enabling it to effectively reason based on ground plane information.

The efficacy of the proposed monocular 3D object detection network is demonstrated through its performance on the KITTI detection benchmark, where it achieved state-of-the-art (SOTA) results among monocular methods. Additionally, the application of the ground-aware convolution module in the monocular depth prediction task yielded competitive outcomes on the KITTI depth prediction benchmark.

It is important to note, however, that the "floor-wall" assumption integral to our approach is primarily applicable to scenes with specific camera poses and is only partially valid in the multifaceted environment of driving scenes. Our methods do not explicitly discern the boundaries between the ground and other objects; rather, they are designed to embed significant information and priors into the network, relying on a data-driven methodology.

In conclusion, while acknowledging the limitations inherent in our assumptions and approach, the methods proposed in this chapter significantly advance the field of 3D detection and depth inference from images. They offer robust and powerful tools for enhancing neural network models in the domains of autonomous driving and mobile robotics.

%% file: chapter/04_stereo3d.tex
\input{stereo3d_secs/sec-0-introduction}

\input{stereo3d_secs/sec-1-related-works}

\input{stereo3d_secs/sec-2-methods}

\input{stereo3d_secs/sec-3-experiments}

\input{stereo3d_secs/sec-4-conclusion}

%% file: stereo3d_secs/sec-0-introduction.tex
\section{Introduction}
\label{section:Introduction}
3D object detection remains a cornerstone challenge in computer vision, with pivotal applications in autonomous vehicles and mobile robotics. The use of a binocular setup, involving two horizontally aligned RGB cameras with a known displacement, enables depth estimation through triangulation based on the pin-hole camera model, making depth inferencing much more accurate compared with the monocular setting. Although less robust compared to LiDAR-based methods, this stereo vision approach offers a cost-effective solution for low-budget applications such as mobile robots and autonomous logistic vehicles.
\input{stereo3d_secs/network.tex}

Many of the state-of-the-art frameworks in stereo 3D object detection stem from the idea of pseudo-LiDAR and are motivated by general stereo-matching algorithms.
However, in 3D object detection, the model should focus on foreground objects. It is expected to be as accurate as possible since a disparity error of one or two pixels would cause a large error in terms of real-world distance.
Many researchers have delved deep into these problems to improve the performance of pseudo-LiDAR-based algorithms; some directly fine-tune the estimation of point clouds to improve performance \cite{You2019PLPP},\cite{Qian2020PLE2E}, while others utilize instance segmentation to focus the stereo matching network on foreground pixels \cite{Sun2020DispRCNN}, \cite{xu2020Zoomnet}, \cite{Pon2019OCStereo}.
However, a high-performance disparity estimation network, e.g., PSMNet \cite{Chang2018PSMNet}, usually takes more than 300 ms per frame on modern hardware on the KITTI dataset \cite{Geiger2012KITTI} and requires a huge GPU memory to train.
These issues hinder the deployment of stereo systems on low-cost robotic applications.

Many of the works mentioned above have shown in practice that transforming images into 3D features is usually sub-optimal and computationally expensive.
To improve the efficiency of stereo 3D detection algorithms while maintaining as much of their performance as possible, we propose selecting a different architecture.
Instead of casting the problem as a 3D detection problem with less accurate point clouds, we take a step back and treat it as a monocular 3D detection task with enhanced stereo features, which is the fundamental motivation of this work. 

The framework, YOLOStereo3D, is a light-weight one-stage stereo 3D detection network (Section~\ref{sec.mono_anchor}).
To efficiently produce powerful stereo features, we re-introduce the pixel-wise correlation module to construct the cost-volume, instead of the popular concatenation-based module (Section~\ref{sec.costvolume}). 
Such a module produces a thin 2D feature map where each channel corresponds to a disparity hypothesis in stereo matching.
Then this module was applied hierarchically to efficiently produce stereo features as 2D feature maps (Section~\ref{sec.ghost}), and these features are densely fused (Section~\ref{sec.hierachical}) to form the base-feature of detection heads. The network is trained end-to-end without the use of LiDAR data (Section~\ref{sec.training}).

The main contributions of this chapter are three-fold. 
\begin{itemize}
    \item For the inference architecture, we incorporate and optimize the inference pipeline from one-stage monocular 3D detection into stereo 3D detection.
    \item For the design of the network, a point-wise correlation module in stereo detection tasks is introduced, and a hierarchical, densely-connected structure is proposed to utilize stereo features from multiple scales.
    \item For the experimental results, the proposed YOLOStereo3D produces competitive results on the KITTI 3D benchmark without using point clouds and with an inference time of less than 0.1 seconds per frame.
\end{itemize}

%% file: stereo3d_secs/network.tex
\begin{figure*}[htb]
    \centering
    \includegraphics[{width=0.95\textwidth}]{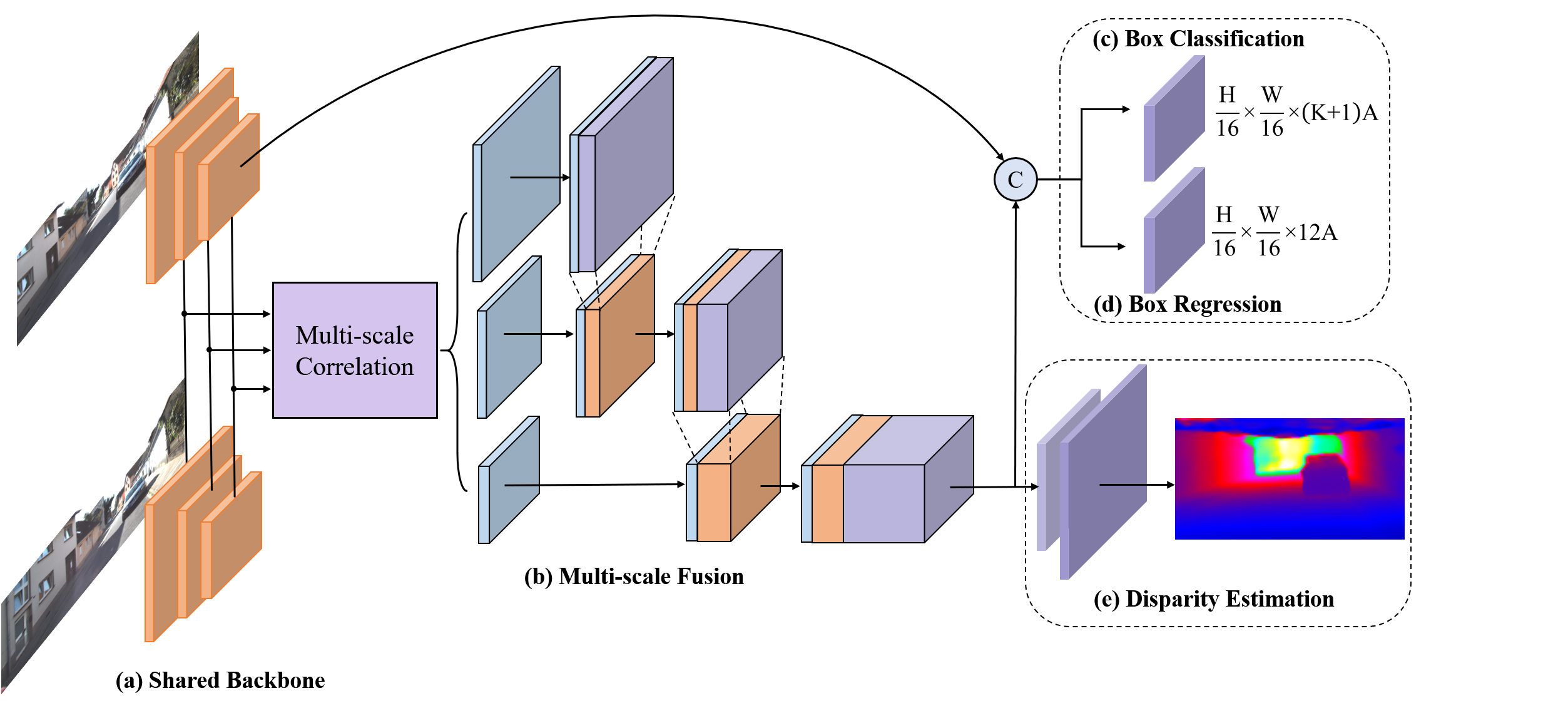}
        
    \caption{Network inference structure of YOLOStereo3D.
    YOLOStereo3D extracts multi-scale features from binocular images with a backbone network (a). 
    These features are passed through a multi-scale stereo matching and fusion module (b) as described in Section~\ref{subsec: multiscale_fusion}.
    Finally, the fused features are concatenated with the last feature from the left image and sent to the classification/regression branch to densely predict the 3D bounding boxes (c/d).
    The network also produces a disparity estimation during training (e).
   }
    \label{fig:architecture}
\end{figure*}

%% file: stereo3d_secs/sec-1-related-works.tex
\section{Related Works}
\label{section:Relate}
\subsection{Stereo Matching}

Stereo matching algorithms focus on estimating the disparity between binocular images.
The current state-of-the-art frameworks for stereo matching apply siamese networks for feature extraction from two images, and construct 3D cost volumes to search the disparity value on each pixel exhaustively.
Early research applied the dot-product between binocular feature maps, with the resulting correlation directly forming an estimation of the disparity distribution \cite{Luo2016EffiStereo} \cite{ZbontarL15DotStereo}.
PSMNet \cite{Chang2018PSMNet} and GCNet \cite{Kendall2017GCNet} constructed concatenation-based cost volumes and applied multiple 3D convolutions to produce disparity outputs.
The recent FADNet managed to perform a fast stereo estimation with a point-wise correlation module \cite{wang2020FADNet}.
Zhang \textit{et al.} proposed stereo focal loss to improve the loss function formulation in disparity estimation \cite{Zhang2019AcfNet}.
Our work, similar to many other stereo 3D object detection algorithms, is developed upon these studies and utilizes the stereo matching features to boost detection performance.

\subsection{Visual 3D Object Detection}

\subsubsection{Stereo 3D Object Detection}

Stereo 3D object detection is usually considered as a tractable but computationally hard problem.
Recent advances in stereo 3D object detection algorithms are based on the idea of pseudo-LiDAR \cite{wang2018pseudo}.
DispRCNN \cite{Sun2020DispRCNN}, ZoomNet \cite{xu2020Zoomnet}, and OC Stereo \cite{Pon2019OCStereo} applied instance segmentation on binocular images to construct a local point cloud for each detected instance to improve the accuracy of disparity estimation on foreground objects.
Pseudo-LiDAR++ \cite{You2019PLPP} recognized that uniform 3D convolution might not be suitable to process the disparity cost volume, and transformation to the depth cost volume may be needed.

We point out that all the aforementioned algorithms require more than 0.3 seconds runtime per frame. Moreover,
Pseudo-LiDAR++ \cite{You2019PLPP}, ZoomNet \cite{xu2020Zoomnet}, OC Stereo \cite{Pon2019OCStereo} and DSGN \cite{Chen2020DSGN} required point cloud data during training or need point cloud data to help the training process.
DispRCNN \cite{Sun2020DispRCNN} and the baseline Pseudo-LiDAR \cite{wang2018pseudo} required off-the-shelf disparity modules, which are usually trained with depth images or point cloud data.

YOLOStereo3D is a light-weight model that performs most of the convolution operation in the perspective view, and the training and inference are significantly lighter and faster than all methods mentioned above.
Moreover, the training process of YOLOStereo3D does not depend on point-cloud data.

\subsubsection{Monocular 3D Object Detection}

Monocular 3D object detection is an ill-posed problem, but it provides many insights into how depth information can be estimated from a single image.
Tom \textit{et al.} \cite{tom2019howdepth} demonstrated that a typical monocular depth estimation network mainly estimates depth from the vertical position of an object.
The authors \cite{tom2019howdepth} provided the theoretical background for pseudo-LiDAR in monocular detection \cite{Vianney2019RefinedMPL}\cite{Weng2019Plidar}.
YOLOStereo3D is built upon the inference structure of M3D-RPN \cite{Brazil2019M3DRPN} and GAC \cite{liu2021GAC} and further enhances the final features with stereo matching results.

%% file: stereo3d_secs/sec-2-methods.tex
\section{Methods}
\input{stereo3d_secs/anchor_filtering.tex}

In this section, we elaborate on the network structure and methods applied in this chapter.
First, we introduce the output definition and data-preprocessing tricks imported into and optimized for YOLOStereo3D \cite{liu2021GAC}.
Second, we re-introduce the light-weight cost volume that speeds up stereo matching and present the hierarchical densely-connected structure that fully exploits such thin features.
Finally, we deliver the loss function as well as the training and inference scheme of YOLOStereo3D.

\subsection{Anchors Definition and Preprocessings}
\label{sec.mono_anchor}
Since we adopt the inference structure of a monocular 3D object detection framework, we need to import the basic definition of anchors and we propose multiple optimized processing methods.
In this subsection, we present some of the preprocessing on the input and output of the network.
\subsubsection{3D Anchors and Statistical Priors}

Each anchor is described by 12 regressed parameters including
$[x_{2d}, y_{2d}, w_{2d}, h_{2d}]$ for the 2D bounding boxes;
$[c_x, c_y, z]$ for the 3D centers of objects on the left image;
$[w_{3d}, h_{3d}, l_{3d}]$ corresponding to the width, height and length of the 3D bounding boxes respectively;
and $[sin(2\alpha), cos(2  \alpha)]$ to estimate the observation angle/orientation of objects. 

We observe that $[sin(\alpha), cos(\alpha)]$  and $[sin(\alpha + \pi), cos(\alpha + \pi)]$ correspond to the same rectangular bounding box results in 3D object detection. As a result, we instead predict $[sin(2\alpha), cos(2\alpha)]$ in the regression branch.
We also add a classification channel to predict if $ |\alpha| > \frac{\pi}{2}$ to eliminate ambiguity, which intuitively means whether or not the object is facing the camera.

We incorporate 3D statistic priors into 2D anchors to improve the regression results.
To collect prior statistics of the anchors, we iterate through the training set, and for each anchor box of different shapes, collect all the objects assigned to this anchor based on the IoU metric.
Then, we compute the mean and the variance of $z$, $sin(2\alpha)$, $cos(2\alpha)$ for each box. 

Furthermore, we explicitly exploit scene-specific knowledge for autonomous vehicles and utilize the statistical information from the anchor boxes.
During training, we project dense anchor boxes into 3D with the mean depth value $\hat z$ and filter out anchor boxes that are far away from the ground plane based on their, as displayed in Figure~\ref{fig:anchor}. 

 For multi-class training, since the statistics for different types of obstacles, e.g., cars and pedestrians, are significantly different, we compute 3D priors for each category, separately.
 During training, we filter out anchor boxes dynamically based on the categories assigned. During inference, we also filter out anchor boxes dynamically based on the anchors' local categorical predictions.

\subsubsection{Data Augmentation for Stereo 3D Detection}

Data augmentation is useful to improve the generalization ability in deep learning applications.
However, the nature of stereo 3D detection limits the number of possible augmentation choices.
We follow \cite{Brazil2019M3DRPN} to apply photometric distortion concurrently on binocular images.
We also follow \cite{Li2019Stereo} to apply random flipping online during training. 
Random flipping includes flipping both RGB images, flipping the position and orientation of objects, and then switching left/right images.

\subsection{Multi-Scale Stereo Feature Extraction}
\label{subsec: multiscale_fusion}
The extraction of stereo features is one of the most time-consuming parts for many pre-existing stereo 3D object detection algorithms.
In this subsection, we re-introduce the cost volume formulation based on dot-product/cosine-similarity and present the hierarchical structure to utilize these features effectively.

\subsubsection{Light-weight Cost Volume}
\label{sec.costvolume}
Current state-of-the-art stereo matching algorithms usually construct 3D cost volume with concatenation, where the module iteratively shifts the right feature map horizontally over the left feature map, and at each step, concatenate the two features at each overlapping pixel. For binocular feature maps with the shape $[B, C, H, W]$, the shape of the output tensor $f_i$ is $[B, 2\cdot C, max\_disp, H, W]$. In this chapter, we follow \cite{Luo2016EffiStereo} and \cite{wang2020FADNet} to apply a normalized dot-product to construct a thin cost volume. Such a module compute \textbf{correlation} between two overlapping pixels of the feature maps instead. The shape of the output tensor $f_i'$ becomes $[B, max\_disp, H, W]$.

The stereo matching process can be much faster. Consider two input feature maps of $[1, 64, 72, 320]$, which is a common shape of a KITTI image down scaled by 4. The forward pass of concatenation-based cost volume construction takes about 200 ms while the correlation-based cost volume takes about 7 ms on an Nvidia-1080Ti.

However, the number of output channels is smaller, which could cause the network to be numerically skewed towards monocular features during the fusion stage and downsampling the stereo matching results could induce further information loss.
We ease these two problems with densely connected ghost modules \cite{Han2020Ghost} and a hierarchical fusion structure.

\subsubsection{Densely Connected Ghost Module}
\label{sec.ghost}
As mentioned in Section~\ref{sec.costvolume}, we need to expand the width of the features to guide the network to skewed towards features produced by stereo matching.

Han \textit{et al.} propose the ghost module, which is an efficient module to produce redundant features \cite{Han2020Ghost}. It applies depthwise convolution to produce extra features, which requires significantly fewer parameters and FLOPs.
We go one step further and densely concatenate the original input features with the output of the original ghost module, thereby tripling the number of channels. As indicated in Figure~\ref{fig:architecture}, the mauve blocks in (b) are the results from ghost module and others denote densly connected residuals. 

Such a module preserves more information before downsampling and also rebalances the number of channels between stereo features and monocular semantic features during the fusion phase. 

\subsubsection{Hierachical Multi-scale Fusion Structure}
\label{sec.hierachical}
To minimize the information loss during the stereo matching phase while keeping the computational time tractable, we engineer a hierarchical fusion scheme.
At the downsampling level of $\frac{1}{4}$ and $\frac{1}{8}$, we construct a light-weight cost volume of a max-disparity of 96 and 192, respectively.
As shown in Figure~\ref{fig:architecture}, they are fed into a densely connected ghost module, downsampled, and concatenated with features at a smaller scale.
At a downsampling level of $\frac{1}{16}$, we first downsample the number of channels with $1\times 1$ convolution.
We then construct a small concatenation-based cost volume (also flattened to be a 2D feature map) to preserve more semantic information from the right images.

This arrangement can also be justified with high-level reasoning.
Features with higher resolution are usually local features with higher frequency portions, which are suitable for dense and accurate disparity estimation. In contrast, features with low resolution contain semantic information at a larger scale. 

\subsection{Training Scheme and Loss Function}
\label{sec.training}
The overall network structure is presented in Figure~\ref{fig:architecture}.
Multi-scale features from binocular images are extracted and fused into stereo features to construct hierarchical cost volumes.
The stereo feature map is concatenated with the last feature map of the left image and fed to the regression/classification branch.
The stereo feature map is also fed into a decoder to predict a disparity map trained with an auxiliary loss. The auxiliary loss can regularize the training process.

\subsubsection{Auxiliary Disparity Supervision in Training}

As pointed out by Chen \textit{et al.} \cite{Chen2020DSGN}, disparity supervision is important to improve detection performance.
We also observe a similar phenomenon in our framework.
Without disparity supervision, the network may not be guided to produce local features useful in stereo matching to fully utilize the geometric potential of binocular images, and the network could be trapped in a local minimum similar to that of a monocular detection network.

We upsample the output of the final stereo features to $[W/4, H/4]$, and supervise the prediction with a sparse "ground truth" disparity derived from the traditional block matching algorithm in OpenCV \cite{opencv_library} during training.
During evaluation and testing, this disparity estimation branch is disabled to improve efficiency.

Though the disparity from the block matching algorithm is coarse and sparse, we empirically show that it significantly improves the network's performance.

\subsubsection{Loss Function}

We apply focal loss \cite{Yun2018Focal}\cite{Lin2018Focal} on classification, and smoothed-L1 loss \cite{Girshick2015Fastrcnn} on bounding box regression.  We follow the scheme of \cite{Zhang2019AcfNet} to apply stereo focal loss on the auxiliary disparity estimation.
First, we compute the expected distribution of disparity with a hard-coded variance $\sigma = 0.5$:
$$
    P(d) =\operatorname{softmax}\left(-\frac{\left|d-d^{g t}\right|}{\sigma}\right) =\frac{\exp \left(-c_{d}^{g t}\right)}{\sum_{d^{\prime}=0}^{D-1} \exp \left(-c_{d^{\prime}}^{g t}\right)}.
$$
Where $d$ represents the disparity and $c_d$ indicates the predicted confidence at disparity $d$. Then, following \cite{Zhang2019AcfNet}, stereo focal loss is defined as:
$$\mathcal{L}_{S F}=\frac{1}{|\mathcal{P}|} \sum_{p \in \mathcal{P}}\left(\sum_{d=0}^{D-1}\left(1-P_{p}(d)\right)^{-\alpha} \cdot\left(-P_{p}(d) \cdot \log \hat{P}_{p}(d)\right)\right)$$
where $D$ is the max-disparity, $\alpha$ is the focus weight, $\mathcal{P}$ presents the set of pixels involved, and $P_p(d)$, $\hat P_p(d)$ represents the expected and predicted distribution map of disaparity $d$.

The final loss function is simply the sum of the three losses.

%% file: stereo3d_secs/anchor_filtering.tex
\begin{figure*}
    \centering
    \includegraphics[{width=1.0\textwidth}]{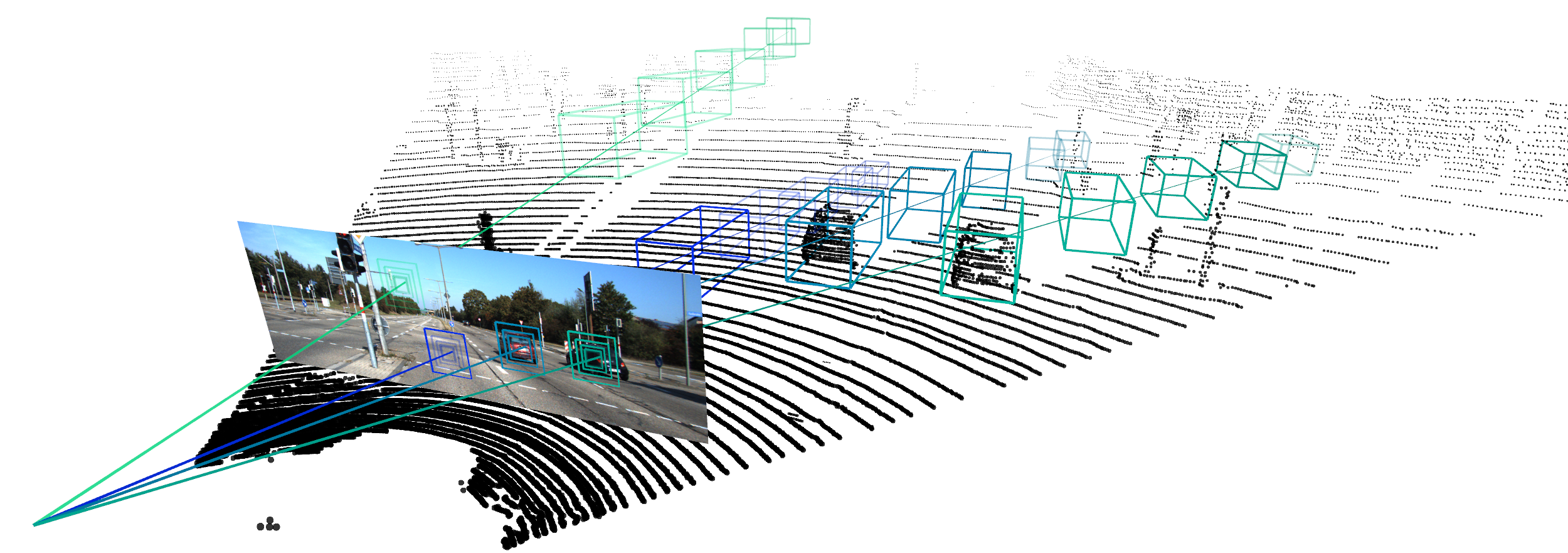}
        
    \caption{We project the center of each anchor box from the left image plane to 3D with its mean distance $\hat z$. We visualize the projected 3D bounding boxes with the mean width/height/length of the cars.
    We filter out anchors that are far from the ground plane during training (transparentized in the figure).
    Point clouds are displayed to indicate 3D positions in the figure. Best viewed in color.
    }
    \label{fig:anchor}
\end{figure*}

%% file: stereo3d_secs/sec-3-experiments.tex
\section{Experiments}
\label{section:Experiments}

We evaluate our method on the KITTI Object Detection Benchmark \cite{Geiger2012KITTI}.
The dataset consists of 7,481 training frames and 7,518 test frames.
Chen \textit{et al.} \cite{Chen2015kittisplit} further split the training set into 3,712 training frames and 3,769 validation frames. 
 In this section, we provide further training details and show the performance of YOLOStereo3D on the test set to compare it with existing models.

 \input{stereo3d_secs/test_result.tex}

 \subsection{Implementation and Training Details}

 Modern deep learning frameworks are sensitive to hyperparameters choices, and critical design choices could profoundly influence the final performance.
 We introduce some crucial design choices before showing the performance, and the code will be made open source upon publication.

 Firstly the structure and the hyperparameters of the network is determined on Chen's split \cite{Chen2015kittisplit}.
 Then, the final network is retrained on the entire training set with the same hyperparameters before uploading the results for testing onto the KITTI server. 
 An ablation study is also conducted on the validation set of Chen's split.
 
 The backbone of the network is ResNet-34 \cite{He2015Resnet}.
 The top 100 pixels of each image are cropped to speed up inference and training. The cropped input images are scaled to $288 \times 1280$.
 The network is trained with a batch size of 4 on a single Nvidia 1080Ti GPU (it takes about 7 GB of GPU memory, significantly less than other SOTA stereo detection algorithms) for 50 epochs on the KITTI training dataset.
 During inference, the network is fed one image at a time, and the total average processing time, including file IO, is about 0.08 s per frame. In contrast, most other stereo-based networks in the KITTI benchmark are several times slower.

 \subsection{Results on Test Set}

 The results are presented in Table~\ref{tab:stereo_test_results} alongside those of other state-of-the-art stereo 3D detection algorithms.

 The proposed YOLOStereo3D is fast and outperforms many pseudo-LiDAR methods or local point cloud methods and is the best-performing algorithm without LiDAR usage.
 It also outperforms DSGN \cite{Chen2020DSGN} on pedestrian detection without an additional training schedule.
 
 \subsection{Test results for Monocular 3D Setting}
 To verify the effectiveness of the proposed anchor pre-processing techniques, they are further tested in the task of monocular 3D object detection.
 Recall that YOLOStereo3D is a monocular detector enhanced with stereo features. By taking away the image from the right camera, the multi-scale fusion module, and the disparity estimation branch, we obtain a standalone monocular detector. We enhance the backbone to be ResNet-101 \cite{He2015Resnet}. Following the proposed YOLOStereo3D, we compute the statistics for each anchor box and filter out deviated anchor boxes during training. 

  \begin{table*}[htb]
    \centering
    \def\arraystretch{1.3}
    \caption{\textbf{Monocular} 3D object detection results of Cars on the KITTI \textbf{test} set.}
    \begin{tabular*}{0.8\textwidth}{  l|c|c }
        \hline
        {\bf Methods} & {\bf $\text{IoU}\ge 0.7$ 3D Easy/Moderate/Hard } & {\bf Time}\\ \hline
        M3D-RPN\cite{Brazil2019M3DRPN}      & 14.76 \% /       9.71 \%       / 7.42 \%  & 0.16s\\
        RTM3D\cite{Li2020RTM3DRM}           & 14.41 \% / 10.34 \% / 8.77 \%   & 0.05s\\
        AM3D\cite{Ma2019AM3D}               & 16.50 \% / 10.74 \% / 9.52 \%   & 0.40s\\
        D4LCN\cite{Ding2019D4LCN}           & 16.65 \% / 11.72 \% / \textbf{9.51} \% & 0.20s\\
        \hline
          \textbf{Ours}           & \textbf{19.24 \%}/  \textbf{12.37 \%}/   8.67 \%  & 0.05s\\
          \hline
  
    \end{tabular*} 
    \label{tab:test_mono}
  \end{table*}

 The network also follow M3D-RPN \cite{Brazil2019M3DRPN} to post-process the prediction results to maximize the 2D-3D coherence. Notice that in YOLOStereo3D, we empirically find this post-processing step deteriorates the final performance, but it is beneficial in the monocular setting.
 
 The results are presented at Table~\ref{tab:test_mono}. 
 As shown in Table~\ref{tab:test_mono}, the proposed framework achieves state-of-the-art performance in KITTI Object Detection Benchmark under the monocular setting. The running time of the proposed monocular detector is about 50 ms per frame.

 \section{Model Analysis and Discussion}
 \label{section:Discussion}

 In this section, we further analyze the performance of YOLOStereo3D and discuss the effectiveness of several important design choices. 
 The baseline model here is only trained on the "Car" type.
We first conduct an ablation study to validate the contribution of anchor preprocessing, hierarchical fusion, and the densely connected ghost module on the validation set.
Then, some qualitative results are presented and discussed.

\input{stereo3d_secs/ablation_study.tex}
\input{stereo3d_secs/cherry_picking.tex}

\subsection{Ablation Study}

\subsubsection{Anchor Preprocessing}
We first test the effectiveness of including statistical information in each anchor.
In the first experiment, instead of predicting a depth value normalized by the depth prior, the network outputs a transformed depth output $\hat z$, where $z = 1/\sigma(\hat z) - 1$, following \cite{Chen2020MonoPair}.
In the second experiment, we do not filter out anchors during training, and the training loss is evaluated with all anchors. We conduct these two experiments in both the stereo setting and the monocular setting. The results are presented in Table~\ref{tab:ablation_study} respectively.

From the two tables, we can observe that anchor priors significantly boost the performance of the network, and filtering out irrelevant anchors during training is also helpful. 
The performance gain can be observed in both monocular 3D detection and stereo 3D detection.
We suggest that we can ease the difficulty of depth inferencing by properly defining and preprocessing anchors specifically for 3D scene understanding in autonomous driving.

The improvement we apply on anchors can also be applied and verified in monocular 3D detection.
We further provide ablation experiments to validate the effectiveness of these processing methods under a monocular 3D detection setting.


\subsubsection{Densely Connected Ghost Module}
Densely connected ghost modules are useful in expanding the number of channels in stereo processing.
Two experiments are conducted to verify its effectiveness.
In the first experiment, a BasicBlock in resnet \cite{He2015Resnet} is adopted to replace the ghost module without expanding the number of channels, resulting in fewer channels during the fusion between RGB features and stereo features.
In the second experiment, we directly upsample the number of channels with $1\times 1$ convolution before feeding the tensor into a BasicBlock. 

We can observe from Table~\ref{tab:ablation_study} that the densely connected ghost module is useful in improving the network's capability. From the first experiment, we demonstrate that expanding the number of channels is crucial for the network's performance.
In the second experiment, we further show that the densely-connected ghost module is better at preserving information than the naive $1\times 1$ convolution.

\subsubsection{Hierachical Fusion}
We respectively disable the stereo matching output on scale 8/16 to produce two networks to justify the usage of multi-scale fusion. We can observe from Table~\ref{tab:ablation_study} that the results of the two ablated models are inferior to that of the baseline model. 
We also point out that the forward pass of stereo-matching modules on scales of 8 and 16 is several times faster than that on the scale of 4.

As a result, we argue that hierarchically fusing stereo features from scales 8 and 16 is worth the effort. 
The baseline structure of hierarchical fusion in YOLOStereo3D achieves a fair balance between speed and performance.

\subsubsection{Disparity Supervision}
An ablation study on the importance of disparity supervision is also conducted. 
Similar to the conclusion in DSGN \cite{Chen2020DSGN}, disparity supervision significantly boosts the performance of the network. 

The experiment shows that such supervision is essential, but the results are not sensitive to the accuracy of the "target" disparity map.
The insight is that the network may only need slight regularizations in stereo-matching submodules. The auxiliary loss is required to drive the network from falling back to a naive local optimal of monocular 3D object detection.

\subsection{Qualitative Results}
We show qualitative validation results in Figure~\ref{fig:examples}. 
The model displayed is YOLOStereo3D sharing the same hyperparameters as the model submitted to the KITTI server, but it is only trained on the training sub-split.

From the RGB images, it can be observed that most of the successful predictions of YOLOStereo3D are visually consistent with the context. As shown in the bird's-eye-view images, though the disparity estimation may not correctly align with the ground truth 3D bounding boxes, the bounding box predictions from YOLOStereo3D are still reasonably accurate. 

The examples suggest that priors in anchor heads and the fusion between stereo matching features and RGB features could help the network to produce more visually consistent predictions and make the network more robust against potentially misleading disparity matching results.

%% file: stereo3d_secs/test_result.tex
\begin{table*}[htb]
    \centering
    \def\arraystretch{1.4}
    \caption{3D object detection results on the KITTI test set on \textbf{Car}.   "*" indicates usage of point cloud data or pretrained disparity estimation module.}
    \begin{tabular*}{0.8\textwidth}{ l|c|r}
        \hline
        {\bf Methods} & {\bf Easy/Moderate/Hard}  & {\bf Time}  \\ \hline
        RT3DStereo\cite{Hendrik2019RT3DStereo}       &29.90 \%/23.28 \%/ 18.96 \%& 0.08s\\
        StereoRCNN\cite{Li2019Stereo}                &47.58 \%/30.23 \%/ 23.72 \%&0.30s\\
        Pseudo-LiDAR*\cite{Weng2019Plidar}           &54.53 \%/34.05 \%/ 28.25 \%&0.40s\\
        OC Stereo*\cite{Pon2019OCStereo}             &55.15 \%/37.60 \%/ 30.25 \%&  0.35s\\
        ZoomNet*\cite{xu2020Zoomnet}                 &55.98 \%/38.64 \%/ 30.97 \%&  0.35s\\
        Disp R-CNN(velo)*\cite{Sun2020DispRCNN}      &59.58 \%/39.34 \%/ 31.99 \%&  0.42s\\
        Pseudo-LiDAR++*\cite{You2019PLPP}            &61.11 \%/42.43 \%/ 36.99 \%&  0.40s\\
        DSGN*\cite{Chen2020DSGN}                     &73.50 \%/52.18 \%/ 45.14 \%&  0.67s\\
        \hline 
        \textbf{Ours YOLOStereo3D}  & \textbf{65.68 \%}/\textbf{41.25 \%}/ \textbf{30.42 \%}& \textbf{0.08s}\\
        \hline
    \end{tabular*}
    
    \label{tab:stereo_test_results}
\end{table*}

\begin{table*}
    \centering
    \def\arraystretch{1.4}
    \caption{3D object detection results on the KITTI test set on \textbf{Pedestrians}.}
    \begin{tabular*}{0.8\textwidth}{ l|c|r}
        \hline
        {\bf Methods} & {\bf Easy/Moderate/Hard} & {\bf Time}  \\ \hline
        RT3DStereo\cite{Hendrik2019RT3DStereo}       &  3.28 \%/  2.45 \%/  2.35 \%& 0.08s\\
        OC Stereo*\cite{Pon2019OCStereo}             & 24.48 \%/ 17.58 \%/ 15.60 \%& 0.35s\\
        DSGN*\cite{Chen2020DSGN}                     & 20.53 \%/ 15.55 \%/ 14.15 \%& 0.67s\\
        \hline 
        \textbf{Ours YOLOStereo3D}  &  \textbf{28.49 \%}/ \textbf{19.75 \%}/ \textbf{16.48 \%}& \textbf{0.08s}\\
        \hline
    \end{tabular*}
    
    \label{tab:test_ped_results}
\end{table*}

%% file: stereo3d_secs/ablation_study.tex
\begin{table*}
    \caption{Ablation study results of cars on the KITTI validation set}
    \def\arraystretch{1.4}
    \centering
    \begin{tabular*}{0.8\textwidth}{  l|c }
        \hline
        {\bf Methods} & {$\text{IoU}\ge 0.7$ 3D Easy/Moderate/Hard } \\ \hline
        \textbf{YOLOStereo3D}               & \textbf{72.06 \%}/  \textbf{46.58 \%}/   \textbf{35.53 \%}  \\
        \hline
        w/o Anchor Prior        & 65.09 \%/   41.38 \%/  30.90 \% \\
        w/o Anchor Filtering    & 71.37 \%/   45.03 \%/  35.83 \% \\
        \hline
        w/o Channel Expand      & 64.16 \%/   39.96 \%/   30.02 \% \\
        w Naive Channel-expand  & 70.70 \%/   45.74 \%/   34.87 \% \\
        \hline
        w/o Scale 8             & 70.80 \%/   45.71 \%/   35.86 \% \\
        w/o Scale 16            & 68.64 \%/   44.54 \%/   33.95 \% \\
        \hline
        w/o Disparity supervision& 62.58 \%/   39.09 \%/   30.34 \% \\
        w PC supervision         & 72.05 \%/   46.59 \%/   35.62 \% \\
        \hline
        \textbf{YOLOMono3D}               & \textbf{21.66 \%}/  \textbf{14.20 \%}/   \textbf{11.07 \%}  \\
        \hline
        Mono w/o Prior        & 19.90 \%/   13.36 \%/   9.68 \% \\
        Mono w/o Filtering    & 20.50 \%/   13.45 \%/  10.50 \% \\
    \hline
    \end{tabular*} 
    \label{tab:ablation_study}
\end{table*}

%% file: stereo3d_secs/cherry_picking.tex
\begin{figure*}
    \centering
  \includegraphics[width=1.0\textwidth]{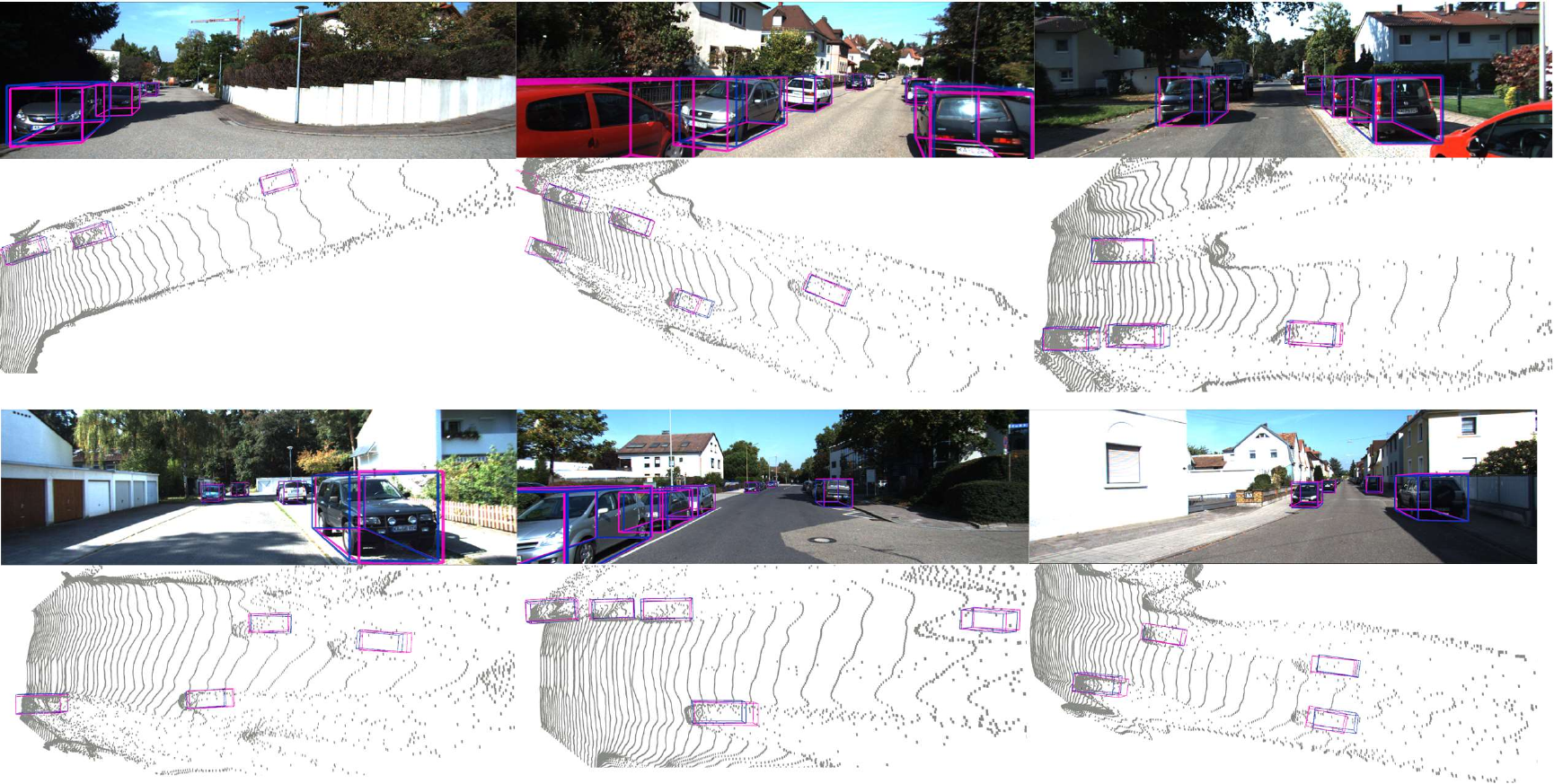}
    \caption{Qualitative examples from the validation set. The RGB images show the detection results and ground truth 3D bounding boxes on the left images. The bird's eye view images show the disparity prediction from the networks, along with detection results.
    The blue bounding boxes are 3D predictions from YOLOStereo3D, the pink bounding boxes are ground truth 3D bounding boxes, and point clouds are predictions from the disparity estimation branch of YOLOStereo3D.
    }
    \label{fig:examples}
\end{figure*}

%% file: stereo3d_secs/sec-4-conclusion.tex
\section{Conclusion}
\label{section:Conclusion}
In this chapter, we introduced YOLOStereo3D, a novel and efficient framework for stereo 3D object detection. The primary contribution of this work lies in reimagining stereo 3D object detection as an extension of monocular detection, shifting away from the less accurate pseudo-LiDAR-based approach. This paradigm shift allowed for the incorporation of real-time monocular 3D object detection principles, utilizing anchor-based priors for depth inference. We further innovated by integrating a point-wise correlation module into the detection framework and employing a hierarchical fusion strategy that skillfully balances information retention and computational efficiency. YOLOStereo3D's performance was rigorously evaluated on the KITTI Object Detection Benchmark, where it demonstrated competitive results among existing stereo frameworks. Notably, it achieves this high level of performance while operating at a rate exceeding ten frames per second, all without relying on LiDAR data.

However, it is important to note a critical aspect of the framework: the asymmetric computational treatment of the stereo images. The model primarily converts features from the right image to the left, resulting in a significant loss of information in the right image. This limitation becomes particularly pronounced in scenarios where an object, occluded in the left image, is more visible in the right image, potentially leading to sub-optimal performance.

Despite this limitation, YOLOStereo3D represents a significant step forward in stereo detection research. Its ability to deliver competitive results with modest computational resources—a single GPU and reduced training time—lowers the entry barrier for research in this field. Moreover, with its accelerated inference speed and robust performance, YOLOStereo3D holds promise for enhancing the deployment of stereo vision systems in autonomous vehicles and mobile robotics, potentially expanding their applicability and efficiency.

%% file: chapter/06_scaleup.tex
\input{scaleup_secs/sec-0-introduction}

\input{scaleup_secs/sec-1-related-works}

\input{scaleup_secs/sec-2-methods}
\input{scaleup_secs/sec-3-experiments}

\input{scaleup_secs/sec-4-conclusion}

%% file: scaleup_secs/sec-0-introduction.tex
\section{Introduction}
\label{section:Introduction}

The pursuit of a comprehensive 3D understanding of dynamic environments stands as a cornerstone in the fields of robotics, autonomous driving, and augmented reality. Contemporary 3D detection algorithms leveraging LiDAR point clouds have shown remarkable performance, owing to the enhanced accuracy and density of these point clouds \cite{Yun2018Focal}. Nevertheless, when contrasted with LiDAR, cameras are found to be cost-effective, power-efficient, and offer greater flexibility in installation. This has led to their widespread use in robots and autonomous vehicles, consequently making 3D detection via single cameras an increasingly compelling area of research within robotics and computer vision.

\begin{figure}[htb]
    \centering
    \includegraphics[{width=1.0\textwidth}]{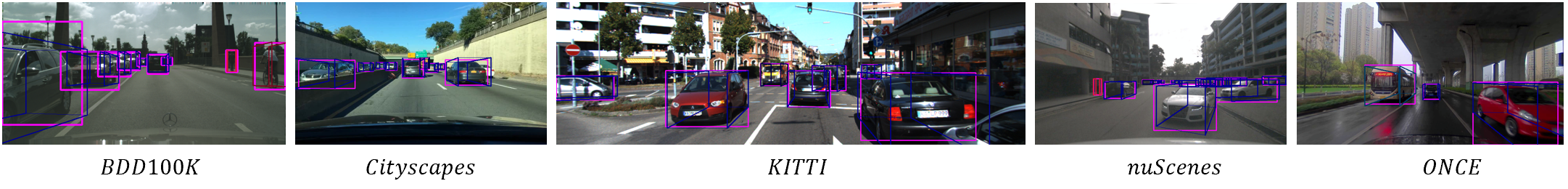}
    \caption{Visualizations of the detection results of our method on five different datasets: BDD100K, Cityscapes, KITTI, nuScenes, and ONCE. Our proposed approach allows for algorithm training across multiple datasets with inconsistent camera settings. Furthermore, our method leverages 2D detection labels for training 3D detection algorithms, significantly reducing the annotation costs associated with 3D detection data. This approach enables the model to exhibit robust 3D detection capabilities when applied to new datasets that only provide 2D annotation labels. The pink box represents the 2D detection results.
    }
    \label{fig:fivdata}
\end{figure}

Over the last few years, monocular 3D object detection has experienced a significant evolution. Models employing Bird-Eye-View (BEV) representation have notably amplified potential in settings involving multi-sensor fusion, frequently encountered in autonomous driving scenarios. However, models utilizing front-view representation, the inherent representation of camera images, are not only faster but also simpler to deploy.

Despite these advancements, deploying monocular 3D object detection continues to grapple with major challenges, with data being a principal concern. When aiming to deploy a 3D detection model on a robot equipped with a single camera in a new environment, obtaining 3D data labeled with LiDAR points becomes a hurdle, and fine-tuning the model with data gathered from the robot is unattainable. It is, therefore, paramount that the system depends on the pretrained model's generalization capability with almost zero-shot fine-tuning. Furthermore, generating a model with robust generalization performance necessitates expansive datasets, which are considerably costly to compile.

In light of these challenges, we put forth methodologies designed to augment the deployment of monocular 3D detection models through effective utilization of existing data. Firstly, we explore the intricacies of training vision-based 3D detection models on a combined assortment of public 3D datasets with varied camera parameters. In doing so, an output representation is devised for the monocular 3D detection model that remains invariant to changes in camera parameters, and, concurrently, a paradigm is established to train models on datasets with different labels. This approach could substantially boost model scales for individual researchers and developers.

Subsequently, a strategy is devised for training 3D models using 2D data. Several existing methods, such as MonoFlex\cite{MonoFlex}, annotate objects on the heatmap based on the 3D center projected on the image as opposed to the center of 2D bounding boxes. We formulate a training strategy that allows these models to fine-tune with 2D labels. This method enables us to fine-tune existing models with cheaper 2D-labeled data procured from on-site robots, thereby transferring 3D knowledge from public 3D datasets to the target environment. Throughout this chapter, experiments conducted to fine-tune a pre-trained model using KITTI 2D data and Cityscapes 2D data are presented as evidence of the potential of our proposed method.

Moreover, by integrating the two proposed methods, we can significantly expand our monocular 3D detection models by training them on a mix of multiple public 3D/2D datasets. This dramatically enlarges the data available during the training phase, thereby significantly enhancing the generalization ability of the resulting model. 
A model is pretrained on a combined set of KITTI\cite{Geiger2012KITTI}, nuScenes\cite{nuscenes2019}, ONCE \cite{mao2021one}, Cityscapes\cite{gahlert2020cityscapes}, and BDD100K \cite{bdd100k} datasets and then fine-tuned on the target dataset using only 2D labels. We subsequently test the model's generalization ability on the target dataset.

The primary contributions of our work can be summarized as:

\begin{itemize}
\item Examination of models such as MonoFlex, and formulation of an output representation robust to varying camera settings, which serves as the foundation for training models on disparate datasets.
\item Development of strategies to train or fine-tune 3D vision models on a blend of 3D/2D datasets.
\item Extensive experiments were conducted on a combined set of KITTI, nuScenes, ONCE, Cityscapes and BDD100K datasets to assess the efficacy of our proposed methods.
\end{itemize}


%% file: scaleup_secs/sec-1-related-works.tex
\section{Related Works}
\label{section:Relate}

\subsection{Monocular 3D Object Detection}

Recent developments in monocular 3D detection can be broadly categorized into two methodologies: bird-eye-view (BEV) methods and perspective-view (PV) methods.

\textbf{Bird-Eye-View Methods:} These methods focus on conducting monocular 3D detection directly within 3D spaces, which simplifies the design of output representation. However, the main challenges arise in the transformation between perspective-view images and features in 3D coordinates. Pseudo-LiDAR represents a series of works that reconstruct 3D RGB-point clouds from monocular depth prediction models, subsequently applying LiDAR-based 3D object detectors to the reconstructed point cloud \cite{Weng2019Plidar,Vianney2019RefinedMPL,Ma2019AM3D,Ku2019MonoPSR}. Another approach directly performs differentiable feature transformation, constructing 3D features from image features, enabling end-to-end training of 3D detection in 3D spaces \cite{Cody2021CaDDN}. These methods often delegate the scale-ambiguity problem to depth prediction sub-networks or attention modules. However, network inferencing in the BEV space heavily relies on 3D labels from LiDAR or direct supervision from LiDAR data, making it challenging to utilize existing 2D datasets or inexpensive 2D labeling tools. Consequently, researchers with limited access to 3D labeled data may encounter difficulties when deploying or fine-tuning networks in new environments.

\textbf{Perspective-View Methods:} These methods, which conduct monocular 3D detection in the original perspective view, are more intuitive. Many one-stage monocular 3D object detection methods are built upon existing 2D object detection frameworks. The main challenge here lies in devising robust and precise encoding/decoding methods to bridge 3D predictions and dense PV features. Various techniques have been proposed, such as SS3D \cite{Jorgensen2019SS3D}, which adds additional 3D regression parameters, and ShiftRCNN \cite{Li2019ShiftRCNN}, which introduces an optimization scheme. Other works like M3DRPN \cite{Brazil2019M3DRPN}, D4LCN \cite{Ding2019D4LCN}, and YoloMono3D \cite{Liu2021YOLOStereo3D} employ statistical priors in anchors to enhance 3D regression accuracy. Additionally, SMOKE \cite{liu2020SMOKE}, RTM3D \cite{Li2020RTM3DRM}, Monopair \cite{Chen2020MonoPair}, and KM3D\cite{KM3D2020Li} utilize heatmap-based keypoints prediction with anchor-free object detection frameworks like CenterNet \cite{zhou2019objects}.

Despite these advancements, most researchers focus on training within a single homogeneous dataset, leading to overfitting in certain camera settings. In this work, we propose a more robust output representation based on MonoFlex \cite{MonoFlex}, enabling network training across different datasets. We also introduce training strategies that allow our model to be trained on 2D datasets as well.

\subsection{Task Incremental Learning}

Task incremental learning represents a specific scenario within incremental learning where models are sequentially trained to recognize new classes across different training phases \cite{riemannian2018IL}. This means that the models are exposed to datasets with varying sets of annotations, particularly when employing a joint set of data labeled with different types. This approach fosters adaptability and flexibility, allowing the model to evolve and accommodate new information without losing its proficiency in previously learned tasks. It represents a significant step towards creating more dynamic and responsive learning systems, capable of adapting to the ever-changing landscape of data and information. Most of the research are focused on Softmax-based multi-class classification \cite{yun2021defense} where activations from different classes all have impacts on each output.

In this chapter, we re-formulate the categorical masking strategy in the setting of joint-dataset training under sigmoid-based classification, allowing more flexible dataset usage.

%% file: scaleup_secs/sec-2-methods.tex
\section{Methods}
\label{sec:methods}

\subsection{Camera Aware Monoflex Detection Baseline}
Monocular 3D object detection involves estimating the 3D location center $(x, y, z)$, dimensions $(w, h, l)$ and planar orientation $\theta$ of objects of interest with a single camera image. Since most SOTA detectors perform prediction in the camera's front-view, we generally predict the projection of the object center on the image plane $(c_x, c_y)$ instead of the 3D position $(x, y)$. The orientation $\theta$ is further replaced with the observation angle
\begin{equation}
    \alpha = \theta - \text{arctan}(\frac{x}{z}),
\end{equation}
which better conforms with the visual appearance of the object \cite{MonoFlex, Brazil2019M3DRPN}.


 MonoFlex \cite{MonoFlex} is an anchor-free method.
It extracts feature maps from the input images with a DLA \cite{DLA2018Yu} backbone, similar to CenterNet\cite{zhou2019objects} and KM3D \cite{KM3D2020Li}. As an anchor-free algorithm, MonoFlex predicts the center positions of target objects with a heat map.
Monoflex primarily consists of the following components: 2D detection, dimension estimation, orientation estimation, keypoint estimation and depth estimation ensemble.
The depth prediction z is simultaneously estimated from both geometry and direct prediction, and these predictions are adaptively ensembled to obtain the final result.

Since we aim to perform joint training on diverse datasets, where data collection involves different cameras with distinct camera settings, our method needs to overcome the challenges posed by varying camera parameters in order to facilitate effective knowledge transfer across these datasets. Given that MonoFlex exhibits insensitivity to camera parameters, we build upon MonoFlex as the foundation of our approach.

Overall, our approach is similar to MonoFlex, with the main difference being that our method has modifications in the depth prediction component, which takes camera parameters into account. Specifically, in the original MonoFlex paper, the direct regression part of depth prediction assumes an absolute depth of
\begin{equation}
    z_{r} = \frac{1}{\sigma(z_{o})} - 1,
\end{equation}
where $z_{o}$ is the unlimited network output following \cite{Chen2020MonoPair, zhou2019objects} and 
\begin{equation}
\sigma(x) = \frac{1}{1 + e^{-x}}.
\end{equation}
While in our method, we have improved it by taking camera's parameters into account as follows:
\begin{equation}
    z_{r} = (\frac{1}{\sigma(z_{o})} - 1) \times \frac{f_{x}}{f_{x0}} .
\end{equation}
where $f_{x}$ is epresents the focal length of the camera used in the training dataset, while $f_{x0}$ is a hyperparameter, we set its value to 500 in our experiments.


\subsection{Selective Training for Joint 3D Dataset Training}
\label{sec:why-base-vtr-failed}
\begin{figure*}[]
    \setlength{\abovecaptionskip}{0pt}
    \setlength{\belowcaptionskip}{0pt}
    \centering
    \includegraphics[width=1.0\textwidth,height=0.22\textheight]{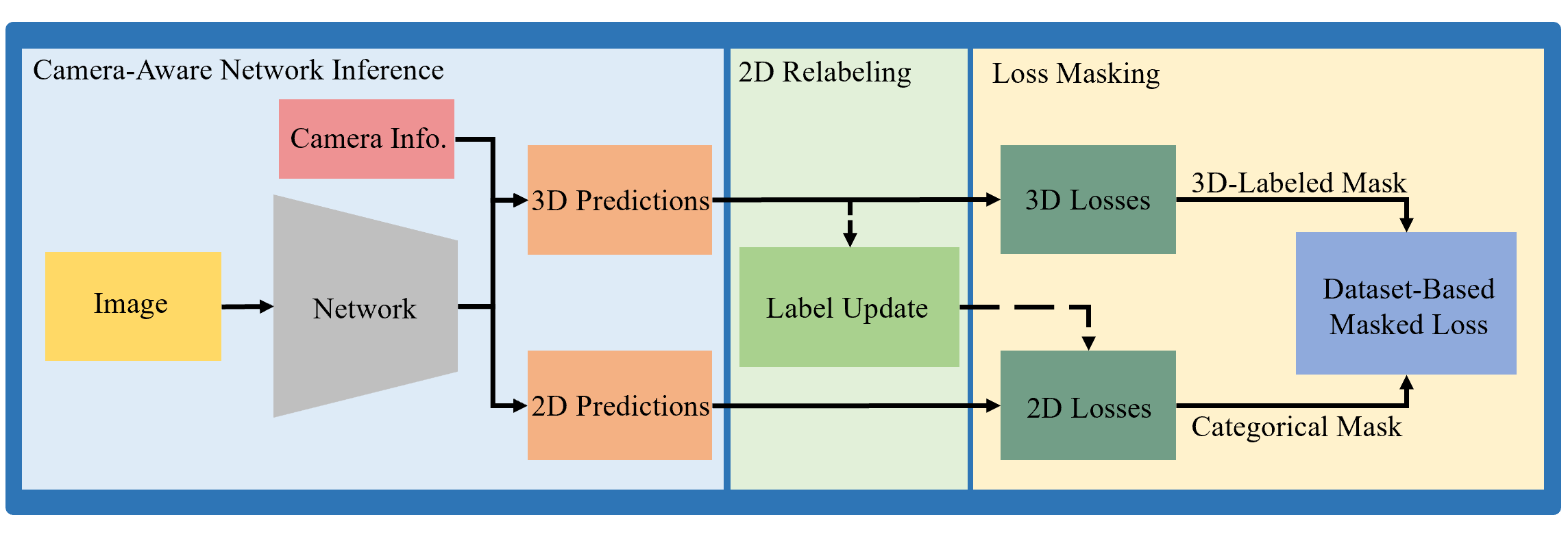}
    \captionsetup{font={footnotesize}}
    \captionsetup{justification=justified} 
    \caption{This figure illustrates the training process of our proposed method. It shows how the pre-trained model's inference, combined with the 2D annotations from the dataset, facilitates the training of a 3D detection model on datasets that lack 3D training labels.}
    \label{training}
\end{figure*}

During dataset training, some parts of the dataset are incompletely annotated. For instance, in the case of KITTI, certain categories like "Tram" are not labeled. However, it is not appropriate to treat the data from KITTI as negative samples for the "Tram" category. Therefore, each data point, in addition to its own annotations, is associated with the categories labeled in the respective dataset. This association provides supervision for the network's classification output and is applied only to the categories with annotations. 
Specifically, for different 3D datasets with varying labeled categories, each data frame stores the categories currently labeled in the dataset. When calculating the loss, we do not penalize (suppress) the detection predictions for unannotated categories.
In summary, this approach effectively handles the issue of incomplete annotations during dataset training. It ensures that only annotated categories influence the model's training, and unannotated categories do not lead to erroneous model behavior. The training procedure is shown in Fig. \ref{training}.

\subsection{Regulating 2D Labels of 2D Datasets for Pseudo 3D training}
\begin{figure}[]
    \setlength{\abovecaptionskip}{0pt}
    \setlength{\belowcaptionskip}{0pt}
    \centering
    \includegraphics[width=1.0\linewidth]{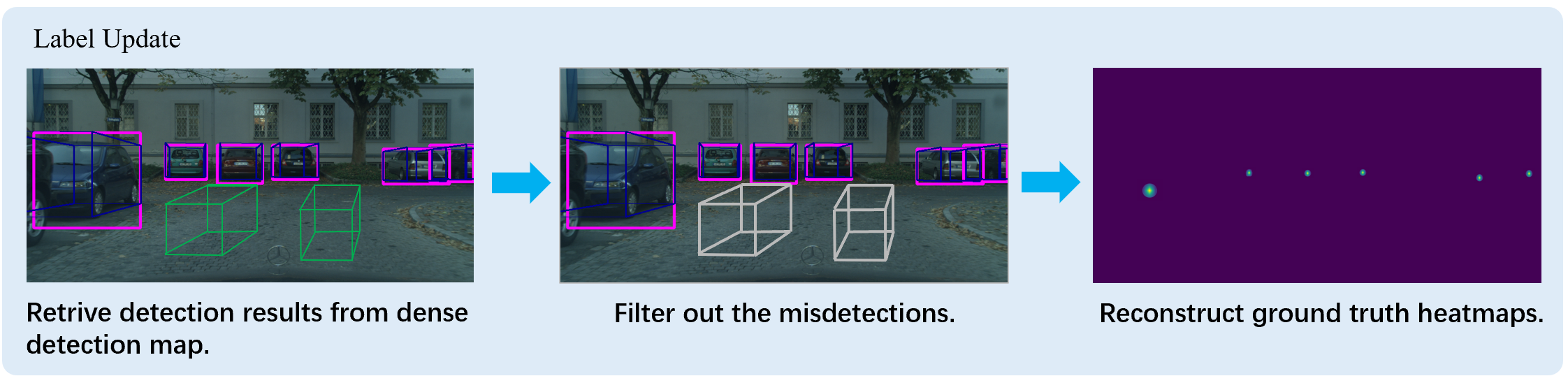}
    \captionsetup{font={footnotesize}}
    \captionsetup{justification=justified} 
    \caption{The figure depicts the training label update process. In the left image, the pre-trained 3D detection model's predictions on new data are shown, which may include some erroneous detections, as indicated by the green boxes. The middle image illustrates the process of identifying and filtering out these erroneous detections, marking them in gray based on the matching results. The right image represents the reconstruction of the ground truth heatmap using the pseudo 3D labels.}
    \label{label_update}
\end{figure}

In the Monoflex method, the center of the heatmap is determined by projecting the 3D center. For data with only 2D annotations, we have no direct means of generating supervision signals for the object's 3D center. 
Therefore, we need to find a way to generate 3D detection supervision information from 2D labels. We propose a novel method to train 3D detection algorithms solely relying on 2D detection labels. Specifically, we start by feeding data with only 2D annotations into a pre-trained 3D detection model and set a very low score threshold to enable the model to produce multiple detection results (including both 2D and 3D detection results). This step may include some erroneous or inaccurate detections.
Next, we use the 2D training labels from the new dataset to match them with the 2D detection results obtained in the previous step, filtering out erroneous or inaccurate detections to obtain pseudo 3D training labels. Finally, we reconstruct the ground truth heatmap and 2D detection map (as shown in Fig. \ref{label_update}), and ultimately, we calculate the loss between the model predictions and the pseudo 3D labels only on the heatmap and 2D detection map.
The method of training a monocular 3D model using 2D annotated data is illustrated in Algorithm \ref{alg:pseudo3d2d}, we refer to it as \textit{Pseudo 3D Training with 2D Labels}.
\input{scaleup_secs/alg_p2D}



\begin{figure}[t]
    \setlength{\abovecaptionskip}{0pt}
    \setlength{\belowcaptionskip}{0pt}
    \centering
    \includegraphics[width=1.0\linewidth]{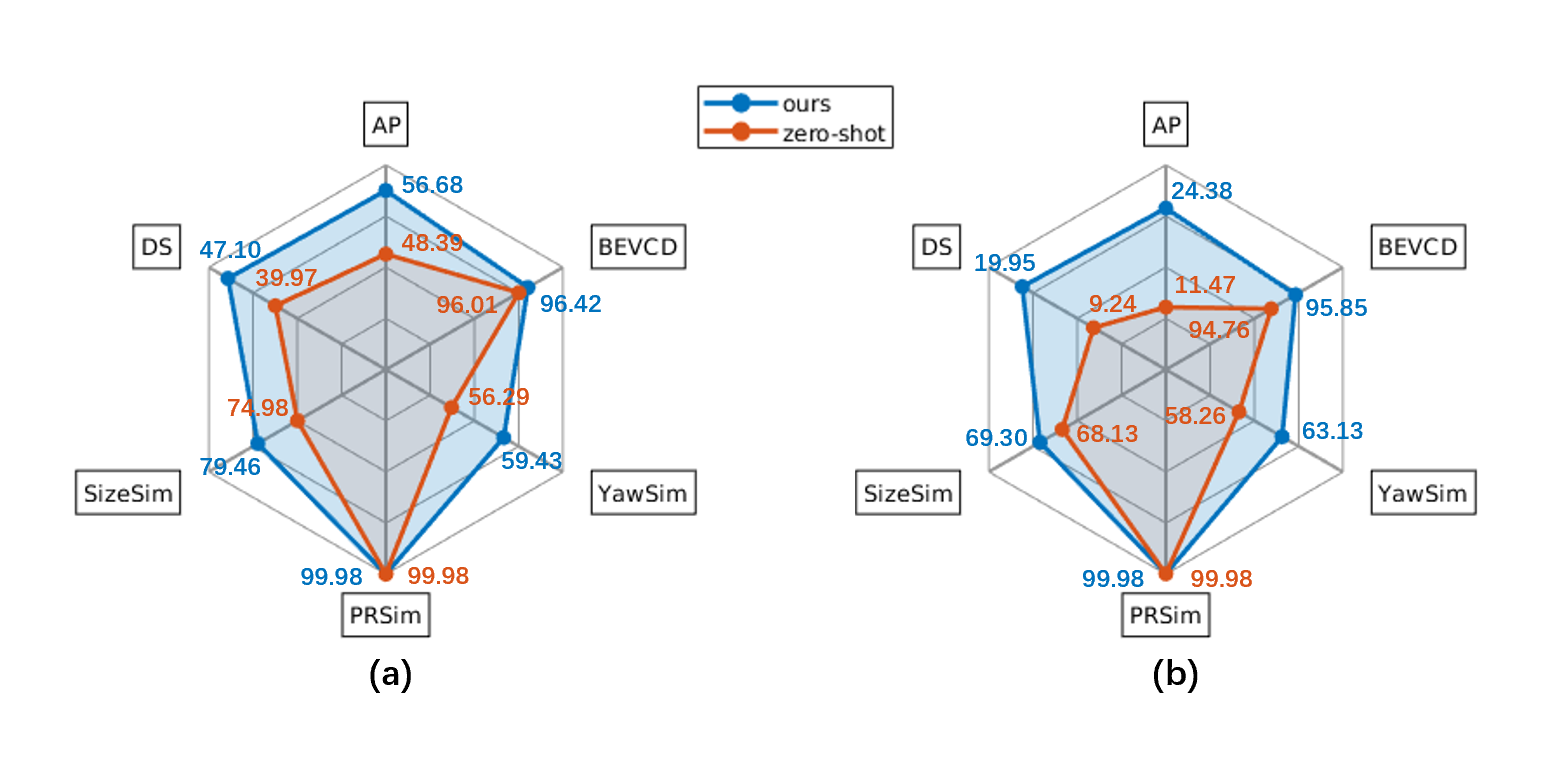}
    \captionsetup{font={footnotesize}}
    \captionsetup{justification=justified} 
    \caption{The comparison between our method and zero-shot on the Cityscapes dataset shows that our method outperforms zero-shot in all the evaluation metrics.}
    \label{spider_plot}
\end{figure}

%% file: scaleup_secs/alg_p2D.tex
\begin{algorithm}
    \caption{Pseudo 3D training with 2D Label}\label{alg:pseudo3d2d}
    \textbf{Input:}  Dense Detection Maps $F$, Labeled 2D boxes $B_{t}$\\
    \textbf{Output:} Loss $l$
    \begin{algorithmic}[1]
    \State {\textbf{Initialization:}}
    \State $B_{p}:$ Detection results from dense detection maps
    \State $B_{t}^{'}:$ Pseudo ground truth
    \State $M:$ IoU matrix
    \State {\textbf{Main Loop:}}
    \State Retrieve detection results $B_{p}$ from dense detection maps $F$.
    \State Compute IoU matrix $M$ between $B_p$ and $B_t$.
    \State Compute the matching with the minimum cost, take 3D centers of $B_p$ as ground truth for $B'_t$.
    \For {each matched $b_p$, $b_t$}
        \If{$cost_i > eps$}
            \State Marked as mis-detection and removed from $B'_t$.
        \EndIf
    \EndFor
    \State Reconstruct ground truth heatmaps and 2D detection maps $F_t$ from $B'_t$.
    \State Compute Loss $l$ with pseudo ground truth $B'_t$ only on heatmaps and 2D detection maps.
    \end{algorithmic}
\end{algorithm}

%% file: scaleup_secs/sec-3-experiments.tex
\section{Experiments}
\label{section:Experiments}

\input{scaleup_secs/tabel_nusc_kitti}

\subsection{Datasets Reviews}

We evaluate the proposed networks on the KITTI 3D object detection benchmark \cite{Geiger2012KITTI} and Cityscapes dataset \cite{gahlert2020cityscapes}. The KITTI dataset consists of 7,481  training frames and 7,518 test frames. Chen \textit{et al.} \cite{Chen2015kittisplit} split the training set into 3,712 training frames and 3,769 validation frames. The Cityscapes dataset contains 5000 images split into 2975 images for training, 500 images for validation, and 1525 images for testing. 




\subsection{Evaluation Metrics}
\subsubsection{KITTI 3D}
All the testing and validation results, are evaluated with 40 recall positions ($AP_{40}$), following Simonelli \textit{et al.} \cite{Simonelli2019MonoDIS} and the KITTI team. Such a protocol is considered to be more stable than the $AP_{11}$ proposed in the Pascal VOC benchmark \cite{Everingham10pascal}.
\subsubsection{Cityscapes 3D}
Following \cite{gahlert2020cityscapes}, we use these five metrics: 2D Average Precision (AP), Center Distance (BEVCD), Yaw Similarity (YawSim), Pitch-Roll Similarity (PRSim), Size Similarity (SizeSim) and Detection Score (DS) to evaluate the performance on Cityscapes 3D dataset. Among them, DS is a combination of the first five metrics and computed as:
\begin{footnotesize}
\begin{equation}
 DS = AP \times \frac{BEVCD+YawSim+PRSim+SizeSim}{4}. 
\end{equation}
\end{footnotesize}
For details, please refer to the paper \cite{gahlert2020cityscapes}.

\subsection{Experiment Setup}
bTo demonstrate the superiority of our method over zero-shot approaches, when testing on the KITTI dataset, we designed our experiments as follows:

\begin{itemize}
        \item We initially pre-trained our model on four datasets: BDD100K, nuScenes, ONCE, and Cityscapes.
        \item With the pre-trained model, we evaluated its zero-shot detection performance on the KITTI dataset.
        \item Subsequently, we fine-tuned the model using our method, which involves training a 3D detection model using 2D training labels from KITTI.
        \item Finally, we obtained detection results on the KITTI dataset based on our method.
    
\end{itemize}

When testing on the Cityscapes dataset, we followed the same experimental setup on the KITTI dataset.



\input{scaleup_secs/table_cityscapes}

\subsection{Experiment Results and Comparison}
\begin{figure}[]
    \setlength{\abovecaptionskip}{0pt}
    \setlength{\belowcaptionskip}{0pt}
    \centering
    \includegraphics[width=1.0\linewidth]{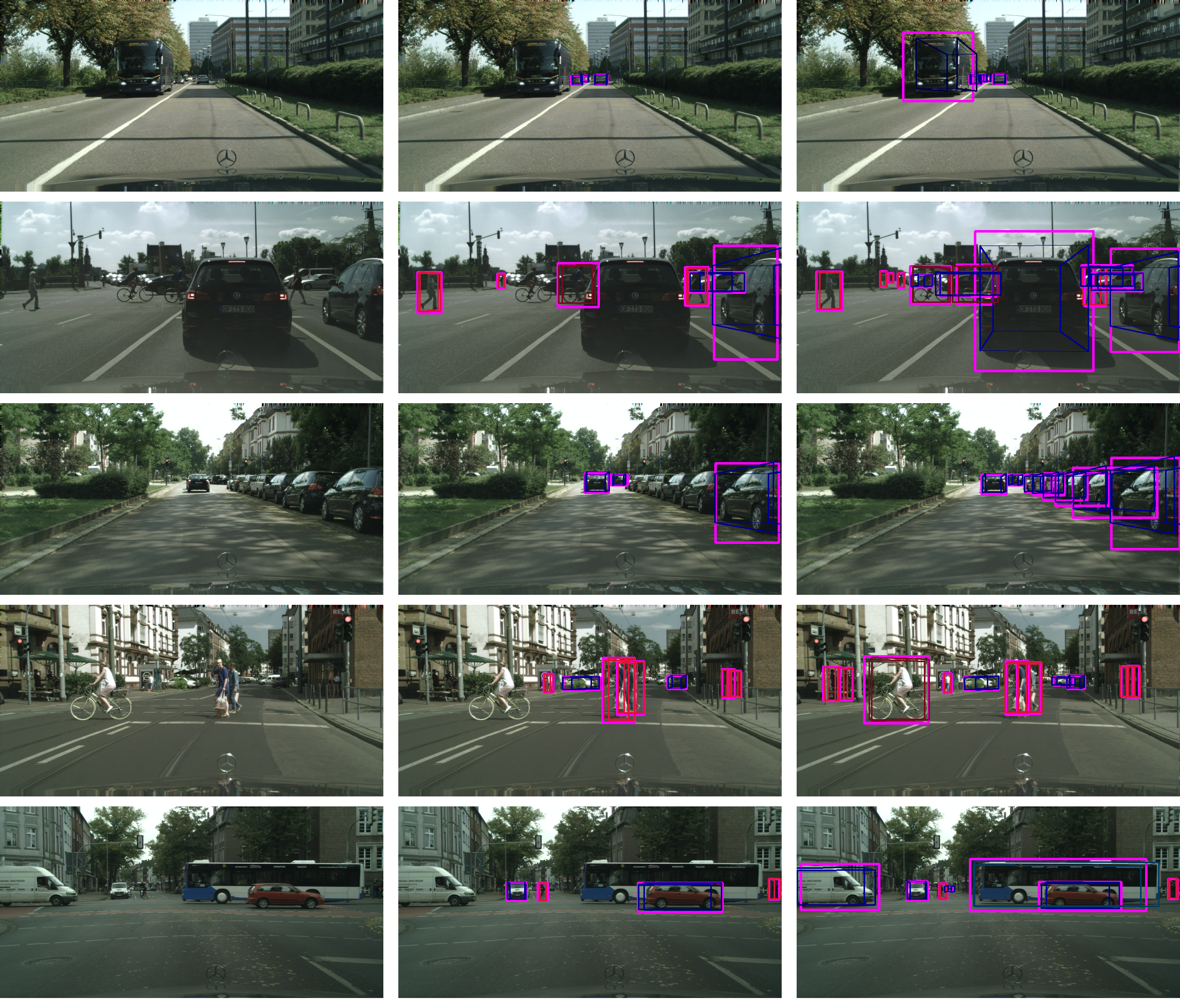}
    \captionsetup{font={footnotesize}}
    \captionsetup{justification=justified} 
    \caption{The qualitative results on the Cityscapes dataset. The leftmost column contains the original images, the middle column displays the zero-shot results, and the rightmost column shows the results obtained using our method. The pink boxes represent 2D detection results.}
    \label{vis_compare}
\end{figure}

The quantitative results of the Car category and Pedestrain category on KITTI dataset are reported in Table~\ref{tab:test_results}, while quantitative results of Car category and Truck category on Cityscapes dataset are shown in Table~\ref{tab:test_results_city_car} and Table~\ref{tab:test_results_city_truck} respectively.
We also utilized radar charts to visualize the performance on the Cityscapes dataset, as depicted in Fig. \ref{spider_plot}. Our proposed method exhibits significant improvements over zero-shot learning across five key metrics: AP (Average Precision), BEVCD (Bird's Eye View Center Distance), YawSim (Yaw Similarity), SizeSim (Size Similarity), and DS (Detection
Score). Fig. \ref{vis_compare} illustrates the qualitative results on the Cityscapes dataset.

From Table~\ref{tab:test_results}, we can observe that our method has achieved significant improvements in both 3D and 2D detection tasks compared to zero-shot learning. Specifically, for the 3D detection task, our method has shown an improvement in $AP_{3D}$ for the "Car" category in the easy, moderate, and hard difficulty levels by 16.42\%, 4.27\%, and 0.20\%, respectively. In the "Pedestrian" category, the AP improvements are even more substantial, with increases of 50.7\%, 32.73\%, and 35.34\% for the easy, moderate, and hard difficulty levels, respectively.
The improvements in $AP_{2D}$ for the "Car" category in the easy, moderate, and hard difficulty levels by 5.54\%, 11.25\%, and 12.39\%, respectively, as well as the AP enhancements of 5.69\%, 6.83\%, and 8.80\% in the "Pedestrian" category across the same difficulty levels for the 2D detection task. These improvements in 2D detection performance are easily comprehensible because we trained the algorithm comprehensively on the new dataset using 2D training labels.

Table~\ref{tab:test_results_city_car} and Table~\ref{tab:test_results_city_truck} present the detection results for the "Car" and "Truck" categories on the Cityscapes dataset. Quantitative results indicate that our method, when compared to the zero-shot approach, has shown significant improvements in various metrics, except for the PRSim metric (as we only focused on the yaw angle, considering pitch and roll angles to be 0). Specifically, in the "Car" category, we observed an increase of 17.13\% in AP, a 4.17\% improvement in BEVCD, a 5.58\% improvement in YamSim, a 5.97\% improvement in SizeSim, and a remarkable 17.84\% enhancement in DS. 
In the "Truck" category, AP, BEVCD, YawSim, SizeSim, and DS have improved by 112.55\%, 1.16\%, 8.36\%, 17.17\%, and 115.91\%, respectively.
From these experimental results, we can also observe that our method demonstrates a more significant improvement in detection performance for categories with fewer instances in the dataset. This is due to the fact that the model has less knowledge about categories with fewer instances, leading to weaker generalization on new datasets compared to categories with more instances. After pseudo 3D learning using 2D labels on the new dataset, we achieved a substantial improvement in detection performance.

%% file: scaleup_secs/tabel_nusc_kitti.tex
\begin{table*}[h!]
    \centering
    \caption{Object detection results on the KITTI dataset. }
    \scriptsize
    \def\arraystretch{1.5}
    \setlength\tabcolsep{4pt} 
    \begin{tabular*}{0.96\textwidth}{ |c|c|c|c|c|c|c|c|}
        \cline{1-8}
       \multirow{2}{*}{\bf } & \multirow{2}{*}{\bf Methods} & \multicolumn{3}{c|}{\bf Car}   & \multicolumn{3}{c|}{\bf Pedestrain}   \\ \cline{3-8}
         & & {\bf Easy (\%)} & {\bf Moderate (\%)}& {\bf Hard (\%)}  &{\bf Easy (\%)}&{\bf Moderate(\%)}& {\bf Hard (\%)}   \\ \cline{1-8}
        \multirow{2}{*}{$3D$} &Zero-shot & 30.69 & 23.88  & 20.34  &  7.83 & 6.66 & 5.49  \\
        \cline{2-8}
        &Ours  & 35.73 \color{table_color}($\uparrow 16.42\%$) & 24.90  \color{table_color}($\uparrow 4.27\%$)
                        & 20.38  \color{table_color}($\uparrow 0.20\%$)   &  11.80 \color{table_color}($\uparrow 50.70\%$) 
                        & 8.84 \color{table_color}($\uparrow 32.73\%$)& 7.43 \color{table_color}($\uparrow 35.34\%$) \\
        \cline{1-8}
        \multirow{2}{*}{$2D$} &Zero-shot & 90.91& 81.28& 71.50 &  59.04 & 50.84 & 43.76  \\
        \cline{2-8}
        &Ours  & 95.95 \color{table_color}($\uparrow 5.54\%$)& 90.42 \color{table_color}($\uparrow 11.25\%$)
                        & 80.36  \color{table_color}($\uparrow 12.39\%$)&  62.40 \color{table_color}($\uparrow 5.69\%$)
                        & 54.31 \color{table_color}($\uparrow 6.83\%$)& 47.61  \color{table_color}($\uparrow 8.80\%$)\\
        \cline{1-8}
    \end{tabular*}
    
    \label{tab:test_results}
\end{table*}

%% file: scaleup_secs/table_cityscapes.tex
\begin{table*}
    \centering
    \caption{Object detection results of class 'Car' on the Cityscapes dataset. }
    \footnotesize
    \def\arraystretch{1.3}
    \setlength\tabcolsep{4pt} 
    \begin{tabular*}{1.0\textwidth}{ |c|c|c|c|c|c|c|c|}
        \cline{1-7}
        \multirow{2}{*}{\bf  Methods} & \multicolumn{6}{c|}{\bf Metrics}      \\ \cline{2-7}
          & {\bf DS (\%)} & {\bf AP (\%)}& {\bf BEVCD (\%)}  &{\bf YawSim (\%)}&{\bf PRSim(\%)}& {\bf SizeSim (\%)}   \\ \cline{1-7}
      
        Zero-shot & 39.97 & 48.39  & 96.01  &  56.29 & 99.98 & 74.98  \\
        \cline{1-7}
        
        Ours  & 47.10 \color{table_color}($\uparrow 17.84\%$) & 56.68  \color{table_color}($\uparrow 17.13 \%$)
                        & 96.42  \color{table_color}($\uparrow 4.27\%$)   & 59.43 \color{table_color}($\uparrow 5.58\%$) 
                        & 99.98 \color{table_color}($\uparrow 0\%$)& 79.46 \color{table_color}($\uparrow 5.97\%$) \\
        \cline{1-7}
    \end{tabular*}
    
    \label{tab:test_results_city_car}
\end{table*}

\begin{table*}
    \centering
    \caption{Object detection results of class 'Truck' on the Cityscapes dataset. }
    \footnotesize
    \def\arraystretch{1.3}
    \setlength\tabcolsep{3pt} 
    \begin{tabular*}{1.0\textwidth}{ |c|c|c|c|c|c|c|c|}
        \cline{1-7}
        \multirow{2}{*}{\bf  Methods} & \multicolumn{6}{c|}{\bf Metrics}      \\ \cline{2-7}
          & {\bf DS (\%)} & {\bf AP (\%)}& {\bf BEVCD (\%)}  &{\bf YawSim (\%)}&{\bf PRSim(\%)}& {\bf SizeSim (\%)}   \\ \cline{1-7}
        
        Zero-shot & 9.24 & 11.47  & 94.76  &  58.26 & 99.98 & 68.13  \\
        \cline{1-7}
        
        Ours  & 19.95 \color{table_color}($\uparrow 115.91\%$) & 24.38  \color{table_color}($\uparrow 112.55\%$)
                        &95.86  \color{table_color}($\uparrow 1.16\%$)   &  63.13 \color{table_color}($\uparrow 8.36\%$) 
                        & 99.98 \color{table_color}($\uparrow 0\%$)& 69.30 \color{table_color}($\uparrow 17.17\%$) \\
        \cline{1-7}
    \end{tabular*}
    
    \label{tab:test_results_city_truck}
\end{table*}

%% file: scaleup_secs/sec-4-conclusion.tex
\section{Conclusion}
\label{section:Conclusion}

In this chapter, we initially conducted research on models such as MonoFlex and developed strategies that are resilient to changes in camera intrinsics. These strategies allow the models to be trained on diverse datasets. Additionally, we designed a learning approach that enables monocular 3D detection models to acquire 3D detection knowledge based solely on 2D labels, even in datasets that only provide 2D training labels. Lastly, we carried out extensive experiments on a combination of datasets, including KITTI, nuScenes, Cityscapes, and others. The experimental results demonstrated the efficacy of our approach.


Despite its success, our work does have limitations. First, our method is currently applicable only to models that are insensitive to camera parameters, such as MonoFlex. For models where the influence of camera parameters cannot be disregarded.
Moreover, when encountering new categories in a novel dataset, the lack of relevant supervision from previous datasets may result in suboptimal detection performance for these new categories. 
Overcoming these limitations demands continuous optimization and improvement of our approach.


%% file: chapter/05_monodepth.tex
\input{monodepth_secs/sec-0-introduction}

\input{monodepth_secs/sec-1-related-works}

\input{monodepth_secs/sec-2-methods}

\input{monodepth_secs/sec-3-experiments}

\input{monodepth_secs/sec-4-conclusion}

\input{monodepth_secs/sec-5-appendix}

%% file: monodepth_secs/sec-0-introduction.tex
\section{Introduction}
\label{section:Introduction}

Dense depth prediction from a single RGB image presents a significant challenge in computer vision and holds great value for the robotics community. While active sensors like LiDAR offer accurate depth measurements of environments, camera-based systems remain popular due to their cost-effectiveness, power efficiency, and adaptability across various robotic platforms. Thus, monocular depth prediction enables more sophisticated 3D perception and planning tasks for many camera-based robots and low-cost self-driving setups \cite{tase1,tase2,tase3}, also including products like Tesla's FSD and Valeo's vision-only system \cite{TeslaAI}. A qualitative prediction sample of our work is presented in Figure ~\ref{fig:kitti360_init_example}, showing the potential of directly perceiving the world in 3D with a monocular camera.

\input{monodepth_secs/image_kitti360_example}

Efficiently deploying a monocular depth prediction network in a new environment raises challenges for existing methods. As pretrained vision models do not usually generalize well enough to new scenes or a new camera setup \cite{MonodepthSurvey2022}, it is extremely useful and convenient to directly train a depth estimation model using raw data collected from the deployed robot in the target environment. However, most of the current self-supervised monocular depth using only monocular image sequences can only provide depth prediction with an ambiguity in the global scale of the depth results unless using stereo images \cite{monodepth2, manydepth2021temporal, watson-2019-depth-hints, ramamonjisoa-2021-wavelet-monodepth} or external point clouds supervision in training \cite{FuCVPR18DORN}. 

A mobile robotic system or a self-driving car usually produces a robot's pose online by a standalone localization module. Moreover, with onboard sensors like the inertial measurement unit (IMU) and wheel encoder, the localization module can produce accurate relative poses between consecutive keyframes. Thus, we focus on the usage of poses to tackle scale ambiguity so that the network can be trained to predict depths with correct global scales.

In this chapter, the FSNet framework is developed step by step from the use of pose. First, we investigate the training process of the unsupervised MonoDepth and demonstrate why directly replacing the output from PoseNet \cite{monodepth2} with real camera poses will cause failure in training. This justifies the multichannel output representation adopted in our chapter. Then, the availability of poses enables the computation of optical-flow-based masks for dynamic object removal. Moreover, a self-distillation framework is introduced to help stabilize the training process and improve final prediction accuracy. Finally, to use historical poses in test time, we introduce an optimization-based post-processing algorithm that uses sparse 3D points from visual odometry (VO) to improve the test time performance and enhance the robustness to changes in the extrinsic parameters of the robot. The main contributions of the chapter are as follows:
\begin{itemize}
    \item An investigation of the training process of the baseline unsupervised monocular depth prediction networks. A multichannel output representation enables stable network training with full-scale depth output.
    \item An optical-flow-based mask for dynamic object removal inspired by the introduction of inter-frame poses.
    \item A self-distillation training strategy with aligned scales to improve the model performance while not introducing additional test-time costs.
    \item An efficient post-processing algorithm that fuses the full-scale depth prediction and sparse 3D points predicted by visual odometry to improve test-time depth accuracy.
    \item A validation and ablation study of the proposed algorithms on the KITTI raw dataset \cite{Geiger2012KITTI}, KITTI-360 dataset \cite{KITTI360}, and the nuScenes dataset\cite{nuscenes2019} to test the performance of the proposed system.
\end{itemize}

The remainder of this chapter is organized as follows:
Section II reviews related work. Section III introduces our FSNet. Section IV presents the experimental results and compares our framework with existing self-supervised monocular depth prediction frameworks. 
Section V presents ablation studies and discussions on the components proposed in the chapter. Finally, we conclude this chapter in the last section.

%% file: monodepth_secs/image_kitti360_example.tex
\begin{figure}[htb]
    \centering
    \includegraphics[{width=0.7\textwidth}]{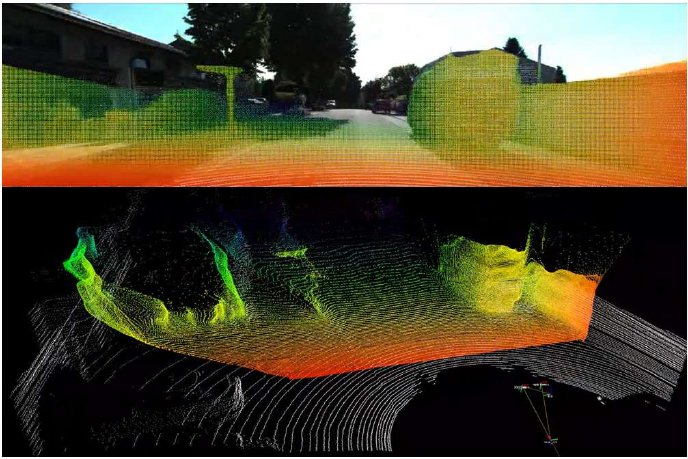}
    \caption{Prediction result sample from KITTI-360 dataset. White Points are points from LiDAR sensing. Colored points are prediction results from FSNet, which correctly express the 3D scenes with dense points.
    }
    \label{fig:kitti360_init_example}
\end{figure}

%% file: monodepth_secs/sec-1-related-works.tex
\section{Related Works}
\label{section:Relate}

\subsection{Self-Supervised Monocular Depth Prediction}
Monodepth2 \cite{monodepth2} sets up the baseline framework for self-supervised monocular depth prediction. The core idea of the training of Monodepth2 is to teach the network to predict dense depths that reconstruct the target image from source views. ManyDepth \cite{manydepth2021temporal} utilizes the geometry in the matching between temporal images to improve the accuracy of single image depth prediction. However, the cost volume construction slows the inference speed and makes it difficult to accelerate on robotics platforms. 

To obtain the correct scale directly from monocular depth prediction, researchers use either LiDAR \cite{Eigen2014DepthPrediction, FuCVPR18DORN, Wong2020UnsupervisedDepthVIO} or stereo cameras \cite{watson-2019-depth-hints, ramamonjisoa-2021-wavelet-monodepth, wimbauer2020monorec, yang20d3vo, ChengDepthPredictionMonoDVO} to provide supervision. 
Among stereo methods, Depth Hints \cite{watson-2019-depth-hints}, and Wavelet Depth \cite{ramamonjisoa-2021-wavelet-monodepth} use traditional stereo matching algorithms to provide direct supervision to the depth prediction network. Monorec \cite{wimbauer2020monorec} uses stereo images to provide scale for the network and to identify moving object pixels in the sequence. DNet \cite{xue2020toward} uses the ground plane estimated from the normal of the image to calibrate the global scale of the predicted depth, but it fails in the nuScenes dataset which contains many image frames without a clear ground plane.

Our proposed FSNet uses poses from robots to regularize the network to predict correct scales without requiring a multiview camera or LiDARs. We point out that relative poses between frames are commonly available for robotic platforms because of sensors like IMUs and wheel encoders.
\input{monodepth_secs/image_struct}
\subsection{Knowledge Distillation}
Knowledge distillation (KD) is a field of pioneering training methods that transfers knowledge from teacher models to target student models \cite{Hinton2015DistillingTK}. KD has been applied in various tasks, including image classification \cite{yun2020regularizing}, object detection \cite{Chen2017LearningEO, Yun2021ConflictKD}, and incremental learning \cite{yun2021KDIncre}. The philosophy of knowledge distillation is training a small target model with an additional loss function to mimic the output of a teacher model. The teacher network could be an ensemble of the target models \cite{Asif2020EnsembleKD}, a larger network \cite{Yang2020KnowledgeDV}, or a model with additional information inputs or different sensors \cite{Chong2022MonoDistillLS}. Self-distillation (SD) is a special case where the teacher model is completely identical to the student model, and no additional input is applied to train the teacher model. Some existing works have demonstrated its performance on image classification \cite{Zhang2019BeYO}. 

In this chapter, we first investigate the training process of an unsupervised monocular depth predictor, and then we apply an offline SD to our proposed model to regularize the training process and improve the final performance.

\subsection{Visual Odometry}
Visual odometry (VO) systems \cite{klein2009parallel,forster2016svo,campos2021orb} are widely used to provide robot-centric pose information for autonomous navigation \cite{engel2012camera} and simultaneously estimate the ego-motion of the camera and the position of 3D landmarks. 
With IMU or Global navigation satellite system (GNSS) information, the absolute scale of the estimations can also be recovered \cite{qin2018vins,cao2022gvins}.

ORB-SLAM3 \cite{campos2021orb} is a typical indirect VO method in which the current input frame is tracked online with a 3D landmark map built incrementally.
In this tracking process, the local features extracted are first associated with landmarks on the map.
Then, the camera pose and correspondences are estimated within a Perspective-n-Point scheme.
The depth of the local feature point can be retrieved from the associated 3D landmark and the camera pose, yielding an image-aligned sparse depth map for each frame.
In this work, we use these sparse depth maps as an input for the full-scale dense depth prediction post-optimization procedure.

\subsection{Monocular Depth Completion}
Monocular depth completion means predicting a dense depth with a monocular image and sparse ground truth LiDAR points. Learning-based methods usually follow the scheme of partial differential equations (PQEs) that extrapolate depth between sparse points based on the constraints learned from RGB pixels \cite{cheng2018dcspn}.  
Other new methods without deep learning or extra ground truth labels exist. IPBasic \cite{ku2018defense} produces dense depth with morphological operations. Bryan et al. \cite{Bryan2021Deter} introduce superpixel segmentation to segment images into different units, and each unit is considered as a plane or a convex hull.

In our proposed post-processing step, the merging between dense network depth prediction and visual odometry can be formulated into the same optimization problem as monocular depth completion but with extra uncertainty and hints. The proposed method is formulated without extra LiDAR point labels.

%% file: monodepth_secs/image_struct.tex
\begin{figure*}
    \centering
    \includegraphics[{width=0.9\textwidth}]{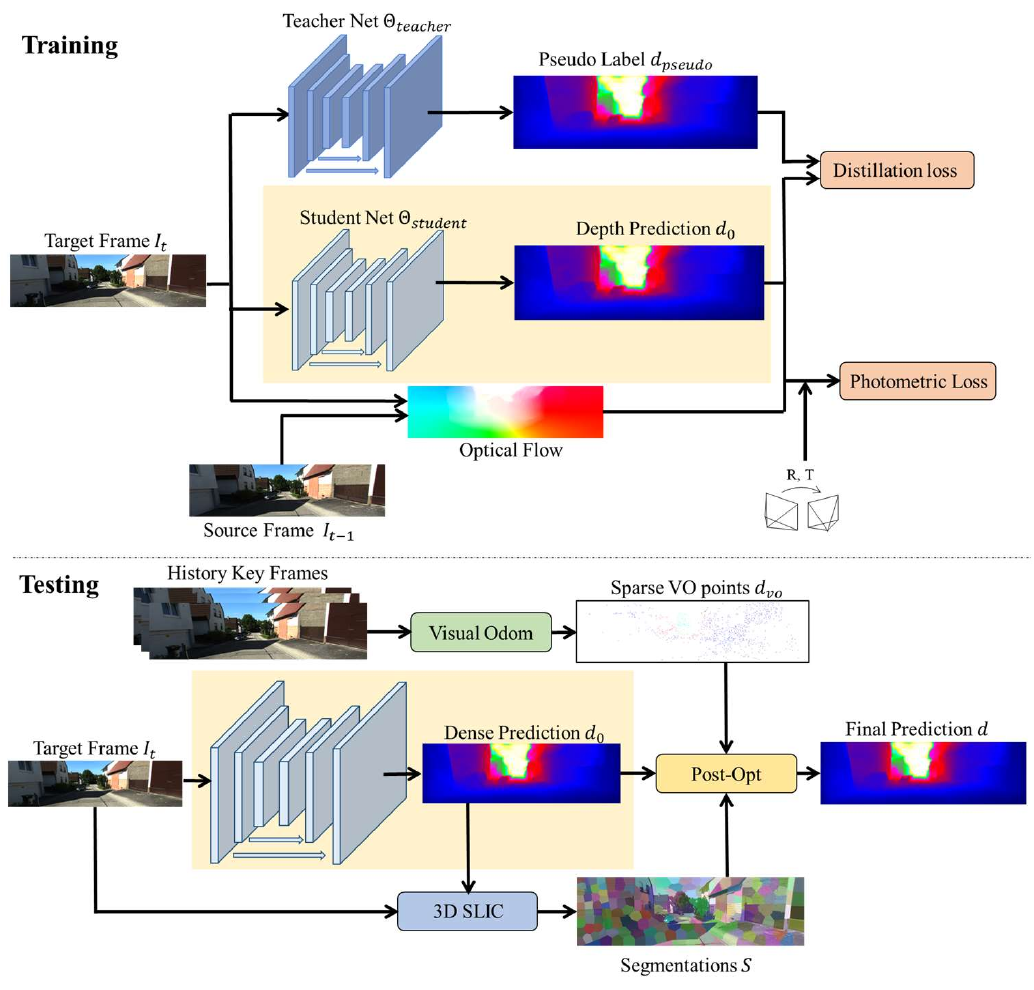}
    \caption{
    The training and inference process of FSNet. During training, the teacher network $\Theta_{teacher}$ and the student network $\Theta_{student}$ utilize a U-Net structure with a ResNet backbone.
    $\Theta_{teacher}$ is pretrained using only the photometric loss and is frozen during the training of the student network. $\Theta_{student}$ is trained with a summation of photometric loss and the distillation loss. During testing, the trained network first predict dense full-scale Depth $d_0$.  The prediction will be merged with sparse VO points $d_{vo}$ to produce accurate final prediction. The 3D SLIC algorithm is used to segment the image and limit the scale of the post-optimization problem.
    }
    \label{fig:monodepth_struct}
\end{figure*}

%% file: monodepth_secs/sec-2-methods.tex
\section{MonoDepth2 Review}

\label{sec:monodepth2}
We first review the pipeline of MonoDepth2. The target is to train a model $\mathcal{F} $ mapping the input target image frame $I_0$ to a dense depth map $d_0$, with reconstruction supervision from neighboring frames. The depth $d$ is decoded from the convolutional output $x$ as 
\begin{equation}
    \frac{1}{d} = \frac{1}{d_{max}} + \text{sigmoid}(x) * (\frac{1}{d_{min}} - \frac{1}{d_{max}}),
\end{equation}
where $d_{max}=100$ and $d_{min}=0.1$ are hyper-parameters for the boundary of the depth prediction. 

The depth map is reprojected to neighboring source keyframes with pose predicted from PoseNet.  Then, the target frame is reconstructed using colors sampled from neighboring source frames and produces $I_0^l$ where $l \in \{-1, 1\}$. The photometric loss is the weighted sum of the structural similarity index measure (SSIM) and the L1 difference between the two images. The loss is expressed as:
\begin{equation}
    l_{photo}(d_0) = \alpha \frac{1 - SSIM(I_0, I_0^l)}{2} + \beta |I_0 - I_0^l|,
\end{equation}
where $SSIM()$ measures the structural difference between the two images using a $3\times 3$ kernel, and $\alpha$ and $\beta$ is a constant with value 0.85 and 0.15, following \cite{monodepth2}.

As shown in our-own experiments and \cite{wei2022surround}, directly substituting PoseNet with poses from the localization module will cause failure in training the network. We notice that the baseline depth decoder method will produce corrupted reconstruction results at initialization if directly fed with the correct pose.  We need to take additional measures to preserve organic reconstruction results during the start of the training.

\section{Methods}
In order to stably train the depth prediction network with inter-frame poses, we reformulate the output of MonoDepth2 as multi-channel outputs. Inter-frame poses are further utilized to filter out dynamic objects during training with an optical flow-based mask. The overall system pipeline is presented in Fig~\ref{fig:monodepth_struct}.
\subsection{Output Modification From MonoDepth2}
\label{sec:modification}
\textbf{Multichannel Output:} In order to preserve organic reconstruction results, we select reformulate the output as multi-channel outputs, to allow larger depth values during initialization and also unsaturated gradients at large depth values.

Assuming the predicted depth $d$ should be within $[d_{min}, d_{max}]$ and the output is decoded from $N$ channels of the output network, then $q= \frac{d_{max}}{d_{min}}^{1/N}$ is defined as the proportion between consecutive depth bins, and the $i$th bin will represent $d_i = d_{min} \times q^{i-1}$. Considering the softmax activated output of the network being $Z \in \mathbb{R}^N$, we interpret the weighted mean of each depth bin as the predicted depth $d = \sum_i z_i \cdot d_i$. 

At initialization, the network will predict an almost-uniform distribution for each depth bin, so the initial decoded depth will be the algorithmic mean of the depth bins $d' = \frac{1}{N} \sum_i d_i$. To guide the selection of hyperparameters, we analyze the initial depth values $d'$ in our framework.

\textbf{Theorem :} The algorithmic mean $d'$ of a proportional array, bounded by $[d_{min}, d_{max}]$ will converge to the following limit as the number of bins $N$ grows: 
\begin{equation}
    \lim_{N\rightarrow\infty} d' = \frac{d_{max} - d_{min}}{\ln{d_{max}} - \ln{d_{min}}}.
\end{equation}

\textbf{Proof:} Denote the boundary of the predicted depth as $[d_{min}, d_{max}]$, the $i$ th bins  will represent $d_i = d_{min} \times q^{i-1} = d_{min} (\frac{d_{max}}{d_{min}})^{i/N}$. As presented in the chapter, the initial depth predicted by the network will be the mean of all depth bins $d' =  \frac{1}{N}\sum_i{d_i}$.

The mean of the proportional array will be 
\begin{align}
\begin{split}
    \lim_{N \rightarrow \infty} d' &= \lim_{N \rightarrow \infty} \frac{1}{N}\sum_i{d_i} = \lim_{N\rightarrow\infty} \frac{1}{N} \sum_i{d_{min} (\frac{d_{max}}{d_{min}})^{i/N}}\\
      &= \lim_{N \rightarrow \infty} \frac{1}{N} \frac{d_{max} - d_{min}}{1 - q} \\
      &=  (d_{max} - d_{min}) \lim_{N\rightarrow\infty} \frac{1}{N(1 - (\frac{d_{max}}{d_{min}})^{1/N})} \\
      &= (d_{max} - d_{min}) \lim_{M\rightarrow 0}\frac{M}{1 - (\frac{d_{max}}{d_{min}})^M} \\
      &\overset{\mathrm{H}}{=} \frac{d_{max}-d_{min}}{\ln{d_{max}} - \ln{d_{min}}}.
\end{split}
\end{align}

L'Hôpital's rule is applied at $\overset{\mathrm{H}}{=}$ to obtain the final result. 

We set $d_{max}=100m$ for the autonomous driving scene. Based on (5.4), we choose $d_{min}=0.1m$ for our network to obtain $d' > 10m$ to ensure stable training initialization.

\textbf{Camera-Aware Depth Adaption:} To train one single depth prediction network on all six cameras of the nuScenes dataset, we point out that we need to take the difference of camera intrinsic matrix into account. Similar object appearance produces similar features in the network, while depth predictions should vary based on the focal length of the camera. As a result, we adapt the value of the depth bins according to the input camera $f_x$ as $d_i = d_i^0 \cdot \frac{f_x}{f_{base}}$, where $f_{base}$ is a hyper-parameter chosen according to the front camera in nuScenes.
\subsection{Optical Flow Mask}
\label{sec:flow}
With the introduction of ego-vehicle poses at training times, we have a stronger prior for dynamic object removal. We know that the reprojection of a static 3D point on neighboring frames must stay on an epipolar line regardless of the distance between that point to the camera. The epipolar line $L$ can be computed as a vector with a length of three $L = \mathbf{F} \cdot [p_x, p_y, 1] ^T$, where $(p_x, p_y)$ are the the coordinate of the pixel in the base frame and $\mathbf{F}$ is the fundamental matrix between the two frame. The fundamental matrix $F$ is further calculated from $\mathbf{F}=K^T \cdot [t]_{x} \cdot R \cdot K$, with camera intrinsics parameters $K$, and relative pose $(R, t)$ between two image frames. 

Therefore, if the reprojected point was far from its epipolar line, we can estimate that this point is probably related to a dynamic object and should probably be omitted during training loss computation.

To construct the optical flow mask for dynamic object removal, we first compute the optical flow $(d_x, d_y)$ between the two image frames using a pretrained unsupervised optical flow estimator ArFlow \cite{liu2020arflow}. The distance $dis_l$ between the reprojected point from the epipolar line is computed as with 
\begin{equation}
    dis_l = \frac{L \cdot [d_x, d_y, 1]^T}{\sqrt{L_0^2 + L_1^2}},
\end{equation}
where $L_0, L_1$ represents the first and second element of vector $L$. Pixels with a deviation larger than 10 pixels are considered dynamic pixels. In this way, we produce a loss mask for photometric reconstruction loss computation.

\subsection{Self-Distillation}
\label{sec:distill}
As discussed in Section~\ref{sec:monodepth2}, the monocular depth prediction network starts training with a generally uniform depth map. 
The loss function, with a kernel size of $3\times 3$, only produces guaranteed optimal gradient directions to the network when the reconstruction error is within several pixels. As a result, at the start of the training, the network is first trained on pixels with ground truth depths close to the initialization state, and the gradients at other pixels will be noisy. The noise in the gradient will affect the stability of the training process. In order to stabilize the training process by increasing the signal-noise ratio (SNR) of the training gradient, we introduce a self-distillation framework. 

Using the baseline self-supervised framework, we first train an FSNet $\Theta_{teacher}$. The first FSNet will suffer from the noisy learning process mentioned above and produce sub-optimal results. Then, we train another FSNet $\Theta_{student}$ from scratch using the self-supervised framework, with pseudo label $d_{pseudo} = \Theta_{teacher}(I_0)$ from the first FSNet to guide the training. With the pseudo label from the teacher net, the student network will receive an additional meaningful training gradient from the beginning of the training phase. Thus, the problem of noisy gradients introduced above can be mitigated. The training process is summarized in the upper half of Fig. ~\ref{fig:monodepth_struct}.

However, since the pseudo label predicted from teacher FSNet is not always accurate, we encourage the student network to predict the uncertainty $\sigma$ for each pixel adaptively, alongside the depth $d_{0}$. We observe that close-up objects have smaller errors in depth prediction than far-away objects, and we encode this empirical result in our uncertainty model.  Thus, we assume that the logarithm of the depth prediction $\log(d)$ follows a Laplacian distribution with mean $\log(d_{pseudo})$ predicted by the teacher. The likelihood $p$ of the predicted depth can be formulated as:
\begin{equation}
    p(\log(d) | \log(d_{pseudo}), \sigma) = - \frac{|\log(d) - \log(d_{pseudo})|}{\sigma} - \log(\sigma).
\end{equation}
Practically, the convolutional network directly predicts the value of $\log(\sigma)$ instead of $\sigma$ to improve the numerical stability. We adopt $l_{distill} = -p(\log(d) | \log(d_{pseudo}), \sigma)$  as the training loss. 

\subsection{Optimization-Based Post Processing}
\label{sec:post}
To deploy the algorithm on a robot, we expect the system to be robust to online permutation to extrinsic parameters that could affect the accuracy of $D_{0}$ predicted by the network. We introduce a post-processing algorithm to provide an option to improve the run-time performance of the depth prediction module by merging the sparse depth map $D_{vo}$ produced from VO.

We formulate the post-processing problem as an optimization problem. 
The optimization problem over all the pixels in the image is described as:
\begin{equation}
    \begin{array}{ll}
        \underset{D_{out}}{\operatorname{minimize}} & \sum_{i} L^{d_{out}^i} \\
        \text { where } & L^{d^i} = \lambda_0 L^i_{consist} + \lambda_1^{i} L^i_{vo} \\
        & L_{consist}^i = \sum_{j} (
            \log( \frac{d^i_{0}}{d^j_{0}}) -
            \log( \frac{d_{out}^i}{d_{out}^j}))^2 \\
        & L_{vo}^i = [\log(d^i_{vo} / d_{out}^i)] ^ 2
    \end{array}.
\end{equation}

The consistency term $L_{consist}$ is defined by the change in relative logarithm between each pixel compared to $D_{0}$. The visual odometry term $L_{vo}$ works only at a subset of points with sparse depth points, and $\lambda_1^i$ is zero for pixels without VO points.

The problem above is a convex optimization problem that can be solved in polynomial time. However, the number of pixels is too large for us to solve the optimization problem at run time. 


Therefore, we downscale the problem by segmenting images with super-pixels and simplifying the computation inside each super-pixel. The full test-time pipeline is summarized in the second half of Fig. ~\ref{fig:monodepth_struct}.

\subsubsection{3D SuperPixel Segmentation}
The segmentation method we proposed is based on Simple linear iterative clustering (SLIC) which is similar to a K-means algorithm operated on the LAB image. We propose to utilize the full-scale dense depth prediction from the network, including the difference in absolute depth $\Delta_{dep}$ to enhance the distance metric. 

\begin{algorithm}
    \caption{3D SLIC with Depth}\label{alg:cap}
    \textbf{Input} LAB image $I_{lab}$, depth image $D_{0}$.  \\
    \textbf{Output} Set of point sets $S$,\\
    \textbf{Parameters} Grid step $s$, \\cost weights $\Lambda_{slic}=\{\lambda_{lab}, \lambda_{d}, \lambda_{pix}\}$, max iteration $E$
    \begin{algorithmic}[1]
    \State Initialize a grid of cluster centers on the image coordinate. $f^k=\{I_{lab}^k, \vec{X^k}, d^k\}$ with the step size $s$.
    \For {$iteration=1,2,\ldots, E$}
		\For {each pixel $i$}
		    \State Compute the distance to cluster centers $l_{ik} = L_{slic3d}(f^k, f^i, \Lambda)$
		    \State Assign to the closest center set $S^k$.
		\EndFor
		
		\For {each center $k$}
		    \State Update the center vector with the mean among point sets $S^k$
		\EndFor
    \EndFor
    \end{algorithmic}
\end{algorithm}

The feature of each pixel is composed of the LAB color channel $I_{lab}^i$, the coordinate in image frame $X^i$ and the depth $d^i$. The distance between each pixel and the cluster center is the weighted sum of the three distances:

\begin{equation}
\begin{aligned}
    L_{slic3d}(f^k, f^i, \Lambda) &=  \lambda_{lab} \cdot norm(I^k_{lab} - I^i_{lab})  \\
     &+ \lambda_{d} \cdot |d^k - d^i| \\
     &+ \lambda_{pix} \cdot norm(X^k, X^i)
\end{aligned}
\end{equation}

The proposed 3D SLIC algorithm is presented in Algorithm ~\ref{alg:cap}. The output will be a set of point sets $S_k$. Notice that we utilize a GPU to accelerate the process.

\subsubsection{Optimization Reformulation}

After obtaining the segmentation results from 3D SLIC, we re-formulate the problem as a nested optimization problem to simplify the computation. The inner optimization computes scale changes for each pixel while the outer optimization considers the constraints between different segments.

\textbf{Inner Optimization:} For each point cluster, we assume it describes a certain geometric unit and the dense depth is correct up to a scale. The inner-optimization target is determining a scale factor $v$ so that the difference between sparse VO points and the predicted depth is minimized. The optimization for points in each cluster can be formulated as:

\begin{equation}
    \begin{array}{ll}
        \underset{\log{v}}{\operatorname{minimize}} & \frac{1}{2} \sum_i^{N_{vo}^k}(\log{d^i_0} + \log{v} - \log{d_{vo}^i})^2 \\
    \end{array}.
\end{equation}

The solution to the inner optimization problem can be derived as:

\begin{align}
   \frac{\partial L}{\partial \log{v}} &= \sum_i^{N_{vo}^k} (\log{d^i_0} + \log{v} - \log{d_{vo}^i}) = 0 \\ 
   \log{v} &= \frac{1}{N_{vo}^k} \sum_i^{N_{vo}^k}\log\frac{d^i_{vo}}{d^i} = lg^k_{vo}.
\end{align}

\textbf{Outer Optimization:} Rewriting the original pixel-wise optimization problem into a segment-wise one forms:

\begin{equation}
    \begin{array}{ll}
        \underset{lg^k}{\operatorname{minimize}} & L = \sum_{k} L^{lg^k} \\
        \text { where } & L^{lg^k} = \lambda_0 L^{lg^k}_{consist} +  \lambda_1^{k} L^i_{vo} +
        \lambda_2(lg^k - lg^k_0)^2\\
        & L_{consist}^i = \sum_{j} [
            (lg^k - lg^j) - (lg^k_0 - lg^j_0)
            ]^2 \\
        & L_{vo}^i = (lg^k_{tar} - lg^k) ^ 2,
    \end{array}
\end{equation}
where $lg^k_{tar}$ indicates the target for each cluster computed by inner optimization, and $\lambda_1^k=\lambda_1$ if there are VO points inside the $k$-th segment and $0$ otherwise. 

For this convex optimization problem, a solution that satisfies the Karush–Kuhn–Tucker (KKT) condition will be the globally optimal solution. The KKT condition implies the differential of $L$ w.r.t. each variable $lg^k$ being zero:
\begin{align*}
   [(N-1) \lambda_0 + \lambda_1^{k} + \lambda_2] lg^k - \lambda_0 \sum_{i\neq k} lg^i &=\\
   \lambda_2 lg^k_0  + \lambda_1^{k} lg^k_{tar} + &\lambda_0 \sum_{i\neq k} (lg^k_0 - lg^i_0),
\end{align*}
which forms $N$ linear equations. We denote 
$$\Lambda = (N-1) \lambda_0 + \lambda_1^{k} + \lambda_2.$$
The solution vector $\vec{lg}=[lg^0, lg^1,...]^T$ to the system of the linear equation will be $\vec{lg} = A^{-1}B$, where:

\begin{align}
    A &= \begin{bmatrix}
        \Lambda & - \lambda_0  & \cdots & -\lambda_0 \\
        - \lambda_0 & \Lambda  & \cdots & -\lambda_0\\
        \vdots & \vdots  & \ddots   & \vdots  \\
        -\lambda_0 & -\lambda_0  & \cdots\  & \Lambda  \\
        \end{bmatrix} \\
    B &= [\cdots, \lambda_2 lg^k_0  + \lambda_1^{k} lg^k_{tar} + \lambda_0 \sum_{i\neq k} (lg^k_0 - lg^i_0), \cdots]^T.
\end{align}

The overall post-optimization algorithm is presented at Algorithm~\ref{alg:post}. 
\input{monodepth_secs/alg_post_opt}

The solution involves the inverse of an $N\times N$ matrix $A$, whose computational complexity scale grows in $\mathcal{O}(n^3)$. This explains the necessity of downscaling the pixel-wise optimization problem into a segment-wise problem with the proposed 3D SLIC algorithm and the use of a nested optimization scheme.

\input{monodepth_secs/table_test}

%% file: monodepth_secs/alg_post_opt.tex
\begin{algorithm}
    \caption{Post-Optimization}\label{alg:post}
    \textbf{Input} Log-depth image $lg_{net}$, VO depth image $D_{vo}$,  point sets $S$\\
    \textbf{Output} Optimized log-Depth $lg$,
    \begin{algorithmic}[1]
    \For {each cluster $S^k$}
        \State Compute mean-log-depth $lg_0^k = \frac{1}{N^k}\sum_i(lg_{net}^i)$.
        \State Compute optimized $v$ from Equation (14).
        \State Obtain target $lg_{tar}^k = lg_0^k \cdot v$.
    \EndFor
    \State Compute the optimized center from outer optimization $\vec{lg}_{seg} = A^{-1}B$
    \For {each cluster $S^k$}
		\For {each pixel $i$ in the cluster}
		    \State Obtain final log-depth $lg^i = lg_{net}^i \cdot \frac{lg_{seg}^k}{lg_0^k} $
		\EndFor
    \EndFor
    \end{algorithmic}
\end{algorithm}

%% file: monodepth_secs/table_test.tex
\begin{table*}[]
\caption{Performance of full-scale MonoDepth on KITTI, KITTI-360 and nuScenes.
Results on nuScenes are averaged over six cameras. For scale factor, "GT" is using LiDAR median scaling methods and "None" means we directly evaluate the error without post-processing scale. "\textbf{*}" indicates methods using sequential images in test time. The \cellcolor{red!25}{pink} columns are error metrics, the lower the better; the \cellcolor{blue!25}{blue} columns are accuracy metrics, the higher the better.}
    \label{tab:nusc_test}
    \centering
    \footnotesize
    \def\arraystretch{1.3}
    \setlength\tabcolsep{2pt}
    \begin{tabular*}{1.0\textwidth}{|l|l|c|c|c|c|c|c|c|c|}
    \cline{1-10} {\bf Data}&{\bf Methods} & {\bf Scale Fac.} &  \cellcolor{red!25}{Abs Rel } & \cellcolor{red!25}{Sq Rel} & \cellcolor{red!25}{\bf RMSE} & \cellcolor{red!25}{RMSE LOG} & \cellcolor{blue!25}{$\delta<1.25$} & \cellcolor{blue!25}{$\delta<1.25^2$} & \cellcolor{blue!25}{$\delta<1.25^3$} \\ 
    \cline{1-10}
    \multirow{9}{*}{KITTI}& Bian et al. \cite{Bian2019Depth}&\multirow{6}{*}{GT} & 0.128 & 1.047 & 5.234 & 0.208 & 0.846 & 0.947 & 0.976 \\
    & CC \cite{cc2019ranjan}& & 0.139 & 1.032 & 5.199 & 0.213 & 0.827 & 0.943 & 0.977 \\
    & MonoDepth \cite{monodepth2} &   &   0.116  &   0.903  &   4.863  &   0.193  &   0.877  &   0.959  &   0.981  \\
    & DNet \cite{xue2020toward} &  &  0.113  &   0.864  &   4.812  &   0.191  &   0.877  &   0.960  &   0.981  \\
    &\textbf{FSNet (single frame)} &   &   \textbf{0.113}  &   \textbf{0.857}  &   \textbf{4.623}  &   \textbf{0.189}  &   \textbf{0.876} &   \textbf{0.960}  &   \textbf{0.982}  \\
    &  \textbf{*FSNet (post-opt)}  &   & \textbf{0.111}  &   \textbf{0.829}  &   \textbf{4.393}  &   \textbf{0.185}  &   \textbf{0.883}  &   \textbf{0.961}  &   \textbf{0.983}  \\
    \cline{2-10}
    & DNet \cite{xue2020toward} &  &  0.118  &   0.925  &   4.918  &   0.199  &   0.862  &   0.953  &   0.979 \\
    &   \textbf{FSNet (single frame)} &  \multirow{2}{*}{None}   & \textbf{0.116}  &   \textbf{0.923}  &   \textbf{4.694}  &   \textbf{0.194}  &   \textbf{0.871}  &   \textbf{0.958}  &   \textbf{0.981}  \\
    &   \textbf{*FSNet (post-opt)}   &   & \textbf{0.109}  &   \textbf{0.866}  &   \textbf{4.450}  &   \textbf{0.189}  &   \textbf{0.879}  &   \textbf{0.959}  &   \textbf{0.982}  \\
    
    \cline{1-10}
    \multirow{5}{*}{K360}& MonoDepth \cite{monodepth2} & \multirow{3}{*}{GT}  &   0.130  &   0.865  &   4.154  &   0.206  &   0.858  &   0.951  &   0.977  \\
    &\textbf{FSNet (single frame)} &   &   \textbf{0.116}  &   \textbf{0.804}  &   \textbf{3.921}  &   \textbf{0.196}  &   \textbf{0.881} &   \textbf{0.955}  &   \textbf{0.978}  \\
    &  \textbf{*FSNet (post-opt)}  &   & \textbf{0.114}  &   \textbf{0.760}  &   \textbf{3.752}  &   \textbf{0.196}  &   \textbf{0.883}  &   \textbf{0.954}  &   \textbf{0.977}  \\
    \cline{2-10}
    &   \textbf{FSNet (single frame)} &  \multirow{2}{*}{None}  & \textbf{0.129}  &   \textbf{0.757}  &   \textbf{3.954}  &   \textbf{0.228}  &   \textbf{0.860}  &   \textbf{0.955}  &   \textbf{0.978}  \\
    &   \textbf{*FSNet (post-opt)}   &   & \textbf{0.122}  &   \textbf{0.731}  &   \textbf{3.865}  &   \textbf{0.226}  &   \textbf{0.862}  &   \textbf{0.945}  &   \textbf{0.973}  \\
      \cline{1-10}
    
    \multirow{5}{*}{Nusc}& MonoDepth \cite{monodepth2} & \multirow{3}{*}{GT}  &   0.233  &   4.144  &   6.979  &   0.308  &   0.782  &   0.901  &   0.943  \\
    &\textbf{FSNet (single frame)} &  &   \textbf{0.239}  &   \textbf{5.104}  &   \textbf{6.979}  &   \textbf{0.308}  &   \textbf{0.794}  &   \textbf{0.904}  &   \textbf{0.942}  \\
    &\textbf{FSNet (multiframe)} &  &   \textbf{0.235}  &   \textbf{4.503}  &   \textbf{6.923}  &   \textbf{0.307}  &   \textbf{0.786}  &   \textbf{0.895}  &   \textbf{0.937}  \\
    \cline{2-10}
    &\textbf{FSNet (single frame)} & \multirow{2}{*}{None} &   \textbf{0.238}  &   \textbf{6.180}  &   \textbf{6.865}  &   \textbf{0.319}  &   \textbf{0.806}  &   \textbf{0.904}  &   \textbf{0.940}  \\
    &\textbf{FSNet (multiframe)} &  &   \textbf{0.238}  &   \textbf{6.198}  &   \textbf{6.489}  &   \textbf{0.311}  &   \textbf{0.811}  &   \textbf{0.910}  &   \textbf{0.944}  \\
    \cline{1-10}
    \end{tabular*}
    
\end{table*}

%% file: monodepth_secs/sec-3-experiments.tex
\section{Experiments}

\subsection{Experiment Settings}
We first present the dataset and background settings of our experiments.

We utilize the following datasets in our experiments to evaluate the performance of our approach:

\begin{itemize}
    \item KITTI Raw dataset \cite{Geiger2012KITTI}: This dataset was designed for autonomous driving and many existing works have produced official results on it. We mainly evaluate FSNet on the Eigen Split \cite{Eigen2014DepthPrediction}. It contains 39810 monocular frames for training and 697 images from multiple sequences for testing. Images are sub-sample to $192\times 640$ during training and inference for the network.
    \item KITTI-360 dataset \cite{KITTI360}: This dataset is collected with a stable camera parameter. The dataset also provides indexes to keyframes to avoid static frames during the training of the depth network. We select eight sequences with 51170 keyframes for training and sub-sample the remaining sequences to obtain 1106 frames for testing. This dataset contains fewer static frames and dynamic objects compared to the KITTI dataset. We show that our method could produce better depth prediction results on cleaner datasets like KITTI-360.
    \item NuScenes dataset \cite{nuscenes2019}: This dataset contains 850 sequences collected with six cameras around the ego-vehicle.  We separate the dataset under the official setting with 700 training sequences and 150 validation sequences. We only select scenes without rain and night scenes during both training and validation.  We uniformly sub-sample the validation set for validation.  Images are sub-sampled to $448 \times 762$. Unlike \cite{Vitor2022FSM}, we treat images at each frame as six independent samples during both training and testing. The model will have to adapt to different camera intrinsic and extrinsic parameters. Furthermore, because the lidar and the cameras are not synchronized, there will be noise in the poses between frames. FSNet needs to overcome these problems to achieve stable training, and also predict scale-aware depths.
\end{itemize}

\textbf{Depth Metrics:} Previous works, including \cite{monodepth2, manydepth2021temporal, wimbauer2020monorec, watson-2019-depth-hints} mainly focus on metrics where the depth prediction is first aligned with the ground truth point clouds using a global median scale before computing errors. In this chapter, we first present data on the scaled metrics for comparison with existing methods. Then, we focus on metrics \textbf{without} median scaling in ablation studies.

\textbf{Data Augmentation:} Besides photometric augmentation adopted in MonoDepth2 \cite{monodepth2}, we implement horizontal flip augmentation for image-pose sequences, where we also horizontally flip the relative poses between image frames.

\textbf{FSNet Setting:} We adopt ResNet-18 \cite{He2015Resnet} as the backbone encoder for KITTI and KITTI-360 datasets following prior mainstream works for a fair comparison, and we adopt ResNet-34 for the nuScenes dataset because of the increasing difficulty. The PoseNet is dropped and poses from the dataset are directly used in image reconstruction, and no additional modules are created except for a frozen teacher net during distillation. In summary, we basically share the inference structure of MonoDepth2, and we do not train a standalone PoseNet. In the KITTI dataset, FSNet spends 0.03s on network inferencing and 0.04s on post-optimization, measured on RTX 2080Ti. We point out that multichannel output does not noticeably increase network inference time with additional width only in the final convolution layer.

\subsection{Single Camera Prediction}
The performance on the KITTI and KITTI-360 datasets is presented in Table~\ref{tab:nusc_test}. We point out that, even though FSNet spends extra network capability to memorize the scale of the objects in the scenes, it can produce even better depth maps than baseline models by utilizing multi-channel output, flow mask, and self-distillation.

We present results on both single-frame settings and multi-frame settings. We could appreciate the improvement from post-optimization in the result table. We further point out that our method  decouples the prediction of a single frame and post-optimization into standalone modules, which makes it flexible for different application settings. 

\input{monodepth_secs/image_kitti360_frames}

We present some qualitative results on the validation split of the KITTI-360 dataset in Figure~\ref{fig:kitti360_frame_examples}. The images in the first row include difficult scenes, including high contrast, large trucks, or complex plants. Images in the second row show the edges around cars or other objects. 
\input{monodepth_secs/image_nusc_vis}
\subsection{Multi-Camera Depth Prediction}

The performance on the nuScenes dataset is presented in Table ~\ref{tab:nusc_test}. We also present the detailed performance result of RMSE and RMSE log on all six cameras on the nuScenes dataset.
Some other methods \cite{Vitor2022FSM, wei2022surround} train with all six camera images in a frame at once, and propagate information between images during training and inference. We treat the data as six \textbf{independent} monocular image streams to train and infer with a single FSNet model. The single FSNet model aligns its predicted depth with different cameras using their intrinsic camera parameter. 

Because the cameras and LiDAR data are not synchronized, the relative poses between frames in the nuScenes dataset are not as accurate as those in the KITTI dataset. However,  experiment data show that FSNet trained with noisy poses can still produce depth predictions with the correct scale as the 3D scenes and it is robust to different camera setups.

Finally, we present some qualitative results on the validation split of the nuScenes dataset in Figure~\ref{fig:nusc_vis}. FSNet predicts depth in each image independently, and we concatenate the predictions together to form the results here. The first three rows demonstrate the network's ability to identify different objects of interest in urban road scenes. The final row presents a failure case where a bus with a huge reflective glass dominates the image. More 3D visualization is presented at the project page \url{https://sites.google.com/view/fsnet/home}, which can further show that the depths predicted from the six cameras are consistent, and we can obtain a detailed 3D perception result of the surrounding environment.

\input{monodepth_secs/table_nuscenes_detail}

\section{Discussions and Ablation Study}

In this section, we start by focusing on how each proposed component improves performance in a single-frame setting. Then, we study the parameters and decision choices in the optimization-based post-processing.

\subsection{Single-Frame Evaluation}
In Section~\ref{sec:modification}, we introduce a multi-channel output to allow organic image reconstruction at initialization to boost-trap the training. Here, we experiment with two other output settings: (1) original monodepth2 output with a bias in the output layer; (2) exponential activation. As presented in Table~\ref{tab:kitti_ablation_single}, multi-channel output performs better at most error metrics. The original output representation of monodepth2 saturates in a common depth range like $d > 10m$, which makes it difficult to train the network. We highlight that the un-scale baseline MonoDepth tends to maintain the output in range with sufficient gradients by adjusting the scale of the pose prediction. The exponential activation, though effective in monocular 3D object detection \cite{Chen2020MonoPair},  results in un-controllable activation growth in background pixels like the sky, where the correct depths are essentially infinity, corrupting the training gradients. The experiments and analysis show that the proposed multi-channel output enables stable training and produces accurate predictions.

We further present the results with a flow mask and self-distillation in Table~\ref{tab:kitti_ablation_single}. Each proposed method incrementally improves the depth estimation results. Squared Relative (Sq Rel) and Root Mean Square Error (RMSE) improve the most, which means that the proposed methods mostly prevent the network from making significant mistakes like with dynamic objects.
We note that both methods do not introduce extra cost in the inference time, but it regulates the training process to improve performance.

\input{monodepth_secs/table_ablation}

\subsection{Post-Optimization Evaluation}

There are many factors contributing to the performance of the post-optimization step. This section investigates two factors: (1) the usage of depth in the 3D SLIC algorithm and (2) the importance of each loss weight in the optimization step.

The results are presented in Table~\ref{tab:kitti_ablation}. The proposed 3D SLIC algorithm improves the segmentation quality by utilizing the depth predicted by the network, thus improving the post-optimization results. When $\lambda_0=0$, the depth prediction of each super segment will be independent, and the optimization from visual odometry cannot fully propagate throughout the map. However, when $\lambda_2=0$, we completely ignore the scale predicted by the network, and errors in visual odometry, especially at dynamic objects, will corrupt the prediction result.  

\input{monodepth_secs/table_ablation_post_opt}

%% file: monodepth_secs/image_kitti360_frames.tex
\begin{figure*}
    \centering
    \includegraphics[{width=1.0\textwidth}]{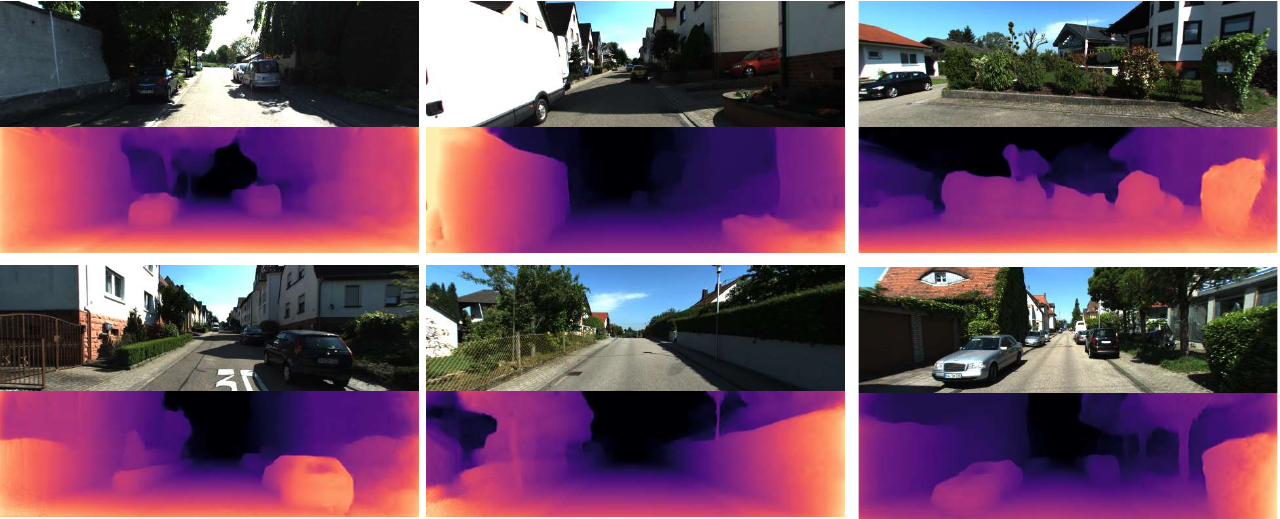}
    \caption{Prediction result sample from KITTI-360 dataset. Pixels colored in white are expected to be closer to the camera. The pictures demonstrates the network's ability to distinguish close-up objects against backgrounds. 
    }
    \label{fig:kitti360_frame_examples}
\end{figure*}

%% file: monodepth_secs/image_nusc_vis.tex
\begin{figure*}
    \centering
    \includegraphics[{width=0.9\textwidth}]{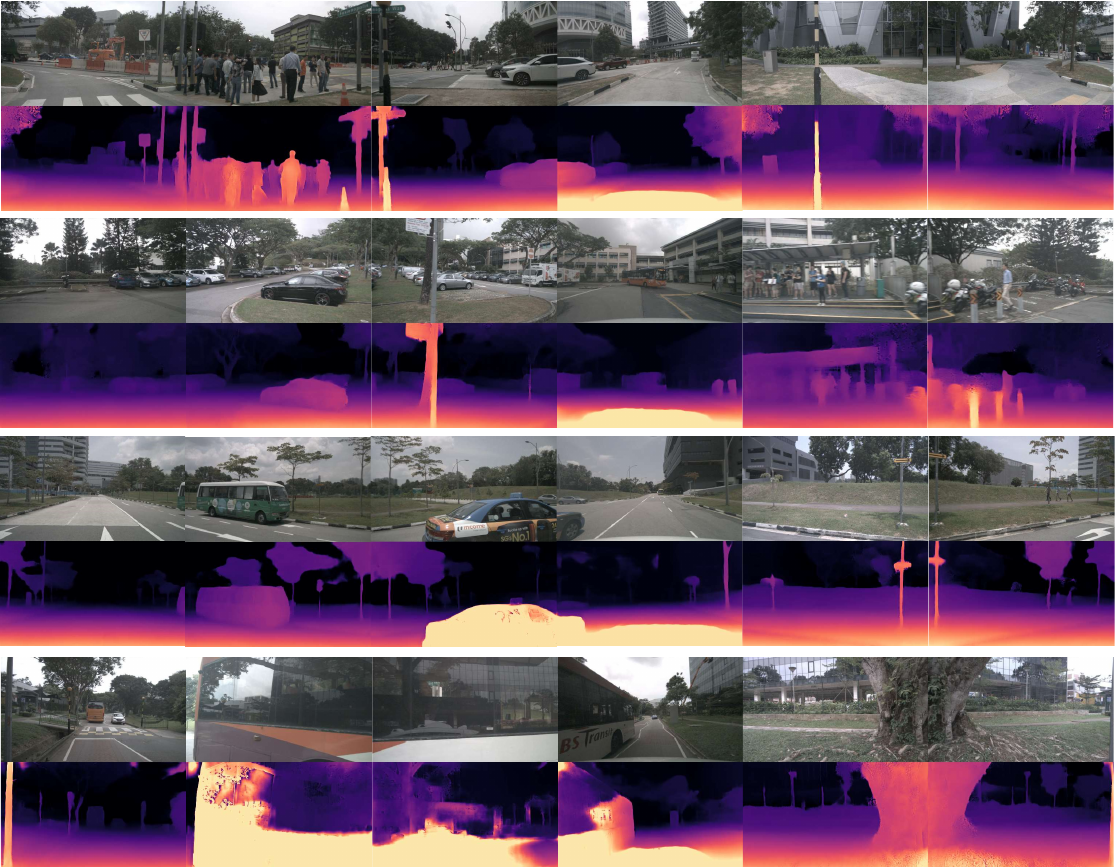}
    \caption{Self-Supervised Depth Estimation FSNet result on the nuScenes dataset. Complicated scenarios, varying camera parameters, close-up objects and large-scale specular reflection pixels make nuScenes a particularly challenging dataset.
    }
    \label{fig:nusc_vis}
\end{figure*}

%% file: monodepth_secs/table_nuscenes_detail.tex
\begin{table*}[]
\caption{ RMSE and RMSE Log of full-scale MonoDepth on nuScenes on all six cameras. For scale factor, "GT" is using LiDAR median scaling methods and "None" means we directly evaluate the error without post-processing scale. "\textbf{*}" indicates methods using sequential images in test time. }
    \label{tab:nusc_test}
    \centering
    \def\arraystretch{1.5}
    \footnotesize
    \setlength\tabcolsep{2pt}
    \begin{tabular*}{1.0\textwidth}{|l|l|cc|cc|cc|cc|cc|cc|cc|}
\cline{1-16}
                                   &                              & \multicolumn{2}{c|}{Front}                                                      & \multicolumn{2}{c|}{F.Left}                                                     & \multicolumn{2}{c|}{F.Right}                                                    & \multicolumn{2}{c|}{B.left}                                                     & \multicolumn{2}{c|}{B.right}                                                    & \multicolumn{2}{c|}{Back}                                                       & \multicolumn{2}{c|}{Avg}                                                        \\ \cline{3-16} 
\multirow{-2}{*}{\textbf{Methods}} & \multirow{-2}{*}{Fac.} & \multicolumn{1}{c|}{\cellcolor[HTML]{FFCCC9}rmse} & \cellcolor[HTML]{FFCCC9}log & \multicolumn{1}{c|}{\cellcolor[HTML]{FFCCC9}rmse} & \cellcolor[HTML]{FFCCC9}log & \multicolumn{1}{c|}{\cellcolor[HTML]{FFCCC9}rmse} & \cellcolor[HTML]{FFCCC9}log & \multicolumn{1}{c|}{\cellcolor[HTML]{FFCCC9}rmse} & \cellcolor[HTML]{FFCCC9}log & \multicolumn{1}{c|}{\cellcolor[HTML]{FFCCC9}rmse} & \cellcolor[HTML]{FFCCC9}log & \multicolumn{1}{c|}{\cellcolor[HTML]{FFCCC9}rmse} & \cellcolor[HTML]{FFCCC9}log & \multicolumn{1}{c|}{\cellcolor[HTML]{FFCCC9}rmse} & \cellcolor[HTML]{FFCCC9}log \\\cline{1-16}
MonoDepth\cite{monodepth2}                          &                              & \multicolumn{1}{c|}{6.992}                        & 0.222                       & \multicolumn{1}{c|}{6.905}                        & 0.319                       & \multicolumn{1}{c|}{7.655}                        & 0.359                       & \multicolumn{1}{c|}{5.996}                        & 0.305                       & \multicolumn{1}{c|}{7.606}                        & 0.375                       & \multicolumn{1}{c|}{6.718}                        & 0.267                       & \multicolumn{1}{c|}{6.979}                        & 0.308                       \\\cline{3-16}
\textbf{FSNet}                     &                              & \multicolumn{1}{c|}{6.999}                        & 0.225                       & \multicolumn{1}{c|}{6.652}                        & 0.312                       & \multicolumn{1}{c|}{7.703}                        & 0.362                       & \multicolumn{1}{c|}{6.137}                        & 0.311                       & \multicolumn{1}{c|}{7.047}                        & 0.362                       & \multicolumn{1}{c|}{7.265}                        & 0.275                       & \multicolumn{1}{c|}{\textbf{6.930}}               & \textbf{0.306}              \\\cline{3-16} 
\textbf{*FSNet-Opt}                & \multirow{-3}{*}{GT}         & \multicolumn{1}{c|}{6.918}                        & 0.223                       & \multicolumn{1}{c|}{6.866}                        & 0.318                       & \multicolumn{1}{c|}{7.586}                        & 0.357                       & \multicolumn{1}{c|}{5.963}                        & 0.304                       & \multicolumn{1}{c|}{7.562}                        & 0.373                       & \multicolumn{1}{c|}{6.644}                        & 0.264                       & \multicolumn{1}{c|}{\textbf{6.923}}               & \textbf{0.307}              \\\cline{1-16}
\textbf{FSNet}                     &                              & \multicolumn{1}{c|}{6.658}                        & 0.223                       & \multicolumn{1}{c|}{6.632}                        & 0.322                       & \multicolumn{1}{c|}{7.853}                        & 0.364                       & \multicolumn{1}{c|}{5.681}                        & 0.303                       & \multicolumn{1}{c|}{7.658}                        & 0.377                       & \multicolumn{1}{c|}{6.706}                        & 0.329                       & \multicolumn{1}{c|}{\textbf{6.865}}               & \textbf{0.319}              \\\cline{3-16} 
\textbf{*FSNet-Opt}                & \multirow{-2}{*}{None}       & \multicolumn{1}{c|}{6.902}                        & 0.235                       & \multicolumn{1}{c|}{6.396}                        & 0.323                       & \multicolumn{1}{c|}{7.317}                        & 0.352                       & \multicolumn{1}{c|}{4.965}                        & 0.282                       & \multicolumn{1}{c|}{6.901}                        & 0.361                       & \multicolumn{1}{c|}{6.584}                        & 0.276                       & \multicolumn{1}{c|}{\textbf{6.489}}               & \textbf{0.311}              \\\cline{1-16}
\end{tabular*} 
    
\end{table*}

%% file: monodepth_secs/table_ablation.tex
\begin{table*}
    \centering
    \caption{Abalation study of single frame prediction in FSNet on KITTI Eigen Split without scale factor.}
    \label{tab:kitti_ablation_single}
    \def\arraystretch{1.3}
    \begin{tabular*}{0.8\textwidth}{ |l|c|c|c|c|}
        \cline{1-5}
        {\bf Variants} & \cellcolor{red!25}{Abs Rel$\downarrow$} & \cellcolor{red!25}{Sq Rel$\downarrow$} & \cellcolor{red!25}{\bf RMSE$\downarrow$} & \cellcolor{red!25}{RMSE log$\downarrow$} \\ 
        \cline{1-5}
        biased MonoDepth &   0.126  &   1.033  &   5.151  &   0.201  \\
        exp MonoDepth &   0.138  &   1.122  &   5.479  &   0.220  \\
        \cline{1-5}
        MultiChannel &   0.117  &   0.936  &   4.910  &   0.202   \\
        + Flow Mask &   0.118  &   0.928  &   4.861  &   0.200\\
        + Distillation & 0.116  &   0.923  &   4.694  &   0.194\\
        \cline{1-5}
    \end{tabular*}
\end{table*}

%% file: monodepth_secs/table_ablation_post_opt.tex
\begin{table*}
    \centering
    \caption{Abalation study of post-processing in FSNet on KITTI Eigen Split without scale factor.}
    \label{tab:kitti_ablation}
    \def\arraystretch{1.3}
    \begin{tabular*}{0.8\textwidth}{ |l|c|c|c|c|}
        \cline{1-5}
        {\bf Methods} & \cellcolor{red!25}{Abs Rel$\downarrow$} & \cellcolor{red!25}{Sq Rel$\downarrow$} & \cellcolor{red!25}{\bf RMSE$\downarrow$} & \cellcolor{red!25}{RMSE log$\downarrow$}  \\ 
        \cline{1-5}
        FSNet(post-opt) &   0.109  &   0.866  &   4.450  &   0.189  \\
        \cline{1-5}
        w 2D SLIC &   0.110  &   0.881  &   4.495  &   0.190   \\
        w $\lambda_0=0$ &   0.111  &   0.924  &   4.513  &   0.194\\
        w $\lambda_2=0$ & 0.121  &   0.977  &   4.585  &   0.199\\
        \cline{1-5}
    \end{tabular*}
\end{table*}

%% file: monodepth_secs/sec-4-conclusion.tex
\section{Conclusion}

Your conclusion effectively summarizes the key aspects and contributions of the chapter on FSNet. Here's a refined version for enhanced clarity and coherence:

In this chapter, we introduced FSNet, a novel approach to self-supervised monocular depth prediction, specifically tailored for robotic applications. This method represents a significant advancement in the field, offering a full-scale, self-supervised monocular depth prediction framework. Our journey began with detailed experiments to deepen our understanding of the training dynamics of unsupervised MonoDepth prediction networks. This led to the development of a multi-channel output representation, which proved crucial for ensuring stable training from the outset.

Further innovations included the integration of an optical-flow-based dynamic object removal mask and a self-distillation training strategy. These enhancements not only bolstered the training performance but also contributed significantly to the overall effectiveness and robustness of FSNet. Additionally, we implemented a novel post-processing technique that leverages sparse 3D points generated through visual odometry, markedly improving the algorithm’s performance during testing.

Extensive experimental evaluations were conducted to validate the effectiveness of FSNet. These tests confirmed the efficacy of our approach in diverse settings, solidifying its potential for wide application in robotic systems.

A notable advantage of FSNet is its reliance on readily available data types: sequences of images and corresponding poses, commonly produced by calibrated robots equipped with localization modules. This reliance ensures the practicality and accessibility of our method for real-world robotic applications. At test time, FSNet is capable of generating detailed 3D environmental information efficiently, without imposing significant additional computational burdens. The modular design of FSNet not only enhances its performance but also simplifies the development and integration processes, making it a valuable addition to the toolkit of robotic and autonomous systems developers.

%% file: monodepth_secs/sec-5-appendix.tex
\section{Appendix}

This appendix extends the application of our proposed FSNet to fisheye cameras, delving into the unique challenges and methodologies adapted for this context.

\begin{figure*}[htb]
    \centering
    \includegraphics[{width=0.9\textwidth}]{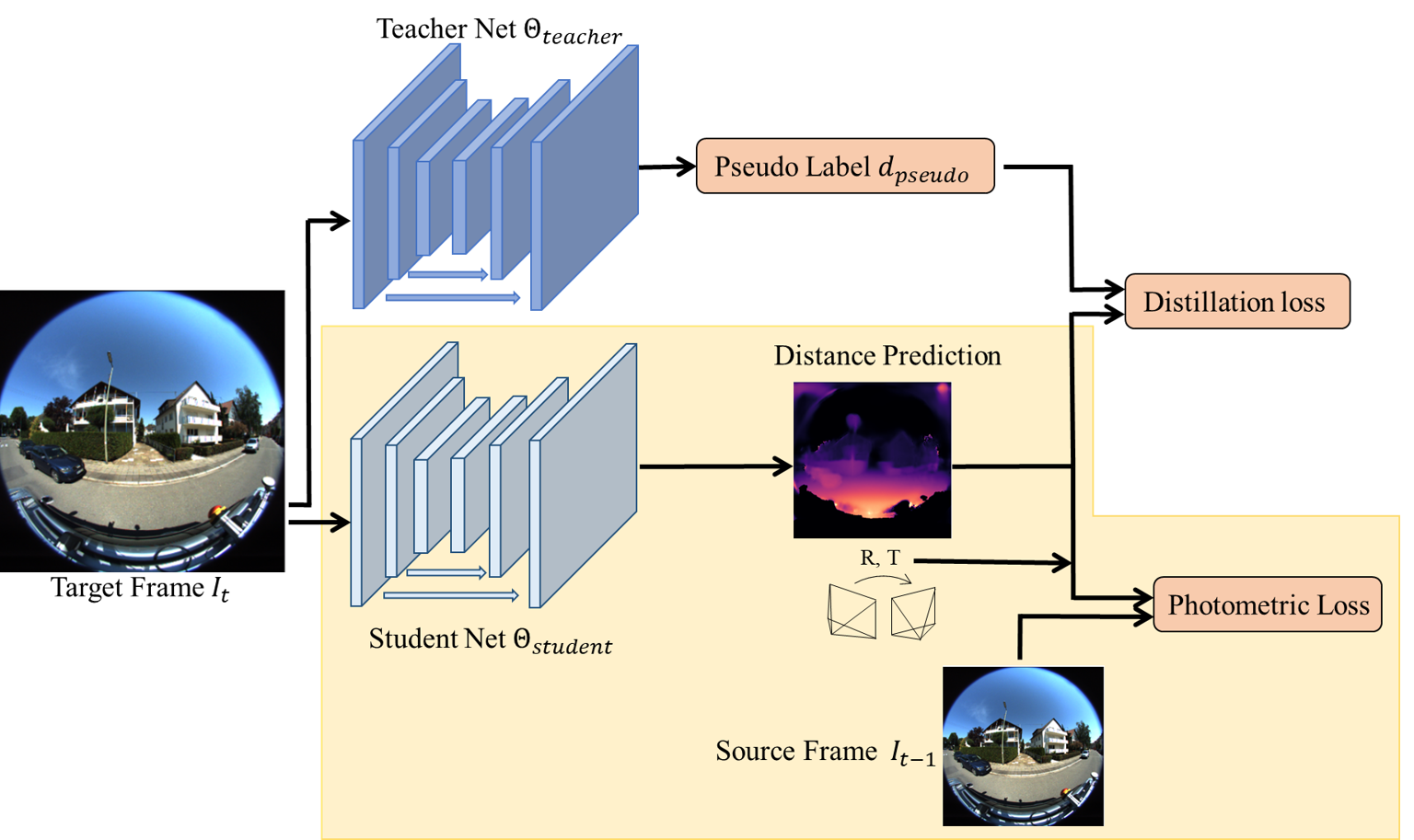}
    \caption{
        Training Scheme for fisheye cameras. Similar to the one for perspective images.
    }
    \label{fig:monodepth_fisheye_struct}
\end{figure*}

\subsection{Extension to Fisheye Cameras}

Our approach involves adapting the FSNet, originally designed for perspective images, to fisheye cameras using a self-supervised and self-distillation framework. Given the complex distortion characteristics inherent to fisheye images, we train and test our network on images that retain their original, unrectified fisheye properties.

\begin{table*}[htb]
\caption{Performance of full-scale monocular depth FSNet on KITTI360 fisheye The pink columns are error metrics, the lower the better; the blue columns are accuracy metrics, the higher the better.}
    \label{tab:fisheye_kitti360_test}
    \centering
    \small
    \def\arraystretch{1.3}
    \setlength\tabcolsep{4pt}
    \begin{tabular*}{1.0\textwidth}{|l|c|c|c|c|c|c|c|}
    \cline{1-8} {\bf Methods} &  \cellcolor{red!25}{Abs Rel } & \cellcolor{red!25}{Sq Rel} & \cellcolor{red!25}{\bf RMSE} & \cellcolor{red!25}{RMSE LOG} & \cellcolor{blue!25}{$\delta<1.25$} & \cellcolor{blue!25}{$\delta<1.25^2$} & \cellcolor{blue!25}{$\delta<1.25^3$} \\ 
    \cline{1-8}
     FSNet (all depth) & 0.161 & 0.594 & 2.348 & 0.823 & 0.771 & 0.881 & 0.927 \\
     \cline{1-8}
    FSNet (within 8m)  & 0.084 & 0.318 & 0.772 & 0.185 & 0.940 & 0.980 & 0.991 \\
      \cline{1-8}
    \end{tabular*}
\end{table*}

We focus on two predominant fisheye camera models: the Mei unified camera model and the pinhole fisheye camera model, as outlined in the OpenCV library \cite{opencv_library}. 

Considering projecting a 3D points $\vec{X}=[x, y, z]$ in the camera frame to image plane $[x_{2d}, y_{2d}]$, the Mei unified camera model can be described by the following set of equations:
\begin{align*}
    \tilde{X} &= \frac{\vec{X}}{\sqrt{x^2 + y^2 + z^2}} \\
    \tilde{x_s} &= \frac{\tilde{x}}{\tilde{z} + \xi} \\
    r &= \sqrt{\tilde{x_s^2} + \tilde{y_s^2}} \\
    x_u &= x_s \cdot (1 + k_1 r^2 + k_2 r^4) \\
    x_{2d} &= \gamma_x \cdot x_u + u_0.
\end{align*}
In contrast, the pinhole model is represented by a different equation set:
\begin{align*}
    r^2 &= (\frac{x}{z})^2 + (\frac{y}{z})^2 \\
    \theta &= \text{atan}(r) \\
    \theta_u &= \theta \cdot (1 + k_1 \theta^2 + k_2 \theta^4 + k_3 \theta^6 + k_4 \theta^8) \\
    x_{2d} &= \frac{\theta}{z\cdot r} \cdot f_x \cdot x + c_x .
\end{align*}
We implement these models' projection and inverse-projection processes, essential for image reconstruction during training.

Unlike traditional depth prediction networks, which focus on the $z$-axis distance, our network for fisheye cameras predicts the radial distance from the camera center, $l=\sqrt{x^2 + y^2 + z^2}$, to accommodate the distinctive projection properties of fisheye lenses. This approach significantly improves training stability.  

Additionally, we address the challenge of ego vehicle capture in fisheye images. Following the methodology of \cite{Vitor2022FSM}, we manually mask the ego vehicle in each camera setup, ensuring the network's accuracy in real-world scenarios.

\subsection{Evaluation on KITTI360}
\begin{figure*}[]
    \centering
    \includegraphics[{width=0.9\textwidth}]{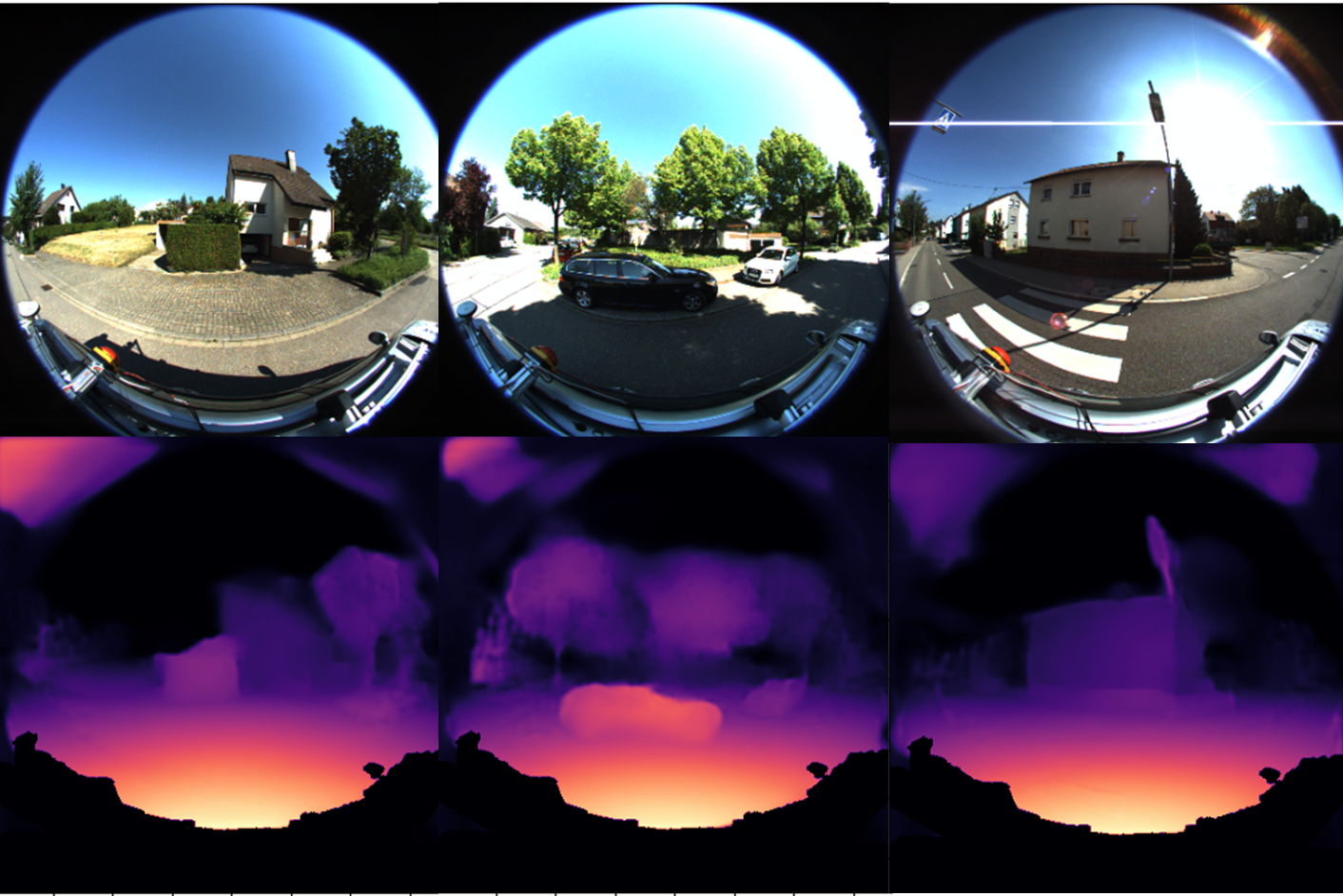}
    \caption{
        Qualitative results of fisheye depth prediction on KITTI360.
    }
    \label{fig:monodepth_fisheye_qualitative}
\end{figure*}

We evaluated our fisheye depth prediction models on the KITTI 360 dataset, which includes fisheye cameras calibrated using the Mei unified model. The training followed the chapter's proposed split, and the results, detailed in figure~\ref{tab:fisheye_kitti360_test}, demonstrate high accuracy, particularly for objects near the camera. Figure~\ref{fig:monodepth_fisheye_qualitative} showcases qualitative results from this evaluation.

\subsection{Deployment to On-Board System}

\begin{figure*}[]
    \centering
    \includegraphics[{width=0.9\textwidth}]{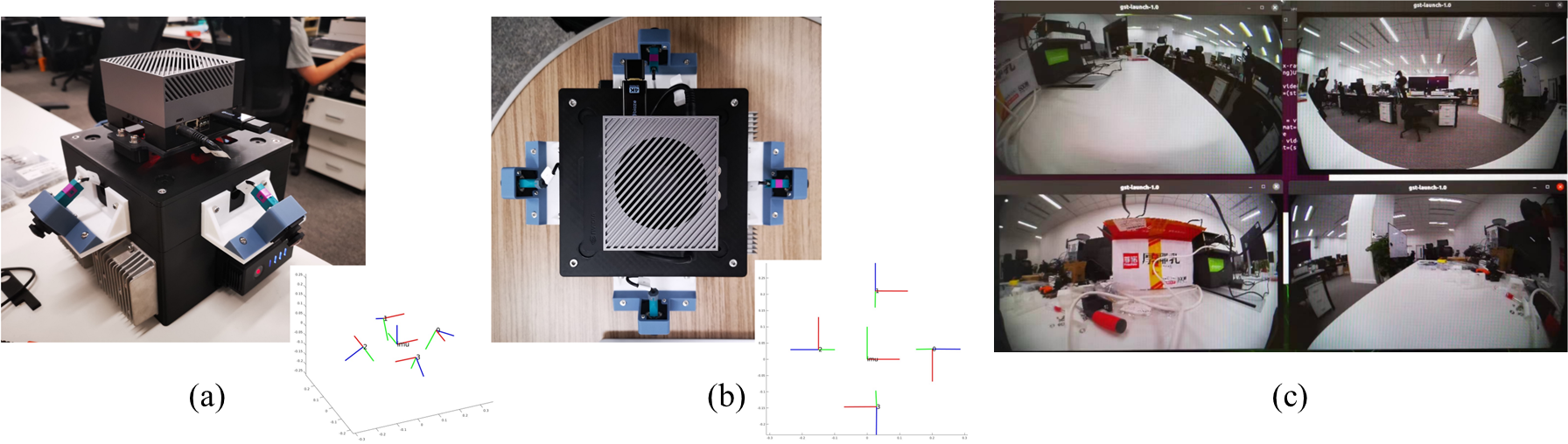}
    \caption{
        The platform we used for testing. It is equipped with four fisheye cameras and an Orin computing board. Some examples of the images are presented in (c).
    }
    \label{fig:monodepth_fisheye_hardware}
\end{figure*}

For real-world applications, FSNet was trained and deployed using a hardware setup illustrated in figure~\ref{fig:monodepth_fisheye_hardware}. This system encompasses four fisheye cameras for comprehensive environmental perception and an Orin computing board for on-the-fly data processing and real-time inference.

\begin{figure*}[]
    \centering
    \includegraphics[{width=0.9\textwidth}]{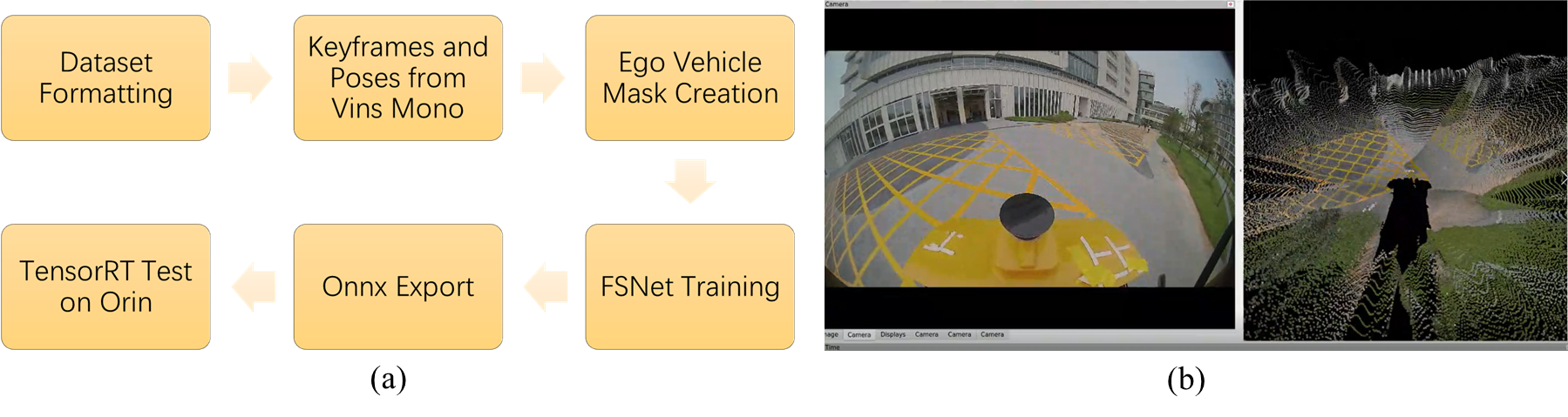}
    \caption{
        The platform we used for testing. It is equipped with four fisheye cameras and an Orin computing board. Some examples of the images are presented in (c).
    }
    \label{fig:monodepth_fisheye_deployment}
\end{figure*}

The deployment process, depicted in figure~\ref{fig:monodepth_fisheye_deployment}, integrates VinsMono for keyframe identification in front-view imagery, facilitating pose estimation between significant frames. Through extensive training and optimization, the model achieved a processing speed of 10Hz for the simultaneous inference of four images on the Orin platform, optimized for half-float computation. Additional visualizations and results are available at  \url{https://www.bilibili.com/video/BV1Qo4y1j7NL/?spm_id_from=333.999.0.0}.

%% file: chapter/07_conclusion.tex
\section{Summary}
This thesis have rigorously explored the construction of high-performance and efficient vision-based 3D perception frameworks specifically tailored for autonomous driving applications. Our focal point has been the development and deployment of models that address both 3D object detection and depth prediction, even on platforms equipped with a single camera.

Initially, Chapter 2 introduced a ground-aware 3D object detector that leverages the concept of 3D anchors and incorporates a Ground-Aware Convolutional (GAC) module. This detector ingeniously uses geometric ground plane priors as supplementary data during network inference to mitigate the accuracy problem of a monocular 3D perception system. We were able to achieve superior detection accuracy without sacrificing computational efficiency. Experiments demonstrate the leading detection accuracy of the proposed detector while keeping computational complexity approachable. 

Subsequently, Chapter 3 presented YOLOStereo3D, a stereo 3D object detector that harmoniously balances efficiency and performance for networks with multi-view geometric reasoning. Drawing inspiration from our monocular 3D detector, this model introduces several pivotal contributions, 1) the introduction of 3D anchors and ground-aware anchor filtering from monocular detectors into stereo detectors; 2) a multi-scale stereo feature extraction pipeline to augment the monocular detection branch, which includes light-weight cost volume construction and hierarchical multi-scale feature fusion structure; 3) a training strategy with auxiliary loss from explicit stereo matching and data augmentation tuned for the task of stereo 3D detection. Our experiments substantiate that YOLOStereo3D sets new benchmarks for speed and accuracy in publicly available tests.

In Chapter 4, acknowledging the limitations associated with obtaining 3D labels for camera-only platforms, which hinder the deployment of the 3D detectors proposed in Chapter 2 and Chapter 3, a versatile joint dataset training regimen for 3D detection networks is introduced to improve the data utilization ability for monocular 3D detection. The detectors were adjusted to be cognizant of camera parameters and geometry, thereby enabling them to generate accurate predictions across diverse datasets. Then the selective training strategy which tailors the model to adapt to differences in annotated classes in different datasets was formulated. Finally, a robust model was trained to be capable of performing high-quality predictions across multiple prominent datasets including KITTI, nuScenes, ONCE, bdd100k, and Cityscapes, essentially transferring 3D knowledge to datasets without 3D labels. Furthermore, the resulting model demonstrates significantly better zero-shot performance on self-collected data even without labeling.

Finally, Chapter 5 unveils the FSNet framework dedicated to unsupervised full-scale depth prediction, which aims for a comprehensive pipeline to utilize image sequence data for 3D perception. FSNet utilizes the related poses of neighboring frames directly as training input and trains the network with reconstruction losses. The framework includes four major contributions: 1) multichannel output for stable training without PoseNet; 2) optical flow masks to tackle dynamic objects using poses as additional priors; 3) self-distillation mechanism for high-performance training; 4) optimization-based post-processing to further optimize the final results online. The pipeline is first tested on multiple public datasets with perspective images. Then the pipeline is extended to train depth predictions on fish-eye cameras, achieving satisfactory results on public datasets. The pipeline was further tested on a self-built platform, achieving high-performing 360-degree 3D perception on embedded computing devices, obviating the need for LiDAR.

In summary, this thesis delineates multiple viable pathways for achieving robust 3D perception in autonomous driving scenarios using only camera-based systems. It presents a suite of algorithms for both 3D object detection and depth prediction, each substantiated by extensive empirical validations on public datasets and real-world applications.

\section{Outlook}

The domain of 3D vision is advancing at a remarkable pace within the autonomous driving industry. This thesis has concentrated on the adaptability of the methods proposed, with a particular emphasis on single-frame input garnered from monocular or stereo cameras.

While our work is grounded in the analysis of instantaneous visual information, other research endeavors are pioneering the integration of 3D perception using surround-view camera arrays and the reconstruction of environments through extended image sequences. These methodologies are gaining traction and offering new perspectives on environmental mapping and navigation.

Bird's Eye View (BEV) and Occupancy Prediction stand at the forefront of research within autonomous driving, fueled by robust data annotation platforms within automotive enterprises. The prevalent use of supervised learning, underpinned by copious LiDAR-annotated datasets, allows for the refinement of these predictive models. The innate advantage of BEV/Occupancy representations is their facilitation of sensor fusion and their seamless incorporation into prediction and planning frameworks. Nevertheless, the economic and logistical demands of data annotation and platform development restrict their application on a more expansive scale. Perspective view-based methodologies, such as those outlined in this thesis, continue to be a mainstay in cost-sensitive autonomous platforms.

Neural Radiance Fields (NeRF) have rapidly become a buzzword in the computer vision arena. We witness an evolution of works that are reconstructing 3D environments from image sequences, some with exceptional speed. Yet, the efficacy of NeRF-based approaches is contingent upon the precise calibration of camera poses and a diverse range of environmental views devoid of dynamic elements — conditions that are challenging to meet in outdoor autonomous driving scenarios.

Nonetheless, the synthesis of burgeoning technologies harbors the potential for transformative developments in vision-only 3D perception methodologies. For instance, training BEV/Occupancy networks with high-fidelity ground truths, derived from extensive NeRF-based 3D reconstructions, or enhancing these networks through differential rendering techniques borrowed from NeRF, could herald a new era of sophisticated vision-centric perception systems. We posit that continued investigation into these integrative approaches may yield a more refined and comprehensive suite of vision-only 3D perception tools.

%% file: chapter/s6_appendices.tex
\appendix

\chapter{List of Publications}
\section*{Journal Publications}

\begin{enumerate}
    \item \textbf{Yuxuan Liu}, Zhenhua Xu, Huaiyang Huang, Lujia Wang and Ming Liu, "FSNet: Redesign Self-Supervised MonoDepth for Full-Scale Depth Prediction for Autonomous Driving," in IEEE Transactions on Automation Science and Engineering, doi: 10.1109/TASE.2023.3290348.
    \item Zhenhua Xu, \textbf{Yuxuan Liu}, Lu Gan, Xiangcheng Hu, Yuxiang Sun, Ming Liu, Lujia Wang, CsBoundary: City-Scale Road-Boundary Detection in Aerial Images for High-Definition Maps, IEEE Robotics and Automation Letters 7, no. 2 (2022): 5063-5070. 
    \item Zhenhua Xu, \textbf{Yuxuan Liu}, Lu Gan, Yuxiang Sun, Lujia Wang, and Ming Liu, "RNGDet: Road Network Graph Detection by Transformer in Aerial Images", IEEE Transactions on Geoscience and Remote Sensing (TGRS), 2022
    \item Xinxing Chen, Huaiyang Huang, \textbf{Yuxuan Liu}, Jiqing Li, Ming Liu. Robot for automatic waste sorting on construction sites. Automation in Construction, Volume 141, 2022, 104387, ISSN 0926-5805.
    \item \textbf{Yuxuan Liu}, Yixuan Yuan and Ming Liu, “Ground-aware Monocular 3D Object Detection for Autonomous Driving,” in IEEE Robotics and Automation Letters (RA-L), vol. 6, no. 2, pp. 919-926, April 2021, doi: 10.1109/LRA.2021.3052442.
    \item Peide Cai, Hengli Wang, Huaiyang Huang, \textbf{Yuxuan LIU}, Ming Liu, “Vision-Based Autonomous Car Racing Using Deep Imitative Reinforcement Learning,” IEEE Robotics and Automation Letters (RA-L), 2021 (Early Access).
    \item Peng Yun, \textbf{Yuxuan LIU} and Ming Liu, “In Defense of Knowledge Distillation for Task Incremental Learning and its Application in 3D Object Detection,” IEEE Robotics and Automation Letters (RA-L), 2021 (Early Access).
    
\end{enumerate}

\section*{Conference Publications}
\begin{enumerate}
    \item Zhenhua Xu, \textbf{Yuxuan Liu}, Yuxiang Sun, Ming Liu and Lujia Wang, CenterLineDet: Road Lane CenterLine Graph Detection With Vehicle-Mounted Sensors by Transformer for High-definition Map Creation, International Conference on Robotics and Automation (ICRA), 2023, London, Britain.
    \item \textbf{Yuxuan Liu}, Zhenhua Xu, Ming Liu, Star-Convolution for Image-Based 3D Object Detection, International Conference on Robotics and Automation (ICRA), 2022, Philadelphia, the US.
    \item Zhenhua Xu, \textbf{Yuxuan Liu}, Lu Gan, Xiangcheng Hu, Yuxiang Sun, Ming Liu, Lujia Wang, CsBoundary: City-Scale Road-Boundary Detection in Aerial Images for High-Definition Maps, International Conference on Robotics and Automation (ICRA), 2022, Philadelphia, the US.
    \item \textbf{Yuxuan LIU}, Lujia Wang, Ming Liu, YOLOStereo3D: A Step Back to 2D for Efficient Stereo 3D Detection, International Conference on Robotics and Automation (ICRA), 2021, Xi An, China.
    \item \textbf{Yuxuan LIU}, Ming Liu, Ground-aware Monocular 3D Object Detection for Autonomous Driving, International Conference on Robotics and Automation (ICRA), 2021 , Xi An, China.
    \item Peide Cai, Hengli Wang, Huaiyang Huang, \textbf{Yuxuan Liu}, Ming Liu, Vision-Based Autonomous Car Racing Using Deep Imitative Reinforcement, IEEE/RSJ International Conference on Intelligent Robots and Systems (IROS), 2021, Prague, Czech Republic.
\end{enumerate}

\section*{Workshop Publications}
    \begin{enumerate}
        \item Hengli Wang*, \textbf{Yuxuan Liu*}, Huaiyang Huang*, Yuheng Pan*, Wenbin Yu, Jialin Jiang, Dianbin Lyu, Mohammud J. Bocus, Ming Liu, Ioannis Pitas, Rui Fan, ATG-PVD: Ticketing Parking Violations on A Drone, European Conference on Computer Vision (ECCV) Workshops, 2020, Glasgow, UK. 
    \end{enumerate}
\endappendix